\documentclass[10pt,journal,compsoc]{IEEEtran}
\usepackage{makecell}
\usepackage{graphicx}

\usepackage{multirow}

\usepackage{hhline}

\ifCLASSOPTIONcompsoc
  \usepackage[nocompress]{cite}
\else
  \usepackage{cite}
\fi
\ifCLASSINFOpdf
\else
\fi
\begin{document}
%
\title{Writer Identification Using Microblogging Texts for Social Media Forensics}
%
%
%
%

\author{Fernando~Alonso-Fernandez,
        Nicole~Mariah~Sharon~Belvisi,
        Kevin~Hernandez-Diaz, \\
        Naveed~Muhammad,
        and Josef~Bigun,~\IEEEmembership{Fellow,~IEEE}
\IEEEcompsocitemizethanks{\IEEEcompsocthanksitem F. Alonso-Fernandez, N. Sharon Belvisi, K. Hernandez-Diaz and J. Bigun are with the \textit{School of Information Technology (ITE)}, \textit{Halmstad University}, Sweden (email: feralo@hh.se, nicbel18@student.hh.se, kevin.hernandez-diaz@hh.se, josef.bigun@hh.se).\protect\\
\IEEEcompsocthanksitem N. Muhammad is with the \textit{Institute of Computer Science}, \textit{University of Tartu, Estonia} (email: naveed.muhammad@ut.ee)}
}

%
%

\markboth{Journal of \LaTeX\ Class Files,~Vol.~XX, No.~X, July~2020}%
{Shell \MakeLowercase{\textit{et al.}}: Bare Demo of IEEEtran.cls for Biometrics Council Journals}
%



\IEEEtitleabstractindextext{%
\begin{abstract}
Establishing authorship of online texts is fundamental to combat cybercrimes. Unfortunately, text length is limited on some platforms, making the challenge harder. We aim at identifying the authorship of Twitter messages limited to 140 characters. We evaluate popular stylometric features, widely used in literary analysis, and specific Twitter features like URLs, hashtags, replies or quotes. 
We use two databases with 93 and 3957 authors, respectively. 
%
%
We test varying sized author sets and varying amounts of training/test texts per author. 
%
%
Performance is further improved by feature combination via automatic selection.
With a large amount of training Tweets ($>$500), a good accuracy (Rank-5$>$80\%) is achievable with only a few dozens of test Tweets, even with several thousands of authors. 
With smaller sample sizes
(10-20 training Tweets), the search space can be diminished by 9-15\% while keeping a high chance that the correct author is retrieved among the candidates. 
In such cases, automatic attribution can provide significant time savings to experts in suspect search.
For completeness, we report verification results. With few training/test Tweets, the EER is above 20-25\%, which is reduced to $<15\%$ if hundreds of training Tweets are available.
%
%
%
%
We also quantify the computational complexity and time permanence of the employed features.
\end{abstract}

}

\maketitle

\IEEEdisplaynontitleabstractindextext

%
\IEEEpeerreviewmaketitle

\IEEEraisesectionheading{\section{Introduction}\label{sec:introduction}}

%
%
%
%
\IEEEPARstart{D}{igital} communication technologies such as social media, SMS, emails, forums, chats, or blog post have enabled faster and more efficient ways to exchange information.
In many cases, it is possible to remain anonymous, something that unfortunately has given rise to a number of cybercrimes. 
In this situation, determining the author of a digital text with sufficient reliability is an important issue for forensic investigation, in order to combat crimes such as cyberbullying, cyberstalking, frauds, ransom notes, etc. \cite{Nirkhi2013}. 

Before the technology era, handwriting texts was the primary form of written communication.
Scanned texts were then analyzed to determine authorship \cite{[Sevilla10],[Alonso10a]}.
However, digital texts demand new methods to look for authorship evidence.
An added difficulty is that some platforms limit the length of messages to just a few characters, making the issue more challenging \cite{[Rocha17tifsTweets]}.
Additionally, uses of proxy Internet addresses (URLs), hashtags, mentions
of other users in the text, or re-posting/quoting a text written by others are common.
%
%
To determine the identity of the individual behind the messages, geo-location or IP addresses could be used, but these can be concealed easily.
Signals captured from the interaction with the device could also be used, such as keystroking \cite{Banerjee12KeystrokeSurvey}
or 
touch-screen signals \cite{Tolosana19_TMC_BioTouchPass},
but they usually demand dedicated applications to capture such data that the criminal can simply deactivate.
Therefore, in many cases, the text is the only evidence available.

Accordingly, this work aims at analyzing methods to identify the writer of 
short digital texts limited to 140 characters
(Twitter posts). 
Authorship analysis relies on 
that every person has a specific writing style which is distinct from 
any other individual, so it is possible to 
connect the text and its author \cite{[Srihari02]}.
We are interested in 
\textit{attribution} or \textit{identification} of the 
author of an anonymous text, in whose case features of the anonymous text have to be compared against a whole database of texts whose authorship is known. This is the traditional identification mode in biometrics \cite{[Jain16]}, and the closest author of the database is assigned as the author of the unknown text.
In a different perspective, the task might be to compare two pieces of text, one anonymous and one written by a known author, to answer if they have been written by the same person. This is known as \textit{authorship verification}, finding applicability for example in plagiarism detection, or in different persons claiming to be the author \cite{HOLMES95}.
Although it is not our main focus, we also report verification accuracy with the employed features, given that identification and verification statistics
do not necessarily map on each other one-to-one
\cite{[DeCannRoss13btasROCCMCmenagerie]}.
%
A third approach, that we will not address, is author \textit{profiling}. This aims at building a psychological or sociological profile of the author since it is believed that indicators such as
gender 
\cite{Cheng11_gender_text}
or education level \cite{Juola03_education_from_texts}
of a person can be revealed from his/her texts.

This paper extends
a conference study \cite{Belvisi20}
%
with two completely new databases, comprising more users (93 and 3957 vs. 40) and more texts (2692 and 1972 vs. 120-200 Tweets per user on average).
We also use a substantially bigger set of stylometric features, and evaluate the impact of the amount of authors in the database and of text employed to model users' identity, both at enrolment and at testing time.
The features employed captures the writing style at different levels, from individual characters or groups or characters, to the type of words used, or the organization of sentences.
We also analyze specific particularities of Tweet messages, namely, the use of URLs, mentions of other users, hashtags, and quotations,
whose use in the literature is restricted 
%
\cite{Sultana14icc,Sultana17thms,Sultana18icsmc}. Our experiments show that such particular features help to improve the performance when added to any of the other stylometric features. 

In overall terms, some features already achieve a 97-99\% Rank-1 identification accuracy if there is sufficient data per user for enrolment and testing ($>$500 Tweets), and if the amount of authors is of few hundreds only.
In an scenario more suited to social media forensic analysis (e.g. 
20 test Tweets per user or less) our experiments show that there is still plenty of room for improvement.
If training data can be kept high ($>$500 Tweets), Rank-5 with the best individual system can be above 65\%  with several thousands of authors (1950). 
However, with the same amount of authors, Rank-5 goes down to less than 35\% if the training data is reduced to a few dozens of Tweets.
Regarding verification experiments, the amount of authors in the database is observed to have much less impact, with the EER increasing a small percentage only when going from hundreds to thousands of authors. 
%
%
%
To further improve the accuracy when few data is available, 
we combine the features to form an extended feature vector, with the best combination found by automatic feature subset selection.
The experiments show that with a subset that represents 35\% of the feature space, identification and verification accuracy can be improved for any amount of authors or training/test Tweets. 
%
%
%
%
%
%
The features are also analyzed from the point of view of computation times and permanence over time.
The extraction time is in the range of milliseconds per Tweet on a regular laptop, and feature comparison is done very quickly (in microseconds).
%
Also, the performance of all features decreases when the time between training and test Tweets increases, although their robustness is observed to be different.

The remainder of this section presents a literature review on the topic of authorship analysis of digital texts, as well as the contributions of this paper.
Section~\ref{sect:features} describes the features employed for author identification.
The experimental framework, including database and protocol, is given in Section~\ref{sect:db_protocol}.
Extensive experimental results are provided in Section~\ref{sect:results}, followed by conclusions in Section~\ref{sect:conclusion}.

\begin{figure}[t]
\centering
        \includegraphics[width=0.3\textwidth]{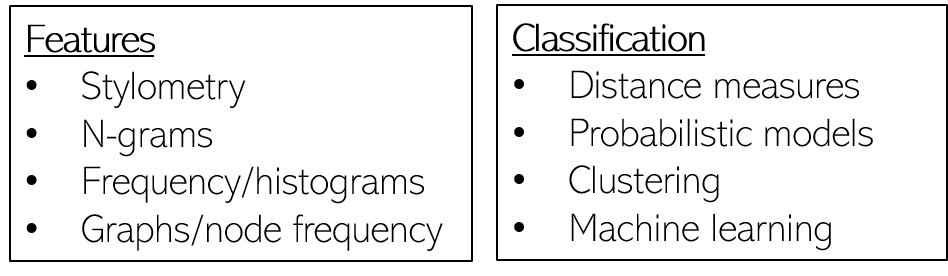}
\caption{Summary of features and classification methods employed for online authorship analysis in the literature.}
\label{fig:SOA}
\end{figure}

\begin{figure}[t]
\centering
        \includegraphics[width=0.3\textwidth]{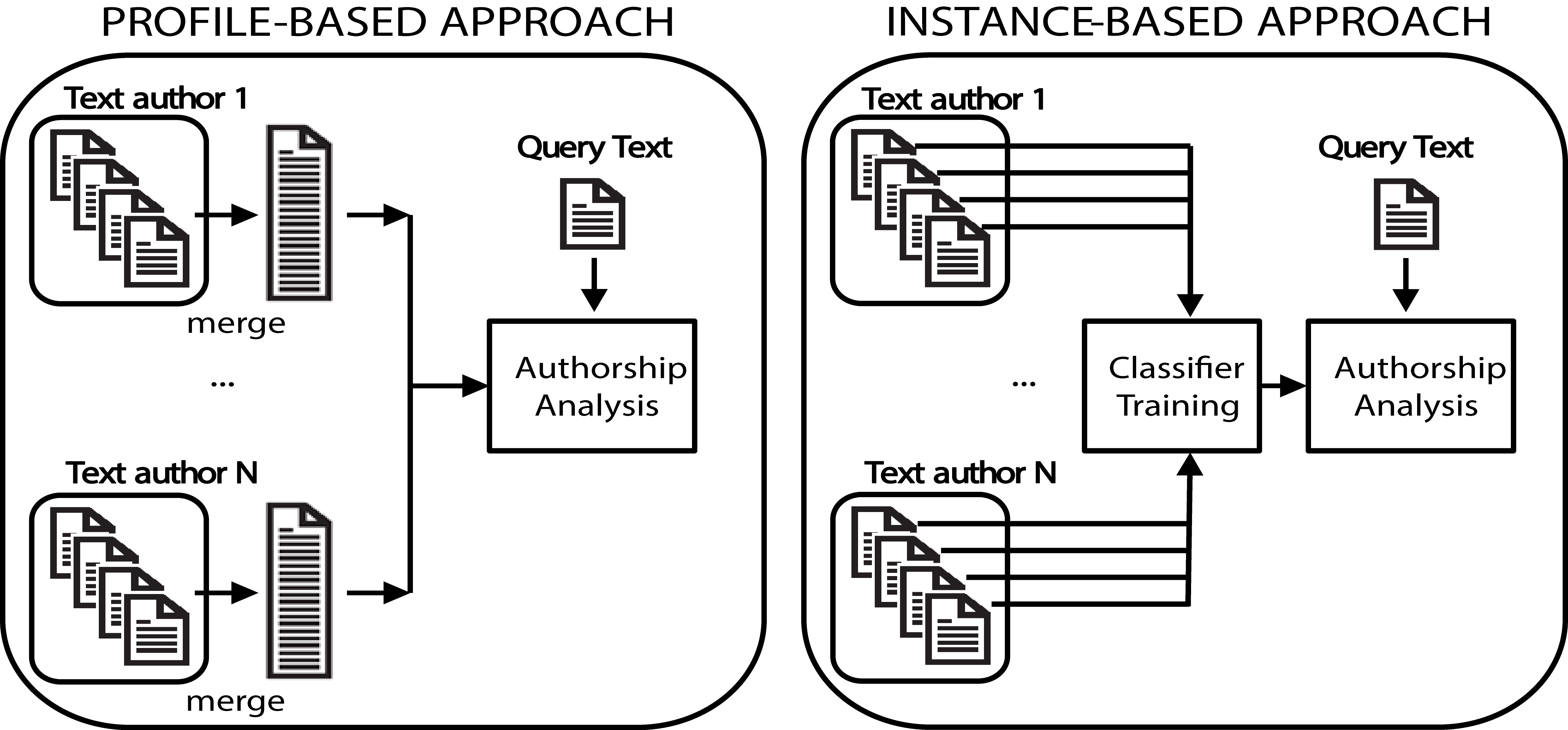}
\caption{Approaches to handle the set of documents available per author.}
\label{fig:approaches_models}
\end{figure}

\begin{table*}[htb]
\scriptsize
\begin{center}
\begin{tabular}{cccccccc}

\multicolumn{8}{c}{} \\

Ref  & Source & Language & Users & Data & Features & Classification & Mode  \\ \hline 

\cite{Abbasi05} & Forums & Ar & 20 & 20 msg/pers &  Stylometry & DT, SVM & Identif \\ \hline

\cite{Mohtasseb09}   & Blog Posts & &  93 & 17.7K total, 200 post/pers  & Stylometry &  SVM, NB  & Identif \\ \hline

\cite{Pillay10}  & Forums &  &  5-100  & 17-161 msg/pers & Stylometry & Clustering, DT, NB  & Identif \\ \hline

\cite{Koppel11}   &  Blog Posts &  &  10K  & 2K words/pers  & $n$-grams & Distance & Identif \\ \hline

\cite{Cristani12}   & Chat  &  It  &  77  & 615 words/pers  &  Stylometry & Distance & Identif \\ \hline

\cite{Amuchi12}  & Chat &  &  10  & 341 posts total  & Stylometry, $n$-grams & Distance, NB, SVM & Identif \\ \hline

\textbf{}\cite{Donais13}  & \textbf{SMS}  &  &  81  & 2K total, $<$50 /pers  & n/a &  Modified NB & Identif \\ \hline

\textbf{\cite{Ragel13}}   & \textbf{SMS}  &  En, Ch  & 70   & $>$50 SMS/pers   & $n$-grams & Distance & Identif \\  \hline

\cite{Inches13}  & Chat &  &  4.6-19K  & 79K-93K msg total & Term frequency &  Distance, KL diverg. & Identif \\ \hline

\cite{Marukatat14}  & Forums  &  Th &  6  & 25 msg/pers, 143 words/msg  & Stylometry &  SVM, DT & Identif \\ \hline

\cite{Johnson14}   & Emails &  & 176    & 63K emails total  &  Stylometry, $n$-grams & Jaccard coeff. & Identif \\ \hline

\textbf{\cite{Okuno14}}   & \textbf{Tweets} &  & 10K   & 120 Tweets/pers  & POS $n$-grams  & Distance & Identif \\ \hline

\cite{Yang14}  & Blog Posts &  &  19.3K  & 678K post, 7250 words/pers & Stylometry, $n$-grams &  LR & Identif \\ \hline

\textbf{\cite{Sultana14icc,Sultana17thms}}  & \textbf{Tweets} & En & 241  & 800 Tweet/pers &  Interaction Profile &  Node Similarity & Ver+Id  \\  

\textbf{\cite{Sultana18icsmc}}  &  &  & 100  &  &  Temporal Profile &  Distance, Jaccard coeff. & Verification  \\  \hline

\textbf{\cite{Brocardo14pst}}, \cite{Brocardo14aina}, \cite{Brocardo14cits}, \textbf{\cite{Brocardo19book}}  & Email & En & 150 & $>$200K msg, 757 msg/pers &  Stylometry, $n$-grams &  SVM, LR, DBN  & Verif  \\ 

  & \textbf{Tweets} & En & 100 & 3.2K Tweet/pers &   &    &   \\  \hline

\textbf{\cite{[Azarbonyad15tweet_time_evolution]}}  & Email & En & 15 & 3200 msg/pers &  $n$-grams &  Maximum Likelihood  & Identif  \\ 

  & \textbf{Tweets} & En & 133 & 1.8K Tweet/pers &   &    &   \\  \hline

\cite{Nirmal15}  & Emails &  & 50   & $>$200 emails/pers & Graphs, NF, PM  &  SVM  & Identif  \\ \hline

\textbf{\cite{[Rocha17tifsTweets]}}  & \textbf{Tweets} & En & 10K & 10M Tweets total &  $n$-grams &  PMSVM, RF, PPM, SCAP  & Identif  \\ \hline

\textbf{\cite{Neal18}}  & \textbf{Blog Posts} & En & 1000 & 4 post/pers, 1.6K char/post &  Stylometry, $n$-grams &  Isolation Forest  & Verif  \\ \hline

\textbf{\cite{Belvisi20}}  & \textbf{Tweets}  & En & 40   & 120-200 Tweet/pers &  Stylometry, $n$-grams &  Distance  & Identif  \\ 

Present  &  &  &  93/3957   & 1-3.2K Tweet/pers &   &    &   \\ \hline


\end{tabular}

\end{center}
\caption{Existing studies in online authorship analysis. References are in chronological order. Works marked in bold refer to studies related with short digital texts.
DBN=Deep Belief Networks. DT=Decision Trees. KL=Kullback-Leibler. LR=Logistic Regression. NB=Naive Bayes. NF=Node Frequency. NN=Neural Networks. PM=Probability Models. PMSVM=Power Mean Support Vector Machines. PPM=Prediction by Partial Matching. RF=Random Forest. SCAP=Source-Code Author Profile. SVM=Support Vector Machines.  
Ar=Arabic.
Ch=Chinese Mandarin.
En=English. 
It=Italian.
Th=Thai.
The number indicated in the column 'users' does not necessarily reflect the total number of classes (authors) used in the experiments, since some papers employ $k$-fold cross-validation to report accuracy. 
}
\label{tab:SOA}
\end{table*}

\subsection{Literature Review}


Authorship analysis originates 
from a linguistic research area called stylometry, which refers to statistical analysis of literary style \cite{Williams75}.
Increasingly, the
research focuses on 
online messages
due to the growth of web applications and social networks \cite{Nirkhi2013,Bouanani2014,Stamatatos2009}.
Table \ref{tab:SOA} provides a summary of studies in online authorship analysis.
%
%
They differ in the features and classifier used (Figure~\ref{fig:SOA}).
%
The most widely used features are
stylometric \cite{Stamatatos2009}, 
specially $n$-grams \cite{Peng03}, 
which are a particular type of stylometric features.

Stylometric features aim at capturing writing patterns at different levels: character and word level (lexical), sentence and paragraph (structural), punctuation and function words (syntactic), topic (content), and unique elements of a particular author (idiosyncratic).
\textit{Lexical} features capture the 
characters and words that an individual employs, describing vocabulary richness and preference for particular symbols, words, etc. 
At the character level, features may include the number or frequency of different characters. 
These are the most primitive characteristics of a text.
Word-level features include the total number of words, average word length, fraction of short/long words, most frequent words, number of unique words, etc. 
Another effective set of lexical features are the $n$-grams.
They represent sequences of $n$ elements next to each other. The elements 
can be of different nature, for example a sequence of characters, words, symbols, syllables, etc.
Given a text, the frequency of each sequence of $n$ consecutive elements is computed, resulting in an histogram that characterizes their distribution.
$N$-grams are very tolerant to typos, misspellings, grammatical and punctuation errors, and they are language independent \cite{Peng03}. They can also capture elements such as punctuation or symbols, since they are not restricted to alphabets or numbers. 
\textit{Structural} features capture the organization of paragraphs and sentences. They include the number of sentences, paragraphs, lines, punctuation, average length of sentences or paragraphs, etc. Other elements can be analyzed as well, such as the use of greetings and farewell text, signature, etc.
Both lexical and structural features represent a text without regard for word order, grammar, or context.
\textit{Syntactic} features, on the other hand, characterize the use of punctuation and function words, which help to define the relationship of elements in a sentence. It also includes Part-of-Speech (POS) tagging, which is categorizing a word according to its function (verb, noun, pronoun, adjective, etc.). 
In this context, some authors have also suggested computing $n$-grams, but of POS tags \cite{Okuno14}.
A weakness of this type of features is that POS is language-dependant, and relies on a language parser which unavoidably introduces errors \cite{Stamatatos2009}.
\textit{Content} features analyze, for example, the distribution or frequency of specific keywords. 
However, they are topic-dependant, thus not applicable to the analysis of general texts such as Twitter posts.
%
%
Lastly, \textit{idiosyncratic} features aim at capturing particularities of an author in terms of e.g. misspelled words, abbreviations used, or other special characters (such as emojis).

To handle the available texts per author, two approaches are employed (Figure~\ref{fig:approaches_models}):
profile- and instance-based \cite{Bouanani2014}.
In \textit{profile-based}, all texts from an author are concatenated into a single text, which is used to extract the properties of the author’s style. The representation could include texts of different nature (formal, informal), creating a more comprehensive profile. Also, this can handle data imbalance or lack of enough data.
%
%
This approach is implemented by using probabilistic techniques such as Naive Bayes, or distance measures.
In \textit{instance-based} approaches, on the other hand, 
features from each individual text are extracted separately,
and used to train a model that represents the identity of the author.
This requires multiple training samples per author in order to develop an accurate model, which may not always be available in forensic scenarios.
Instance-based approaches are usually implemented using clustering algorithms or machine learning classifiers such as Support Vector Machines (SVMs), Artificial Neural Networks (ANNs), Decision Trees (DTs), or Random Forest (RF).

Some studies have been conducted specifically using short digital texts
such as Tweets \cite{Okuno14,Brocardo19book,Brocardo14pst,Sultana14icc,Sultana17thms,Sultana18icsmc,[Azarbonyad15tweet_time_evolution],[Rocha17tifsTweets],Belvisi20}, 
SMS \cite{Donais13,Ragel13},
or small pieces 
from blog posts \cite{Neal18}.
They are indicated in bold in Table~\ref{tab:SOA}.
In an early work, the authors of \cite{Okuno14} reported an accuracy of 53.2\% in identifying the authorship among 10000 Twitter users, with the maximum span between Tweets of 1 month. 
As features, they employed POS tags $n$-grams.
%
To handle short texts, 
$n$-grams were weighted based on the length of the elements.
With 120 Tweets available per user, they used 90 for training, and 30 for testing. 
Recently, we employed a set of stylometric features and character and word $n$-grams, with a database of 40 users and 120-200 Tweets per user \cite{Belvisi20}. 
We reported an accuracy of 92-98.5\% (depending on the features) in author identification. 
The number of Tweets per user in such study was in the same range than \cite{Okuno14}, but the number of users was substantially different, which may explain the different accuracy. 
The work \cite{Donais13} reported an accuracy of 20.25\% in the classification of 2000 SMS messages from 81 authors, with maximum 50 messages per author. The time span between messages is not specified.  
Also concerned with authorship of SMS, the work \cite{Ragel13} employed a corpus of 70 persons, with at least 50 SMS per person captured across a year.
%
As features, they employed unigram word counts, which is simply the frequency of each word available in the text.
%
They carried out experiments varying the number of users in the database, and the size of training and test data.
For example, with 20 users having more than 500 SMS each for training, accuracy was $\sim$88\% when the testing set contained 20 SMS per user. If only one SMS is available for testing, accuracy goes down to about 41\%.
They also studied the effect of reducing the training set. With 100 training messages per author, accuracy barely goes above 61\% when 20 SMS per user are combined for testing. With 50 training messages, accuracy goes below 48\%.
The authors acknowledge that these results may be positively biased since they deliberately chose the SMSs having the maximum length, highlighting the difficulty of the problem.
Temporal changes of words usage in Tweets were analyzed in \cite{[Azarbonyad15tweet_time_evolution]}, with character $n$-grams as feature method. Using 133 users and 1820 average Tweets per user posted over a period of 4 years, the authors proposed a time-aware attribution algorithm that builds a model for different time periods and applies a weight decay factor specific of each user. The employed model provided improvements (measured by precision and recall curves) in comparison to non applying time-aware compensation.
Lastly, the work \cite{[Rocha17tifsTweets]}
studied character, word and POS tags $n$-grams, combined with several classifiers. They varied the number of users in the database between 50 and 1000, and the number of training Tweets per user between 1 and 1000.
%
%
With 500 training Tweets, the best accuracy ranged from 57\% (50 users) to 35\% (1000 users). If there are only 50 training Tweets, these numbers were reduced to 46\% and 28\% respectively.
One of the databases of the present paper is the same than \cite{[Rocha17tifsTweets]}, with our experiments reaching up to 1950 users in the database.

Regarding authorship verification, 
Brocardo \emph{et al}. investigated the use of emails and Tweets from 150 and 100 authors respectively \cite{Brocardo19book,Brocardo14pst,Brocardo14aina,Brocardo14cits}.
The database of Tweets of this set of studies is also employed in the present paper, 
with messages posted across the course of several years.
They made use of an extensive set of stylometric features, as well as character and word $n$-grams, with a total dimensionality of 1072. 
They carried out a selection based on correlation measures to keep the most discriminative features.
They also investigated different classifiers .
The best reported accuracy is EER=8.21\% (emails) and 10.08\% (Tweets) \cite{Brocardo19book}.
Recently, the work \cite{Neal18} by Neal \emph{et al}. employed short blocks of 50-100 characters from blog posts to carry out continuous user verification. 
The aim is to simulate for example login sessions where the user is verified as s/he produces new text. 
As features, they used a set of stylometric features and character $n$-grams, with a total dimensionality of 671. 
As classifier, they employed Isolation Forest, trained with 3-10 blocks of text, and tested with the next block (to simulate continuous sessions).
In contrast to regular verification, the classifier is trained to classify test samples as normal or abnormal. Accordingly, all test samples should be classified as normal, without the need of training a classifier on both positive and negative samples.
The accuracy in classifying genuine users exceeded 98\%

In another set of works, Sultana \emph{et al}. \cite{Sultana14icc,Sultana17thms,Sultana18icsmc}
proposed several features aimed at quantifying interaction patterns and temporal behaviour of Twitter users, rather than analyzing the actual content of messages.
Interaction patterns were measured by creating profiles of friendship (other users with whom a user maintains frequent relationships via retweet, reply and mention), contextual information (shared hashtags and URLs), and temporal interaction (posting patterns) \cite{Sultana14icc,Sultana17thms}.
Temporal interaction is further analyzed in \cite{Sultana18icsmc}
where 
temporal profiles of users are created
by measuring features such as average probabilities of tweeting/retweeting/replying/mentioning per day, per hour, or per week. 
They employed a self-collected database of more than 240 users, with Tweets spanning over 4 months. 
They carried out both identification and verification experiments, with a
reported accuracy of 94\% (rank-1 identification) and 4\% (EER) in the best case \cite{Sultana18icsmc}.

\subsection{Contributions}

\begin{figure}[t]
\centering
        \includegraphics[width=0.35\textwidth]{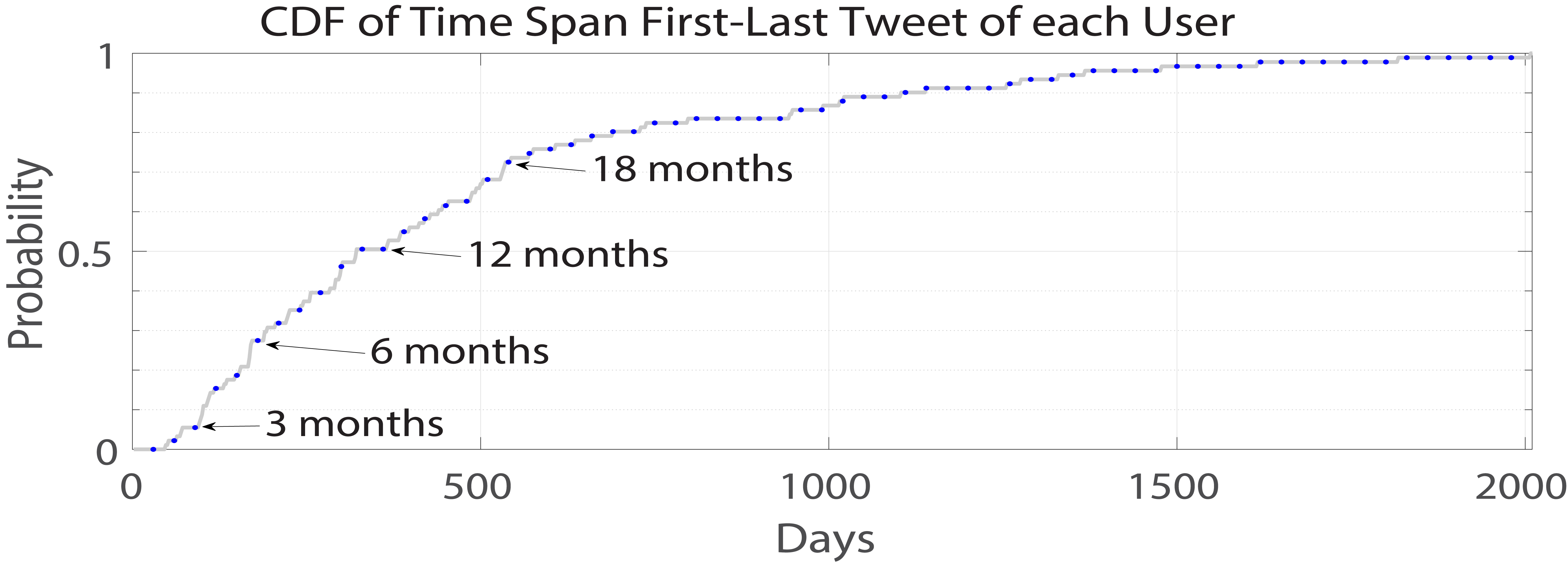}
\caption{CDF of the time span between the first and the last Tweet of each user (ISOT database \cite{Brocardo14pst}). The circular markers are indicative of entire months (30 days).
}
\label{fig:dbstats_time}
\end{figure}

This paper is based on a previous conference study 
\cite{Belvisi20}, where 
we employed 18 stylometric features with 
40 Twitter users and 120-200 Tweets per user. 
%
%
The study is extended substantially here with new developments, data, and experiments.
One of the new databases includes 93 authors, with 1142-3209 Tweets 
per author (242K Tweets in total). 
%
%
This database is employed to do a preliminary analysis of
the influence of the elements that constitute the different
feature types evaluated in this paper.
It is also used to explore feature permanence over time, when Tweets are being posted across the course of several years (Figure~\ref{fig:dbstats_time}).
Once the optimal set of elements of each feature type has been set, we carry out a more comprehensive set of experiments with a second new database of 3957 authors, with 1000-3249 Tweets per author (7,8M Tweets).
The number of features is increased here to 173, plus character $n$-grams at different levels ($n$ = 2 to 6), with a dimensionality of 2000 each (Table~\ref{tab:features}).
We adapt stylometric features from the literature \cite{Stamatatos2009} to the particularities of short digital texts 
\cite{Brocardo19book,Neal18,Okuno14}. 
%
%
Our set of features is also significantly larger than previous studies comparable to ours in contents since these are primarily restricted to character or word $n$-grams \cite{[Rocha17tifsTweets]}.

The feature set is also enhanced with Twitter-specific features 
such as the use of hashtags, URLs, reply mentions and quotes.
%
%
%
%
To the best of our knowledge, only \cite{Sultana14icc,Sultana17thms,Sultana18icsmc} have previously used Twitter-specific features for the task of writer analysis. 
In contrast to them, we are agnostic of the individual content, and we just consolidate the total count of each element.
In our experiments, such features are observed to provide a substantial improvement when added to all type of stylometric features. 
%

While the majority of studies concerned with the analysis of Twitter texts operate in verification mode 
%
%
(Table~\ref{tab:SOA}), this work focuses on author identification as well.
In addition, we carry out an evaluation of the different features. We vary the number of enrollment and test Tweets per author across a wide range to quantify the impact of the amount of available text on the accuracy.
We range from 
1000 Tweets per user to 
100 or 50.
We further reduce the amount to simulate cases more suited to social media forensics, where data might be scarce \cite{[Jain15]}, such as 20, 10 or 5 Tweets only.
Moreover, to test how accuracy of the employed features correlates with the authors set size, we consider different amounts of authors in the database.
Starting from 100, we augment the database to 500, 1000 and 1950 classes.
%
%
%
%
The maximum amount of authors we have used, 1950, is also the highest in comparison to our reviewed studies (Table~\ref{tab:SOA}), with the closest using 1000 authors \cite{[Rocha17tifsTweets]}. Thus, we believe that our results contribute to the study of authorship of short digital messages in being supported by large enrolment and test data combinations produced by a wide range of sizes of author sets.

\begin{figure}[t]
\centering
        \includegraphics[width=0.3\textwidth]{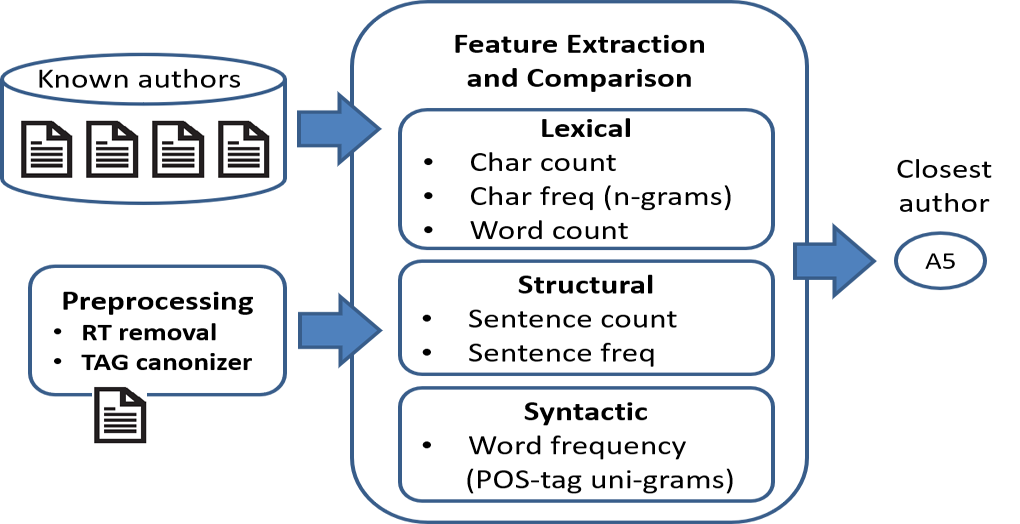}
\caption{Structure of the author identification system.}
\label{fig:system}
\end{figure}

Among the features evaluated, $n$-grams have been observed to provide the best identification performance.
This is in consonance with the literature \cite{Nirkhi2013,Bouanani2014,Stamatatos2009}, and explains the popularity of $n$-grams in authorship studies with all kinds of texts (Table~\ref{tab:SOA}).
Interestingly, this superiority is not translated to verification experiments, being surpassed by other features such as uni-grams or sentence-level features. Indeed, sentence-level features are the best ones for author verification, while they rank third in identification tests.
%
%
We have studied here, both character and word $n$-grams.
$N$-grams have the advantage that they can cope with different text lengths, as well as misspellings and other errors. 
In particular, we evaluate character $n$-grams up to $n$=6.
For $n$=1 (individual characters), we also analyze the influence of different elements, including alphabets (both case-dependent and independent), digits, vowels, spaces, special characters, and particular elements of Tweets (hashtags, link, mentions, and quotations). 
To the best of our knowledge, this is the first work, together with \cite{[Rocha17tifsTweets]}, conducting an exhaustive study of $n$-grams applied to 
online short messages.
Regarding word $n$-grams, we study Part-of-Speech (POS) uni-grams \cite{Okuno14}, which converts the word to its POS tag (noun, verb, etc.), and then computes the frequency of each tag in the text.
This results in a substantially smaller feature space than if we had to compute the frequencies of a given dictionary of individual 
words. 

\begin{table}[t]
\begin{center}
\begin{tabular}{|c|c|c|l|c|}

\multicolumn{5}{c}{} \\ \hline
 
\multicolumn{3}{|c|}{\textbf{Category}}  & \textbf{Feature}  &  \textbf{Total}   \\ \hline


 & \multirow{16}{*}{\rotatebox{90}{Character}}  &  & Alphabets & 1 \\
 
 &  &  & Uppercase alphabets & 1 \\
 
 &  &  & Lowercase alphabets & 1 \\
 
 &  & Count & Vowels & 1 \\
 
 &  &  & Uppercase vowels & 1 \\
 
 &  &  & Lowercase vowels & 1 \\
 
 &  &  & Digits & 1 \\
 
 &  &  & White spaces & 1 \\
 
 &  &  & Special characters & 1 \\ 
 
 &  &  &  \multicolumn{2}{r|}{\textbf{9 features}}   \\ \cline{3-5}
 
 &  &  & Alphabets & 26  \\
 
  &  &  & Uppercase alphabets & 26 \\
 
 &  & Freq & Lowercase alphabets & 26 \\
 
 
 
 
Lexical &  & $n$-grams & Digits & 10 \\

&  &  & White spaces & 1 \\
 
 &  &  & Special characters & 33 \\ 
 
 &  &  & \multicolumn{2}{r|}{\textbf{122 features}} \\ \cline{4-5}
 
 &  &  & Char $n$-grams ($n>$1) & 2000 \\  
 
 &  &  & \multicolumn{2}{r|}{\textbf{5 $\times$ 2000  features}} \\  \cline{2-5}

 & \multirow{14}{*}{\rotatebox{90}{Word}}  &  & URLTAG & 1 \\ 
 
 &   &  & USERTAG (mention) & 1 \\ 
 
 &   &  & TRENDTAG (hashtag) & 1 \\ 
 
 &   &  & QUOTETAG & 1 \\ 
 
 &   &  & Words (W) & 1 \\ 
 
 &   &  & Average word length (L) & 1 \\ 
 
 &   & Count & Ratio W/L & 1 \\ 
 
 &   &  & Fraction of short words & 1 \\ 
 
 &   &  & Fraction of long words & 1 \\ 
 
 &   &  & Unique words & 1 \\ 
 
 &   &  & Capitalized words & 1 \\ 
 
 &   &  & Uppercase words & 1 \\ 
 
 &   &  & Lowercase words & 1 \\ 
 
 &   &  & Othercase words & 1 \\  
 
 &   &  & \multicolumn{2}{r|}{\textbf{14 features}} \\  \hline

 & \multirow{7}{*}{\rotatebox{90}{Sentence}}  &  & Sentences & 1 \\ 
 
 &   &  & Punctuation symbols & 1 \\ 
 
Struc- &   & Count & Words per sentence & 1 \\ 
 
 tural &   &  & Characters per sentence & 1 \\ 
 
 &   &  & Uppercase sentence & 1 \\ 
 
 &   &  & Lowercase sentence & 1 \\ 
 
  &   &  & \multicolumn{2}{r|}{\textbf{6 features}} \\  \cline{3-5} 
 
 &   & Freq & Punctuation symbols & 9 \\ 
 
 &   &  & \multicolumn{2}{r|}{\textbf{9 features}} \\  \hline 

 Syntactic &  Word & Freq  & POS tag uni-grams & 13 \\ 
 &   &  & \multicolumn{2}{r|}{\textbf{13 features}} \\  \hline 



\end{tabular}

\end{center}
\caption{List of properties of Twitter messages measured in this work.}
\label{tab:features}
\end{table}
\normalsize

At the database level, we have observed that features that analyze word usage (word-level features and POS tag uni-grams) behave differently on each database. 
%
%
One of the databases contains messages from influential tweeters, while the other database is from the general public.
%
%
Influential tweeters can be expected to 
have a specific writing style in order to build a particular brand, which is reflected in a different word usage. 
%
%
%
Interestingly, the accuracy of word-related features is better with the database of general public. 
One may reason that 
connected users tend to use a common vocabulary (due to shared interests) that may hinder their identification when analyzing language usage. The influential tweeters are all from the UK, although we have not analyzed their connections, while the other database is from random authors, so they are unconnected in principle.
The effect is exacerbated when there are few training or test Tweets, when it is expected that the amount of employed words that differ from a 'common' vocabulary is less. Consequently, connected authors become closer to each other in the feature space.
%
This effect is also observed with $n$-grams, since they are computed using a common dictionary for the two databases, although it does not happen until we employ very few Tweets for training and test. In this sense, $n$-grams are revealed again as a very robust feature, at least if sufficient data is available.
%
%
This comparative analysis between two databases of the same type of texts with two different writing styles can be also considered novel, since previous works mainly use one database only.

We also study feature combination 
by automatic feature selection. Using Sequential Forward Floating Selection (SFFS) \cite{[Pudil94]}, we find the optimum set of elements that maximize the identification accuracy.
The search space consists of 672 elements, and includes the different levels of stylometric analysis evaluated (Table~\ref{tab:features}).
Feature importance is analyzed, having observed that combining different features leads to better results, even if the individual performance of some of them is comparatively worse.
With appropriate selection mechanisms, the best individual stylometric feature ($n$-grams) can be surpassed with a reduced feature set of only 15 elements. The peak in performance is observed with 239 elements ($\sim$35\% of the search space).
Another novelty is the analysis of permanence of features over time. 
%
In a related study \cite{[Azarbonyad15tweet_time_evolution]} which employed only character $n$-grams ($n$=4), the authors studied the evolution of writing style over time.
In comparison, we contribute with the evaluation of our extended set, quantifying the differences in robustness between the different types of features.
Lastly, we analyze time complexity of the employed features. The entire set of features takes 89 ms per Tweet to be extracted on a regular laptop, and vector comparisons are done in microseconds due to the use of distance measures (involving a fixed number of computations). This would scale well with bigger databases. For example, with 1M users, a query against the entire database would take a few seconds, which could be speeded up further by using dedicated hardware. 

\begin{figure}[t]
\centering
        \includegraphics[width=0.35\textwidth]{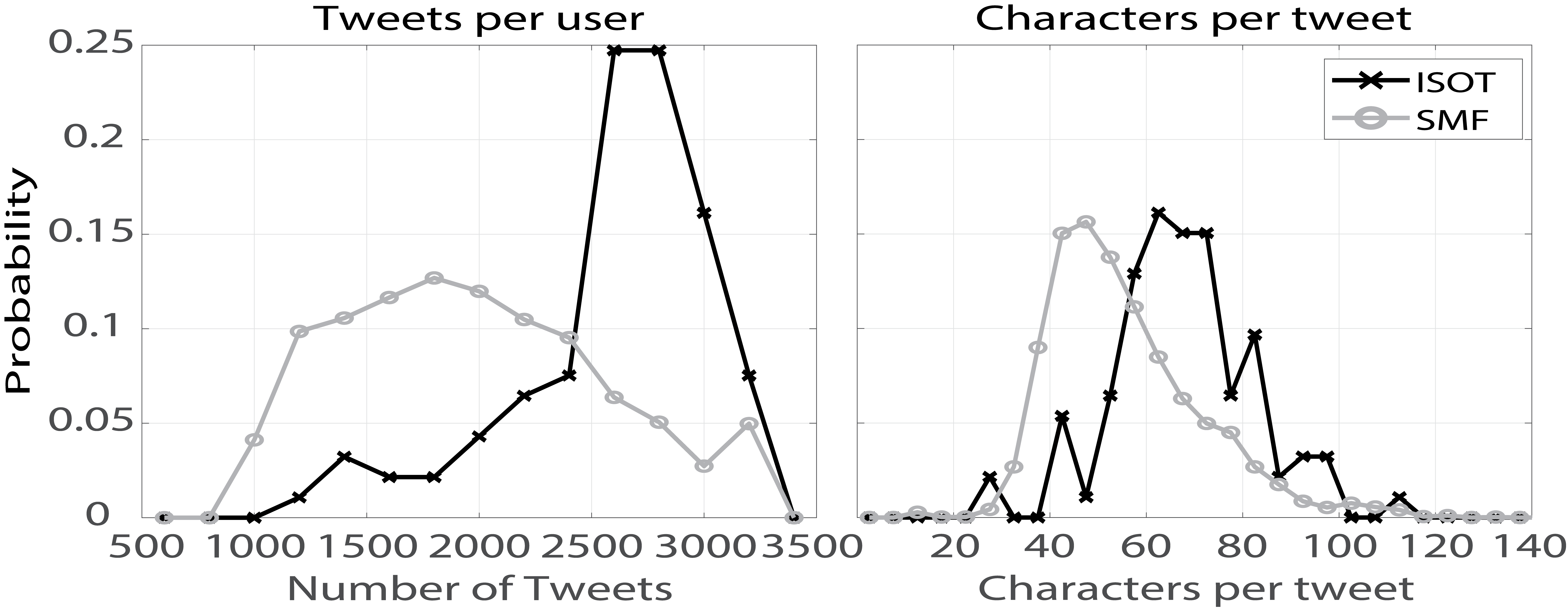}
\caption{Statistics of the ISOT \cite{Brocardo14pst} and SMF \cite{[Rocha17tifsTweets]} databases.}
\label{fig:dbstats}
\end{figure}

\begin{figure}[t]
\centering
        \includegraphics[width=0.35\textwidth]{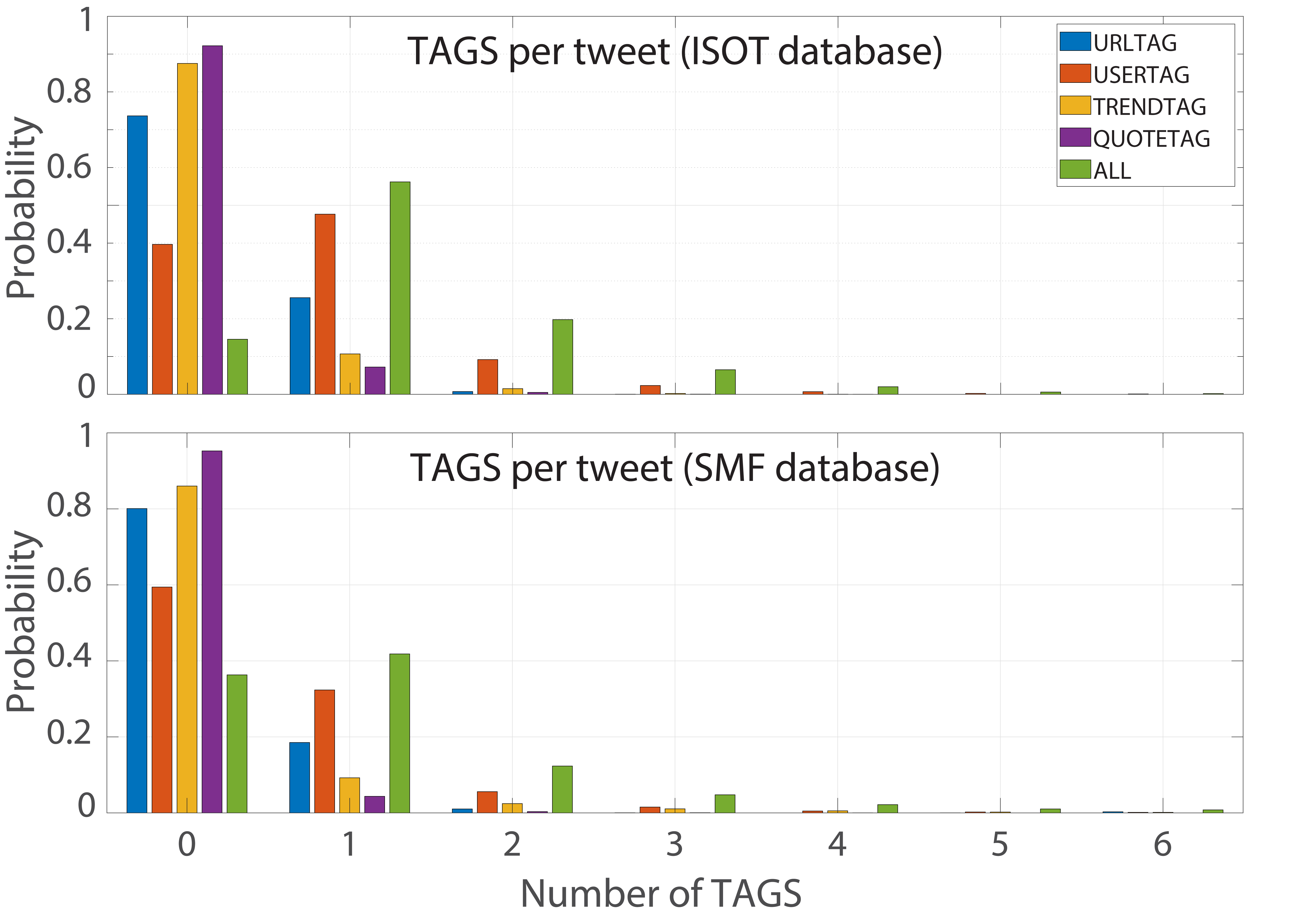}
\caption{Histogram of number of tags per Tweet.
}
\label{fig:dbstats_tags}
\end{figure}

\section{Authorship Identification Using Microblogging Texts}
\label{sect:features}

Asserting writer identity based on Twitter messages require four main phases: 1) pre-processing, 2) feature extraction, 3) feature comparison, 4) and writer identification.
In Figure~\ref{fig:system}, the overall model of our author identification system is depicted.
In this section, we present the pre-processing steps and feature extraction methods.
We measure properties 
extracted from Twitter messages to characterize writer identity in a manner that is independent of the amount of text.
This allows to accommodate messages of different length, and to vary the number of Tweets that models writer's identity.
Thus, with `feature vector' we denote a vector 
capturing several properties of the Twitter messages generated by an author.
At the same type, this allows us to employ simple distance measures to compute the similarity between two given feature vectors.
The particular properties chosen for this work are given in Table~\ref{tab:features}.
We have selected a set of lexical, structural, and syntactic stylometric features suitable for Twitter messages. For example, given the limited length (140-280 characters), we do not consider structural features related to paragraph properties, since rarely there is more than one paragraph. 
We do not consider content or idiosyncratic features either, since these are dependant on particular topics, or rely on the use of a language parser. 
The only feature considered that has such need is POS tag uni-grams  \cite{Okuno14}. This is because character and word $n$-grams have shown high efficiency in the literature with short digital texts \cite{Ragel13,Brocardo14pst,Brocardo19book,Brocardo14pst,Neal18,Belvisi20}, so we are interested in evaluating their efficiency using syntactic elements as well.

\subsection{Pre-processing}
\label{sect:prepro}

Elements which are not reliable to model identity are removed, either because they are produced by another writer, or because they are used by many.
This includes:

\begin{itemize}
  \item Replacing URLs with the meta tag `URLTAG'.
  
  \item Removing re-tweeted text. A re-tweet is a re-posting of a Tweet written by another user. This can be identified by the letters `RT' followed by the username of the person that wrote the original message. 
  We only keep the text before the RT flag. If there is no other text, then we delete the entire Tweet.
  
  \item Replacing usernames (`@username') with the meta tag `USERTAG'.
  A username can only contain alphanumeric characters and underscores, and cannot be longer than 15 characters, facilitating its detection. When usernames of other users are included in Tweets, they receive a notification, allowing to track mentions by others. 
  
  \item Replacing hashtags (`$\#$word') with the meta tag `TRENDTAG'. Hashtags are used to highlight keywords that categorize the topic of the text. They allow to follow threads related to a particular topic, to search for Tweets with a particular hashtag, and they help for relevant Tweets to appear in Twitter searches as well. When many people use the same hashtag, it is used by the Twitter algorithm to determine popular trends ('trending topics') in a particular location or at a particular time.
  
  \item Replacing information among quotes (`word') with the meta tag `QUOTETAG'. 
  This is because the quoted text is likely to be generated by somebody else, thus not reflecting the writing style of the user that quoted the information.
  
\end{itemize}

\begin{table}[t]
\begin{center}
\begin{tabular}{|c|c|c|c|c|c|c|c|}

\multicolumn{8}{c}{} \\  \cline{1-2} \cline{4-5} \cline{7-8}
 
\textbf{$n$=2} & \textbf{Count} &  & \textbf{$n$=3} & \textbf{Count}  & & \textbf{$n$=4} & \textbf{Count} \\ \cline{1-2} \cline{4-5} \cline{7-8}

'th' & 10176 &  & 'the' & 5166  & & 'that' & 964 \\
'in' & 9521  & & 'ing' & 4299  & & 'tion' & 943 \\
'he' & 7177  & & 'you' & 2163  & & 'ther' & 822 \\
'er' & 6897  & & 'tha' & 2041  & & 'than' & 811 \\
'on' & 6522  & & 'and' & 1819  & & 'with' & 773 \\
're' & 6521  & & 'for' & 1673  & & 'ight' & 724 \\
'an' & 6444  & & 'hat' & 1479  & & 'this' & 698 \\
'st' & 5591  & & 'thi' & 1354  & & 'here' & 694 \\
'to' & 5340  & & 'rea' & 1347  & & 'ingt' & 675 \\
'at' & 5228  & & 'her' & 1322  & & 'nthe' & 640 \\ \cline{1-2} \cline{4-5} \cline{7-8}

\end{tabular}

\end{center}
\caption{Ten most frequent $n$-grams of the ISOT database \cite{Brocardo14pst}.}
\label{tab:frequent_ngrams}
\end{table}
\normalsize

\subsection{Lexical Features}

These features describe the set of characters and words that an individual uses. 
At the character level, we extract two types of features: total count of elements (alphabets, numbers, vowels, digits, white spaces, and special characters), and frequency of each individual element (character $n$-grams).
Alphabets and vowels are considered both case-independent (e.g. `a' and `A' count towards the same character), and case dependent (uppercase and lowercase elements are counted separately).
As special characters, we consider the ASCII elements 33-47, 58-64, 91-96, and 123-126, plus the symbol £.
For $n>$1, we compute the frequency of the 2000 most frequent character $n$-grams (up to $n$=6), which have been obtained from a training set (Section~\ref{sect:db_protocol}). 
Regarding word-level features, we calculate the number of words per Tweet, the average word length, the fraction of short and long words ($<=$3 characters and $>$6 characters respectively) \cite{Brocardo19book}, the number of unique words, and different counts related to word casing \cite{Neal18}.
We also employ a number of features which are particular of Tweet messages: number of URLs, user mentions, hashtags, and quotations per Tweet.
Despite being very specific of Tweet messages, such features have not been widely used in related works studying authorship of Twitter messages \cite{Sultana14icc,Sultana17thms,Sultana18icsmc}.

\subsection{Structural Features}

These features inform about the organization of elements in a text, both at sentence- and paragraph-level.
Here, given the short length of Tweets, we only consider sentence features.
In particular, we calculate the number of sentences per Tweet, the number of punctuation symbols, the words and characters per sentence, and the number of sentences commencing with uppercase or lowercase. 
We also calculate the frequency of different punctuation and other sentence separation symbols (, - . : ; $<$ $>$ ? !), which could be considered a type of `punctuation' uni-gram.

\subsection{Syntactic Features}

Finally, we compute the frequency of thirteen syntactic elements found in each Tweet \cite{Okuno14}. For this purpose, each word is converted to its Part-of-Speech (POS) tag \cite{Unicode,ICUtokenizer} and then, the frequency of each element is counted. 
As POS tags, we employ adjective, preposition/conjunction, adverb, verb, determiner, interjection, noun, numeral, particle, pronoun, punctuation, symbol, and `other'.

\begin{figure}[t]
\centering
        \includegraphics[width=0.4\textwidth]{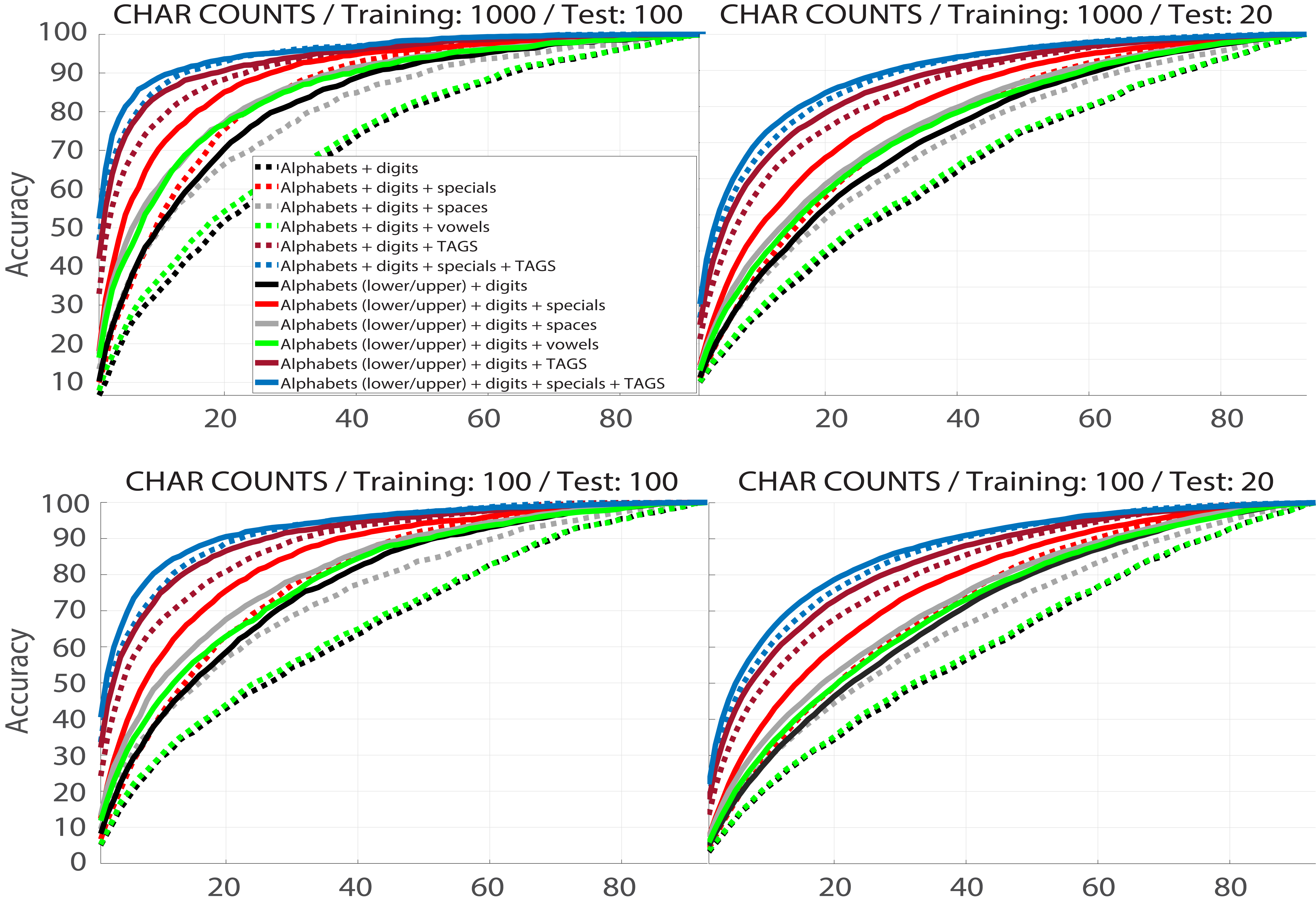}
\caption{Identification accuracy of character count features for a different number of training and test Tweets ($\chi^2$ distance). Results are given for the ISOT \cite{Brocardo14pst} database with 93 test authors. Best in colour.}
\label{fig:results_char_counts}
\end{figure}

\begin{table}[htb]
\begin{center}
\begin{tabular}{|>{\centering}m{14pt}|>{\centering}m{11pt}|||>{\centering}m{9pt}|>{\centering}m{9pt}||>{\centering}m{9pt}|>{\centering}m{9pt}||>{\centering}m{9pt}|>{\centering}m{9pt}|||>{\centering}m{13pt}|>{\centering\arraybackslash}m{13pt}|}

\multicolumn{2}{c}{} & \multicolumn{8}{c}{\textbf{Character Count Features}} \\ \cline{3-10}

\multicolumn{2}{c|}{} & \multicolumn{6}{c|||}{\textbf{ISOT DB}} & \multicolumn{2}{c|}{\textbf{SMF DB}} \\ \cline{3-10}

\multicolumn{2}{c|}{} & \multicolumn{2}{c||}{$\chi^2$} & \multicolumn{2}{c||}{Cosine} & \multicolumn{2}{c|||}{Euclidean} & \multicolumn{2}{c|}{$\chi^2$} \\ \hline
 
\textbf{Train} & \textbf{Test} & \textbf{R1} & \textbf{R5} & \textbf{R1} & \textbf{R5} & \textbf{R1} & \textbf{R5}  & \textbf{R1} & \textbf{R5} \\ \hline

1000 & 1000 & \textbf{79.8} & \textbf{97}  & 65.6 & 90.9 & 31.3 & 45.5  &  -0.1 & -5.4
  \\ \cline{2-10}
 & 500 & \textbf{73.8} & \textbf{91.8}  & 54.3 & 81.6 & 22.7 & 41.8  &  -0.3 & -3.2
    \\ \cline{2-10}
 & 100 & \textbf{52.4} & \textbf{81.3}  & 34.3 & 64.3 & 14.6 & 29.0   &  +4.9 & -2.9
  \\ \cline{2-10}
 & 50 & \textbf{40.3} & \textbf{72.1}  & 26.1 & 54.2 & 11.8 & 25.7   &  +8.7 & -0.3
  \\ \cline{2-10}
 & 20 & \textbf{25.3} & \textbf{55.9}  & 16.0 & 39.9 & 8.1 & 20.7   &  +13.1 & +5.4
  \\ \hline
 
500 & 500 & \textbf{72.8} & \textbf{90.3}  & 50.1 & 79.4 & 15.8 & 34.7   &  -6.1 & -7.3
 \\ \cline{2-10}
 & 100 & \textbf{49.7} & \textbf{79.5}  & 33.3 & 62.6 & 12.6 & 28.3    &  +2.1 & -5.8
  \\ \cline{2-10}
 & 50 & \textbf{39.4} & \textbf{70.7}  & 25.8 & 52.0 & 10.6 & 24.8   &  +4.9 & -3.3
  \\ \cline{2-10}
 & 20 & \textbf{24.8} & \textbf{55.9}  & 16.4 & 39.8 & 7.9 &  19.9   &  +10 & +1.7
 \\ \hline
 
100 & 100 & \textbf{40.5} & \textbf{69.5}  & 25.2 & 50.1 & 9.8 &  26.0   &  +1.1 & -5.5
 \\ \cline{2-10}
 & 50 & \textbf{32.6} & \textbf{62.2}  & 20.4 & 44.9 & 9.8 &  23.0   &  +3.6 & -2.7
 \\ \cline{2-10}
 & 20 & \textbf{22} & \textbf{49.8}  & 13.9 & 35.5 & 7.3 & 19.5   &  +7.2 & +1.9
  \\ \hline

\end{tabular}

\end{center}
\caption{Identification accuracy (Rank-1, Rank-5) of character count features for a different number of training/test Tweets and different distance measures. The features employed are: Alphabets (separated lowercase/uppercase) + digits + special characters + TAGS (solid blue curves in Figure~\ref{fig:results_char_counts}). Results are 
with 93 test authors. The best case of each row with the ISOT database is marked in bold. Results with the SMF database indicate accuracy change w.r.t. the ISOT database.}
\label{tab:results_char_counts}
\end{table}
\normalsize

\section{Databases and Protocol}
\label{sect:db_protocol}

We use 
the ISOT Twitter database 
\cite{Brocardo14pst} and the Social Media Forensics (SMF) database \cite{[Rocha17tifsTweets]}, with 100 and 10K authors respectively.
%
%
%
%
The distribution of the ISOT database consists of a list of users and corresponding Tweet IDs, so researchers can crawl them using the Twitter API. 
The ISOT database has tweets from 100 names randomly selected from the 
2011 and 2012 lists of the UK’s most influential tweeters compiled by Ian Burrell (The Independent newspaper).
%
%
We were able to crawl 93 accounts, with the Tweets per user ranging from 1580 to 3239.
Some accounts or Tweet IDs outputted an error, either due to issues with the Twitter API, or because they had been deleted. 
Regarding the SMF database, it consists of English speaking public accounts, selected by searching Twitter for language function words. A processed version without link to the actual user or Tweet ID is provided by its authors under request. 
We gathered the data of the SMF database and to ensure a sufficient amount of data per user, we discarded the authors with less than 1000 Tweets. As a result, the dataset obtained consists of 3957 users. 

Figure~\ref{fig:dbstats} shows the statistics of the databases after the pre-processing of Section~\ref{sect:prepro}.
The ISOT database contains 241,985 Tweets, with 1142-3209 Tweets per user (2692 on average).
The average number of characters per Tweet/per user ranges from 26 to 111 (68 on average). 
Regarding SMF, it contains 7,801,919 Tweets, with 1000-3249 Tweets per user (1972 on average).
The average number of characters per Tweet/per user ranges from 13 to 131 (56 on average).
The character counts do not include meta tags, which explains why the average number of characters per Tweet of the databases is 
40-50\% of the 140 characters allowance.
By looking at the histograms of Figure~\ref{fig:dbstats}, it is evident that ISOT has more Tweets per user, and more characters per Tweet. One possible explanation could be that authors of the ISOT database are influential tweeters, who, in principle, can be expected to be more active in social media.

%
%
%
%
%
%
%
%

\begin{figure*}[t]
\centering
        \includegraphics[width=0.90\textwidth]{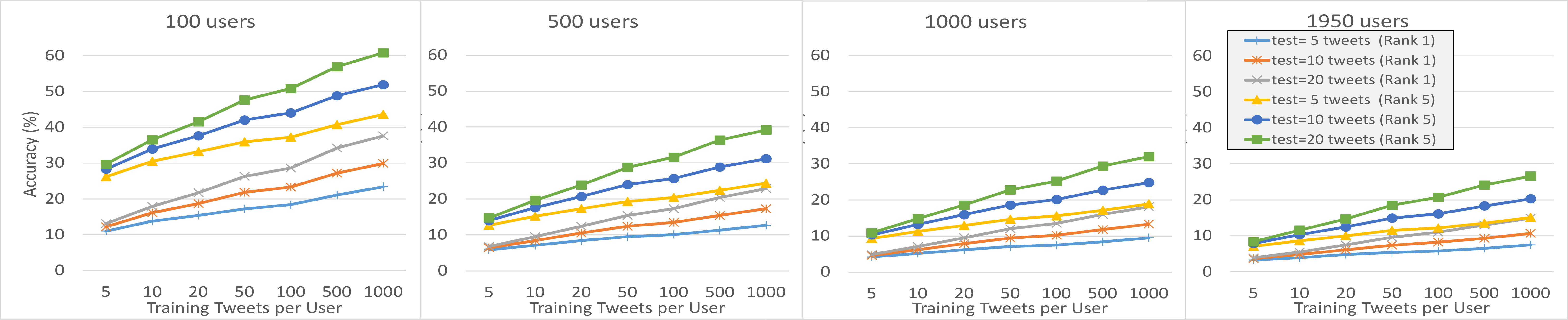}
\caption{Identification accuracy (Rank-1 and Rank-5) of character count features for a reduced number of test Tweets ($\chi^2$ distance). Results are given for the SMF \cite{[Rocha17tifsTweets]} database. Each figure corresponds to a different number of test authors (from left to right: 100, 500, 1000, and 1950 authors). Best in colour.}
\label{fig:results_char_counts_rocha_ranks}
\end{figure*}

\begin{figure}[t]
\centering
        \includegraphics[width=0.4\textwidth]{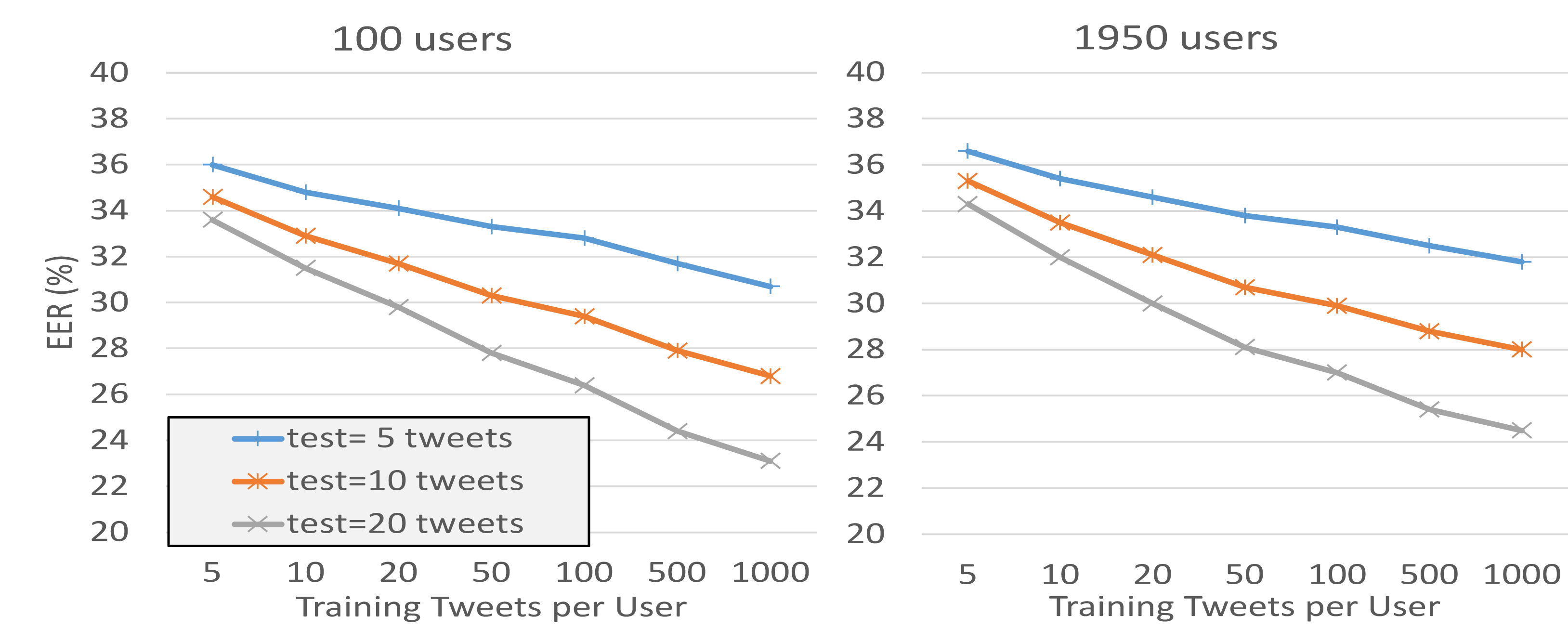}
\caption{Verification accuracy (EER) of character count features for a reduced number of test Tweets ($\chi^2$ distance). Results are given for the SMF \cite{[Rocha17tifsTweets]} database. Each figure corresponds to a different number of test authors. Best in colour.}
\label{fig:results_char_counts_rocha_eer}
\end{figure}

Regarding meta tags,
Figure~\ref{fig:dbstats_tags}
shows their actual distribution per Tweet. 
%
%
%
%
%
The majority of Tweets do not contain all tags at the same time, evidenced by the high probability of some bars at `Number of TAGS'=0. More than 70\% of the Tweets do not contain URLTAG (blue bar), and more than 85\% do not contain TRENDTAG (yellow) or QUOTETAG (purple).
However, only 15\% of the Tweets of the ISOT database do not contain any tag (green bar), and 35\% of the Tweets of the SMF database. 
Around 56\% of the Tweets of the ISOT database contain at least one tag, and 29\% contain more than one. 
With the SMF database, these number are a little bit smaller: around 42\% of the Tweets contain at least one tag, and 20\% contain more than one.
This highlights again the 'celebrity' nature of the ISOT authors, who may be more prone to use these particular Twitter elements than the general public in order to increase their reach.
The most used tag in both databases is USERTAG (mentions to other users, red bar), followed by URLTAG (inclusion of URLs in the Tweet, blue bar).
About 48\% of the Tweets of the ISOT database contain at least one USERTAG, and 13\% contain more than one.
Regarding URLTAG, 26\% of the Tweets contain at least one.
Again, these numbers are smaller with the SMF database, but still they are representative: 35\% of the Tweets contain at least one USERTAG, and 19\% of the Tweets contain at least one URLTAG. 
This highlight the potential usefulness of these elements for recognition tasks.

We also plot in Figure~\ref{fig:dbstats_time} the CDF of the time span between the first and the last Tweet of each user of the ISOT database (such information is not available for the SMF database).
The minimum and maximum spans are 47 and 2007 days, respectively.
As it can be seen, around 25\% of the users have a span of 6 months or less, while another 25\% have 6-12 months.
Half of the users have a span of more than 12 months, with $\sim$30\% having more than 18 months, and $\sim$15\% more than 33 months (990 days).

Pre-processed Tweets are then analyzed, extracting the features of Table~\ref{tab:features} for each Tweet. 
The meta tags introduced in Section~\ref{sect:prepro} are only used to account for their number, according to the four lexical features URLTAG, USERTAG, TRENDTAG, and QUOTETAG count.
They are not used for any other feature computation, e.g. they do not count towards the number of characters per Tweet, frequency of characters, number of words, etc. 
A \textit{profile-based} approach (Figure~\ref{fig:approaches_models}) has been used in this work. 
The first $J$ Tweets of each user are employed for enrolment,
which are averaged to generate a single enrolment vector. 
The remaining Tweets 
are divided into groups of $K$ non-overlapping consecutive Tweets, which are then averaged and used to simulate test identification attempts. 
Enrolment and test vectors are compared by distance or similarity measures. 
We have observed 
that the $\chi^2$ distance is a good choice, in line with previous works \cite{[Hernandez18],[Bulacu07]}, but for comparative purposes, we also give results with
%
the Euclidean distance and the cosine similarity. 
%
%
%
%
%
%

Given a test sample, identification experiments are done by comparing it against every enrolment sample, producing a sorted list of identities ordered from the best match (having the smallest distance score) to the worst match (highest score). Results are reported in the form of Cumulative Match Characteristic (CMC) curve, and Rank-$N$ measures (probability that the correct identity is within the first $N$ matches).
%
%
For completeness \cite{[DeCannRoss13btasROCCMCmenagerie]}, we also report author verification results. For this purpose, the scores obtained during the identification experiments are rearranged into \emph{mated} and \emph{non-mated} scores, from which the Equal Error Rate (EER) and Detection Error Trade-off (DET) graph are computed.
Mated scores refer to scores generated when comparing two vectors from the same individual, whereas non-mated scores are when the vectors belong to different individuals.

We use different combinations of $J$ and $K$, with 
$J, K \in \{5, 10, 20, 50, 100, 500, 1000\}$. 
%
The aim is to test a different number of enrolment and test Tweets, from a very large amount (hundreds to thousands) to just a few Tweets. The latter is tailored to social media forensic scenarios, as in related works \cite{[Rocha17tifsTweets]}, where the amount of available data may be scarce.
%
%
The ISOT database (93 authors) is used first to evaluate different combinations of each feature type, after which the best combinations are carried forward for further experiments with the larger SMF database.
We also carry out experiments varying the number $M$ of authors of the SMF database ($M \in \{100, 500, 1000, 1950\}$) to evaluate how the features handle an increasing number of classes. 
Experiments with the SMF database are done by $k$-fold cross-validation, repeating each test $k$ times with the authors of each fold chosen randomly (without overlap between folds), and reporting the average accuracy.
The value of $k$ is
$k$=5 ($M$=100, 500),
$k$=3 ($M$=1000), and
$k$=2 ($M$=1950),
chosen to maximize the number of possible folds per $M$. 
%
%
To compute the dictionaries of the 2000 most frequent character $n$-grams ($n>$1), we use the first 100 Tweets of each user of the ISOT database (9300 Tweets in total). %
We convert all text to lowercase and remove white spaces before computing the $n$-grams.
%
%
Table~\ref{tab:frequent_ngrams} shows the ten most frequent $n$-grams for $n$ = 2, 3, 4.
Regarding the sofware employed, we use Matlab r2019b. 
The POS tag information is obtained with the functions \textit{tokenizedDocument}, \textit{addPartOfSpeechDetails} and \textit{tokenDetails}.
The text is tokenized using rules based on Unicode Standard Annex 29 \cite{Unicode} and the ICU tokenizer \cite{ICUtokenizer}.

%
%
%
%
%
%
%
%

\section{Experimental Results}
\label{sect:results}

%
%
We are interested in a comparative analysis
of the features for a different number 
of enrolment/test Tweets, and for a different number of authors in the database.
We are
also interested in the improvements 
obtained
by combining multiple features. 
Thus, we consider first the
individual features and then their combinations.
The time complexity and feature permanence over time is also analyzed.

\subsection{Character Counts}
\label{sect:results:char_count}

These refer to the Lexical-Character-Count features of Table~\ref{tab:features}.
Identification results with the ISOT database are given in Figure~\ref{fig:results_char_counts} and Table~\ref{tab:results_char_counts} (left).
We have tested different combinations involving count of alphabets, digits, spaces, vowels, and special characters.
We also distinguish the case where uppercase and lowercase alphabets are counted separately (solid curves, indicated as `lower/upper' in the graphs) or together (dashed curves). 
We have also added to the feature vector the Lexical-Word-Count meta tags that compute the number of URLs, user mentions, hashtags, and quotations per Tweet.
These tags are treated as separate entities in the feature extraction process, as mentioned in Section~\ref{sect:db_protocol}, so they do not count towards the Lexical-Character-Count features. 
Since they are very particular of Tweet messages, we are interested in testing if they can complement other features.

\begin{figure}[t]
\centering
        \includegraphics[width=0.4\textwidth]{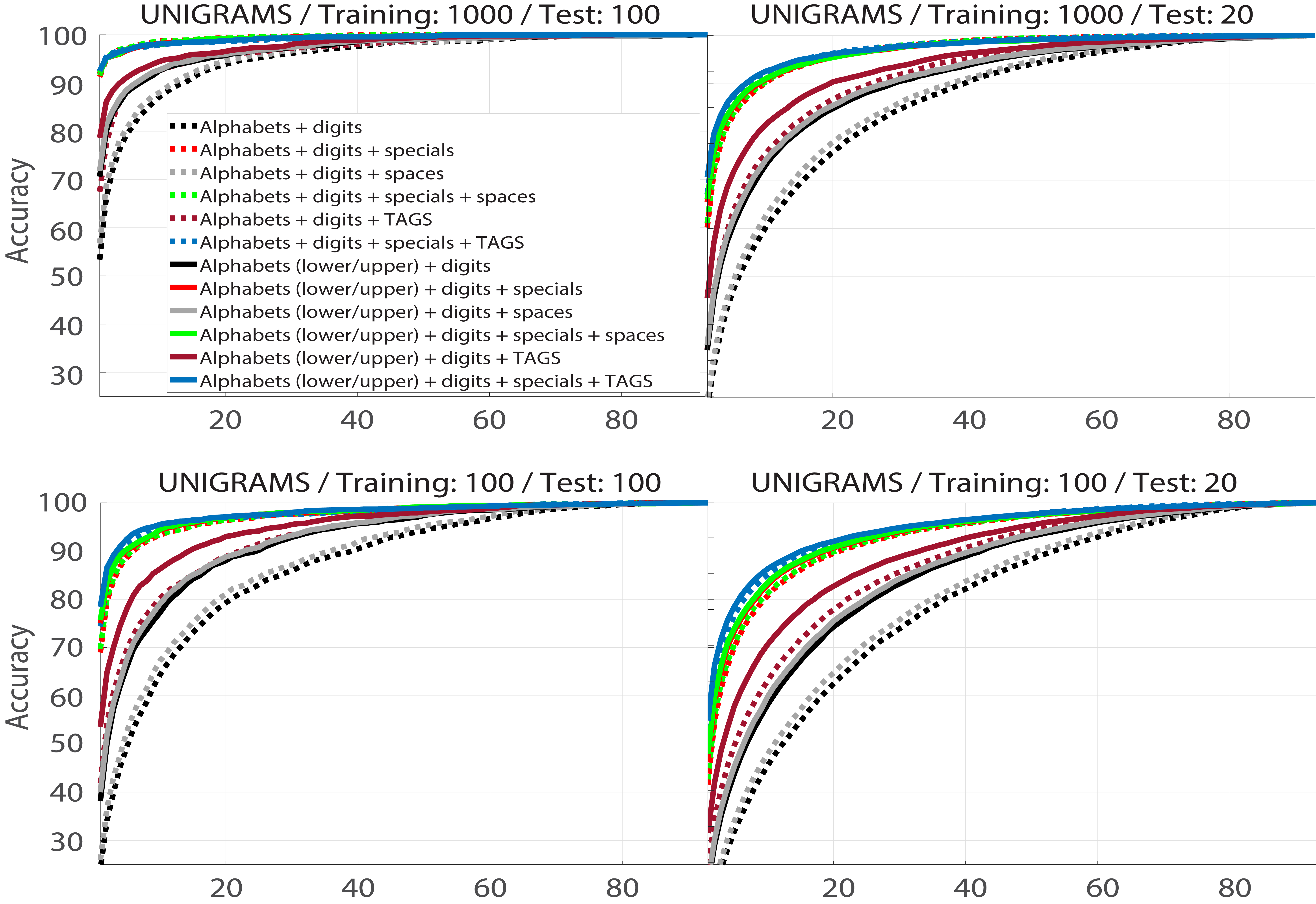}
\caption{Identification accuracy of character frequency features (uni-grams) for different training and test Tweets ($\chi^2$ distance). Results are given for the ISOT \cite{Brocardo14pst} database with 93 test authors. Best in colour.}
\label{fig:results_char_unigrams}
\end{figure}

\begin{table}[t]
\begin{center}
\begin{tabular}{|>{\centering}m{14pt}|>{\centering}m{11pt}|||>{\centering}m{9pt}|>{\centering}m{9pt}||>{\centering}m{9pt}|>{\centering}m{9pt}||>{\centering}m{9pt}|>{\centering}m{9pt}|||>{\centering}m{13pt}|>{\centering\arraybackslash}m{13pt}|}

\multicolumn{2}{c}{} & \multicolumn{8}{c}{\textbf{Character Frequency (uni-grams)}} \\ \cline{3-10}

\multicolumn{2}{c|}{} & \multicolumn{6}{c|||}{\textbf{ISOT DB}} & \multicolumn{2}{c|}{\textbf{SMF DB}} \\ \cline{3-10}

\multicolumn{2}{c|}{} & \multicolumn{2}{c||}{$\chi^2$} & \multicolumn{2}{c||}{Cosine} & \multicolumn{2}{c|||}{Euclidean} & \multicolumn{2}{c|}{$\chi^2$} \\ \hline
 
\textbf{Train} & \textbf{Test} & \textbf{R1} & \textbf{R5} & \textbf{R1} & \textbf{R5} & \textbf{R1} & \textbf{R5}  & \textbf{R1} & \textbf{R5} \\ \hline

1000 & 1000 & \textbf{99} & \textbf{99} & 97.0 & 98.0 & 70.0 & 80.0 & -2 & +0.1
  \\ \cline{2-10}
 & 500 & \textbf{96.9} & \textbf{98.4} & 93.8 & 97.3 & 64.5 & 76.2 &  -2.2 & -0.9
 \\ \cline{2-10}
 & 100 & \textbf{92.5} & \textbf{96.8}  & 81.1 & 92.7 & 45.8 & 61.3 & -2.5 & -1.3
 \\ \cline{2-10}
 & 50 & \textbf{86.6} & \textbf{95}  & 66.9 & 86.5 & 36.7 &  52.9 & -1.4 & -1
 \\ \cline{2-10}
 & 20 & \textbf{70.6} & \textbf{87.8}  & 45.0 & 71.3 & 25.7 & 42.4 &  +3.5 & +1.3
 \\ \hline
 
500 & 500 & \textbf{96.6} & \textbf{98.9}  & 92.0 & 96.0 & 56.7 &  72.2 & -3.8 & -2.2
 \\ \cline{2-10}
 & 100 & \textbf{91.3} & \textbf{96.8}  & 75.6 & 90.7 & 44.9 & 59.4 & -5 & -2.8
 \\ \cline{2-10}
 & 50 & \textbf{85.4} & \textbf{94.2} & 63.2 & 84.3 & 35.1 & 52.1 & -4.4 & -2.3
   \\ \cline{2-10}
 & 20 & \textbf{68.7} & \textbf{86.9}  & 42.8 & 68.7 & 24.7 & 41.6 & +1 & -0.5
  \\ \hline
 
100 & 100 & \textbf{78.4} & \textbf{92.5}  & 50.4 & 75.8 & 32.6 & 51.8 & -2.2 & -2.8
  \\ \cline{2-10}
 & 50 & \textbf{71} & \textbf{88.4}  & 40.4 & 66.8 & 27.1 & 48.6 & -0.3 & -1.4
  \\ \cline{2-10}
 & 20 & \textbf{54.9} & \textbf{78.5}  & 27.7 & 52.8 & 18.8 &  37.1 &  +5.5 & +1.7
\\ \hline

\end{tabular}

\end{center}
\caption{Identification accuracy (Rank-1 and Rank-5) of character frequency features (uni-grams) for a different number of training and test Tweets, and for different distance measures. Features employed: Alphabets (lowercase + uppercase) + digits + special characters + TAGS (solid blue curves in Figure~\ref{fig:results_char_unigrams}). Results are given for the ISOT \cite{Brocardo14pst} and SMF \cite{[Rocha17tifsTweets]} databases with 93 test authors. The best case of each row with the ISOT database is marked in bold. Results with the SMF database indicate the accuracy change with respect to the ISOT dataase.}
\label{tab:results_char_unigrams}
\end{table}
\normalsize

Figure~\ref{fig:results_char_counts} shows the CMCs of all the feature combinations evaluated for a selected number of training and test Tweets (using $\chi^2$ only due to space limitations), while Table~\ref{tab:results_char_counts} 
provides results of the best feature combination of Figure~\ref{fig:results_char_counts} (solid blue curves).
The number of training Tweets in this first experimentation starts at 100 Tweets, while the number of test Tweets starts at 20.
%
%
A first observation from the graphs is that separating the count of lowercase and uppercase alphabets (solid curves) systematically results in better performance than if counted together (dashed curves). 
The improvement is higher for combinations with more modest performance, e.g. black or green curves, suggesting that accounting for lowercase and uppercase alphabets separately can provide a significant boost in accuracy.
Regarding the different combinations of count features, the use of only alphabets+digits (black curves) results in the worst performance. Performance is boosted as we add counts of (in this order): vowels (green curves), spaces (gray), and special characters (red). 
But the biggest boost is obtained after the inclusion of the meta tags. The power of such meta tags can be seen by the fact that if we add them just to the count of alphabets+digits (the worst combination), the performance is boosted to top positions (brown curves). 
The best combination overall comes just after the addition of the special character's count (blue curves), but the improvement is marginal in comparison to the brown vs. black curves. 

\begin{figure*}[t]
\centering
        \includegraphics[width=0.9\textwidth]{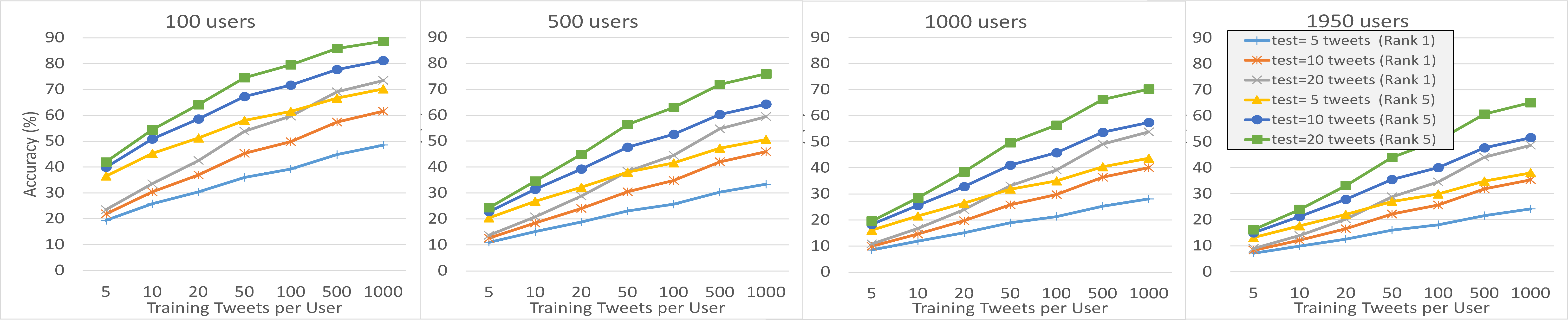}
\caption{Identification accuracy (Rank-1 and Rank-5) of character frequency features (uni-grams) for a reduced number of test Tweets ($\chi^2$ distance). Results are given for the SMF \cite{[Rocha17tifsTweets]} database. Each figure corresponds to a different number of test authors (from left to right: 100, 500, 1000, and 1950 authors). Best in colour.}
\label{fig:results_char_unigrams_rocha_ranks}
\end{figure*}

\begin{figure}[t]
\centering
        \includegraphics[width=0.4\textwidth]{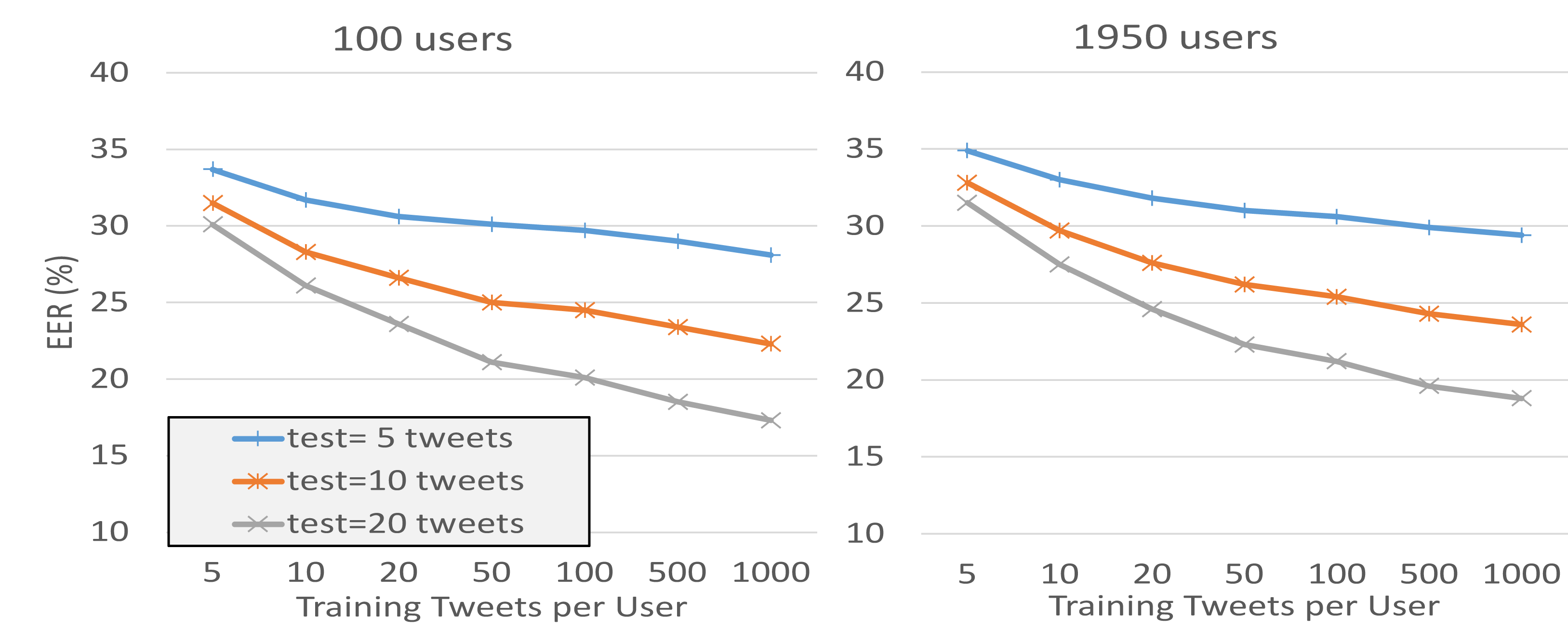}
\caption{Verification accuracy (EER) of character frequency features (uni-grams) for a reduced number of test Tweets ($\chi^2$ distance). Results are given for the SMF \cite{[Rocha17tifsTweets]} database. Each figure corresponds to a different number of test authors. Best in colour.}
\label{fig:results_char_unigrams_rocha_eer}
\end{figure}

If we look at the different plots of Figure~\ref{fig:results_char_counts} (representing a different number of training and test Tweets), the relative positioning of the curves is always the same. 
In other words, changing the amount of training or test data does not make that one particular combination of features surpasses others. Obviously, the higher the amount of training or test data, the better the performance.
With the maximum amount of training and test data available (1000 Tweets in each case), the Rank-1 is 79.8\%, and the Rank-5 is 97\% with the $\chi^2$ distance (Table~\ref{tab:results_char_counts}).
If we decrease the number of test Tweets to 20 (keeping 1000 training Tweets) such ranks decrease to 25.3\% and 55.9\%, respectively.
Also, the performance seems to saturate after the use of 500 Tweets for training. For example, the ranks of 500/500 (training/test) and 1000/500 is similar by 1-2\%. The same goes for other cases.
On the other hand, with just 100 Tweets for training, the performance deteriorates significantly. For example, with 100/100, the ranks in Table~\ref{tab:results_char_counts} are 40.5\% (Rank-1) and 69.5\% (Rank-5).
If test data is more scarce (100/20), the ranks are just 22\% and 49.8\%, respectively.
It should be noted though that the dimensionality of the feature vector employed is just 8 elements (uppercase alphabets, lowercase alphabets, digits, special characters, and the four meta tags).

Regarding the distance and similarity measures employed, Table~\ref{tab:results_char_counts} shows that $\chi^2$ always gives the best results, and Euclidean performs significantly worse. 
This has been also observed in the other feature combinations of Figure~\ref{fig:results_char_counts}, although it is not shown here due to page limit. 

We further evaluate the best feature combination 
on the larger SMF database. For comparative purposes, we first report in Table~\ref{tab:results_char_counts} (right) the accuracy with the same number of authors than the ISOT database (93 authors, 5-fold cross-validation). A first observation is that there is no database for which the accuracy is clearly better. In most cases, Rank-1 is better with SMF, while Rank-5 with the ISOT database.
The biggest difference (in favour of SMF) happens when only 20 Tweet are used for testing (regardless of the number of training Tweets). In the remaining cases, the difference between the two databases is mostly within 2-4\%. 
To evaluate scenarios more suited to social media forensic cases \cite{[Rocha17tifsTweets]}, we then report experiments where the number of test Tweets can be very small (we evaluate 5, 10 and 20 test Tweets). 
Identification accuracy is reported in Figure~\ref{fig:results_char_counts_rocha_ranks}, whereas verification accuracy is given in Figure~\ref{fig:results_char_counts_rocha_eer}. We also vary the number of authors in the database from 100 to 1950, according to the protocol of Section~\ref{sect:db_protocol}.
As the number of authors increases, the identification accuracy decreases accordingly. 
A bigger amount of training Tweets provides better chances of identification, but it is not always the case that investigators have as many training samples at their disposal. 
In all cases, however, there is plenty of room for improvement, since the best Rank-1 in any experiment is below 40\%, being less than 15\% with 1950 authors in the database. 
Regarding verification experiments, the amount of authors in the database does not have the same impact as with identification tests. As it can be seen in Figure~\ref{fig:results_char_counts_rocha_eer}, the EER for each training/test combination does not increase substantially when going from 100 to 1950 authors. As it can be expected, the EER correlates with the amount of data. With 5 training Tweets, it ranges around 34-36\%, going down to 22-24\% with 1000 training and 20 test Tweets.

\subsection{Character Frequency (uni-grams)}
\label{sect:results:unigrams}

Character uni-grams compute the frequency of individual alphabets, digits, and special characters, rather than their total counts. 
They are denoted as Lexical-Character-Frequency in Table~\ref{tab:features}.
As in the previous sub-section,
to populate uni-grams, we consider different combinations of elements, including the use of meta tags.
We also distinguish the case where the frequency of uppercase and lowercase alphabets is computed separately vs. computed together.
Identification results with the ISOT database 
are given in Figure~\ref{fig:results_char_unigrams} (all combinations) and Table~\ref{tab:results_char_unigrams} (best one).

The majority of observations of the previous sub-section apply here too, i.e. $i$) separating lowercase and uppercase alphabets results in better performance (solid vs. dashed curves); $ii$) the use of only alphabets+digits results in the worst performance, which is improved after the inclusion of white spaces (grey curves), special characters (red), and meta tags (blue); and $iii$) meta tags is a valuable source, as seen by the fact that its inclusion boosts performance significantly, e.g. black vs, brown curves, or red vs. blue.
It can also be observed that white spaces does not provide significant value. For example, compare black/grey curves, or red/green; both pairs employ the same combination of features, with the only difference being the inclusion of data about white spaces.
White spaces contain (indirectly) information about the number of words in the text, but their addition here does not result in a better performance.

As in the previous sub-section, the best combination includes the use of alphabets (separating lowercase/uppercase), digits, special characters, and meta tags.
In addition, the $\chi^2$ distance also stands out.
However, a significant difference is that character uni-grams provide a much better accuracy. 
For example, with a training and test size of 1000 Tweets, Rank-1 is already 99\% (see Table~\ref{tab:results_char_unigrams}).
It should be considered though that the dimensionality of the feature vector with uni-grams is of 99, while the vector of the previous sub-section was of 8 elements.
Decreasing the test Tweets to 50 (keeping 1000 training Tweets) still gives a Rank-5 of 95\%.
Therefore, if we can ensure such a high amount of training data, uni-grams can provide very good performance with a relatively small amount of test data. 

With the 1000/20 case, accuracy is 70.6\% (Rank-1) and 87.8\% (Rank-5). In the previous sub-section, it was 25.3\%/55.9\% only.
It seems to be as well a saturation in performance here after the use of 500 training Tweets, since the use of 500 or 1000 training Tweets does not show a significant difference (given the same number of test Tweets).
With just 100 training Tweets, performance is still good if we use 100 test Tweets as well (Rank-5 of 92.5\% vs. 69.5\% in the previous sub-section). 
With fewer data (100/20), the ranks go down to 54.9\% (Rank-1) and 78.5\% (Rank-5); 
but if we allow a list of 20 candidates, then $\sim$92\% can be obtained
(Figure~\ref{fig:results_char_unigrams}, bottom right plot).
This shows as well the capabilities of uni-grams when fewer data is available.

\begin{figure}[t]
\centering
        \includegraphics[width=0.4\textwidth]{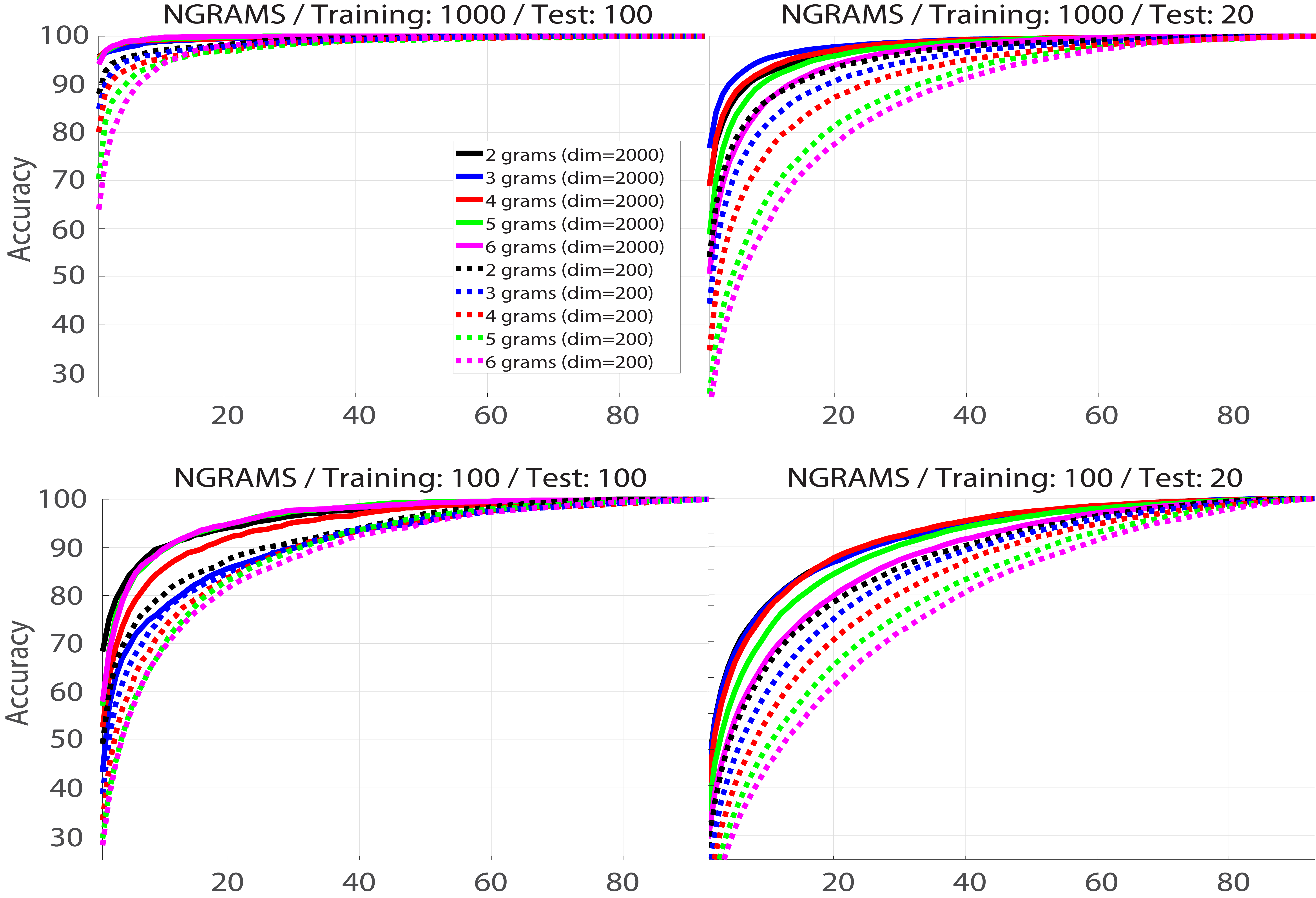}
\caption{Identification accuracy of character frequency features ($n$-grams, $n>1$) for a different number of training and test Tweets ($\chi^2$ distance). Results are given for the ISOT \cite{Brocardo14pst} database with 93 test authors. Best in colour.}
\label{fig:results_char_ngrams}
\end{figure}

\setlength{\tabcolsep}{0pt}
\begin{table}[t]
\begin{center}
\begin{tabular}{|>{\centering}m{20pt}|>{\centering}m{20pt}||>{\centering}m{20pt}|>{\centering}m{20pt}||>{\centering}m{20pt}|>{\centering}m{20pt}|||>{\centering}m{20pt}|>{\centering}m{20pt}||>{\centering}m{20pt}|>{\centering\arraybackslash}m{20pt}|}

\multicolumn{2}{c}{} & \multicolumn{8}{c}{\textbf{Character Frequency ($n$-grams, $n>1$)}} \\ \cline{3-10}

\multicolumn{2}{c|}{} & \multicolumn{4}{c|||}{3-grams (dim=2000)} & \multicolumn{4}{c|}{2-grams (dim=200)}  \\ \cline{3-10}

\multicolumn{2}{c|}{} & \multicolumn{2}{c||}{\textbf{ISOT DB}} & \multicolumn{2}{c|||}{\textbf{SMF DB}} & \multicolumn{2}{c||}{\textbf{ISOT DB}} & \multicolumn{2}{c|}{\textbf{SMF DB}}  \\ \hline


\textbf{Train} & \textbf{Test} & \textbf{R-1} & \textbf{R-5} &  \textbf{R-1} & \textbf{R-5}  & \textbf{R-1} & \textbf{R-5}  & \textbf{R-1} & \textbf{R-5} \\ \hline

%
%

1000 & 1000 & 98 & 98 & -0.4 & +1.1  & 98 & 98  & -0.4 & +1.4  \\ \cline{2-10}
 & 500 & 97.3 & 98.4 & +0.4 & +0.2  &   97.7 & 98  & -0.8 & +0.5  \\ \cline{2-10}
 & 100 & 94.9 & 97.6 & +0.2 & +0.2  &   88 & 95.6  & +4.1 & +1.7  \\ \cline{2-10}
 & 50 & 92.3 & 96.9 & +0.3 & +0.3  &   77.4 & 92.5  & +8 & +2.9   \\ \cline{2-10}
 & 20 & 76.7 & 91.4 & +3.8 & +2.2  &   54.1 & 78.7  & +16.3 & +9.8  \\  \hline
 
500 & 500 & 94 & 97.7 & +0.3 & -0.6  &   96 & 98  & -2.9 & -1.9   \\ \cline{2-10}
 & 100 & 90.9 & 96.3 & +1 & -0.2  &   84.4 & 94.9  & +1.9 & -0.5   \\ \cline{2-10}
 & 50 & 86.7 & 95.4 & +1.9 & -0.3  &   71.5 & 89.7  & +7.7 & +2  \\ \cline{2-10}
 & 20 & 69.8 & 88 & +5.6 & +2.5   &   48.9 & 75.1  & +15.1 & +8.6  \\  \hline
 
100 & 100 & 43.3 & 69.4 & +13.6 & +8.3  &   49.2 & 71.7  & +12.6 & +9.9  \\ \cline{2-10}
 & 50 & 48 & 72 & +13 & +8.8  &   39.1 & 65  & +16.8 & +12.8  \\ \cline{2-10}
 & 20 & 42.8 & 67.8 & +13.6 & +10.1  &   27.5 & 52.3  & +19 & +17.1   \\  \hline

\end{tabular}

\end{center}
\caption{Identification accuracy (Rank-1 and Rank-5) of character frequency features ($n$-grams, $n>1$) for a different number of training and test Tweets ($\chi^2$ distance). Results are given for the ISOT \cite{Brocardo14pst} and SMF \cite{[Rocha17tifsTweets]} databases with 93 test authors. 
Results with the SMF database indicate the accuracy change with respect to the ISOT database.}
\label{tab:results_char_ngrams}
\end{table}
\normalsize
\setlength{\tabcolsep}{6pt}

We then evaluate the best feature combination of this section with the larger SMF database. First, Table~\ref{tab:results_char_unigrams} (right) shows the identification accuracy with 93 authors to compare with the same setup than the ISOT database.
As in the previous sub-section, the difference between the two databases is small, and in this case the margin is even smaller (2\% or less in many cases). As observed earlier, the features of this sub-section are more powerful, so the difference between the two databases is reduced. 
Even if the writing style of influential tweeters or the general public may differ, these features are agnostic of such differences, at least with the amount of training and test data employed in Table~\ref{tab:results_char_unigrams}. 
Next, in Figures~\ref{fig:results_char_unigrams_rocha_ranks} and \ref{fig:results_char_unigrams_rocha_eer}, we provide the author identification and verification accuracy with a small number of test Tweets (5, 10 and 20). 
Apart from the expected correlation between accuracy and authors in the database or number of training/test Tweets, the plots reflect the better accuracy of character unigrams, in comparison to features of the previous sub-section. 
Here, with a sufficient number of training Tweets (1000), the Rank-1 accuracy with 100 authors is between 50-70\%, and the Rank-5 is between 70-90\%. With 1950 authors, the Rank-5 is 50\% with 5 test Tweets only, and it goes up to 66\% with 20 Tweets. 
Naturally, when the amount of training Tweets is less, the accuracy is worsened. With 100 authors, the Rank-5 can be kept above 50\% with 10 or 20 training Tweets only. With 1950 authors, however, the Rank-5 is reduced to 25-35\% in the same conditions. While this is an improvement in comparison to the previous sub-section, there is still work to do in author identification with short texts when the number of authors in the database increases to thousands of people and beyond. 
Regarding verification experiments, there is also certain resilience with these features to the number of authors in the database, but with better EER values. With 5 training Tweets for example, the EER ranges around 30-35\% (34-36\% in the previous sub-section), going down to 17-19\% with 1000 training Tweets and 20 test Tweets (22-24\% in the previous sub-section).

\subsection{Character Frequency ($n$-grams, $n>1$)}
\label{sect:results:ngrams}

These features capture the occurrence of the most frequent character $n$-grams. 
Depending on the value of $n$, they use pairs of consecutive characters (called 2-grams or bi-grams), triplets (3-grams or tri-grams), etc.
The ten most frequent $n$-grams of the ISOT database for $n$ = 2, 3, 4 are given in Table~\ref{tab:frequent_ngrams}.
%
Some recognizable words of the English language (or parts of them) appear in the list, such as `in', `the', `you', `and', `for', `with', etc.
Some works prune the dictionaries to remove common words that are likely to be used by many, such as articles, prepositions, etc. (called function or stop words).
However, since they tend to be the words that occur more frequently in general language (as corroborated by Table~\ref{tab:frequent_ngrams}), they can be reliable features for authorship attribution \cite{[Rocha17tifsTweets]}, specially with short texts where most of the words appear only once.
Indeed, statistics of function words were one of the earliest features suggested for manual attribution \cite{[Mascol1888functionwords]}.
%
%
%
Therefore, removing them can be counterproductive, since the amount of data is already limited \cite{Forstall09featuresfrom}. 
%
%
For these reasons, in this work, we do not remove any function word prior to generating the dictionaries. This also helps to keep the extraction simpler, since they 
%
%
are built by pure analysis of raw texts, without further considering higher levels of syntactical or semantic analysis that are language-dependant and might be prone to errors.

\begin{figure*}[t]
\centering
        \includegraphics[width=0.9\textwidth]{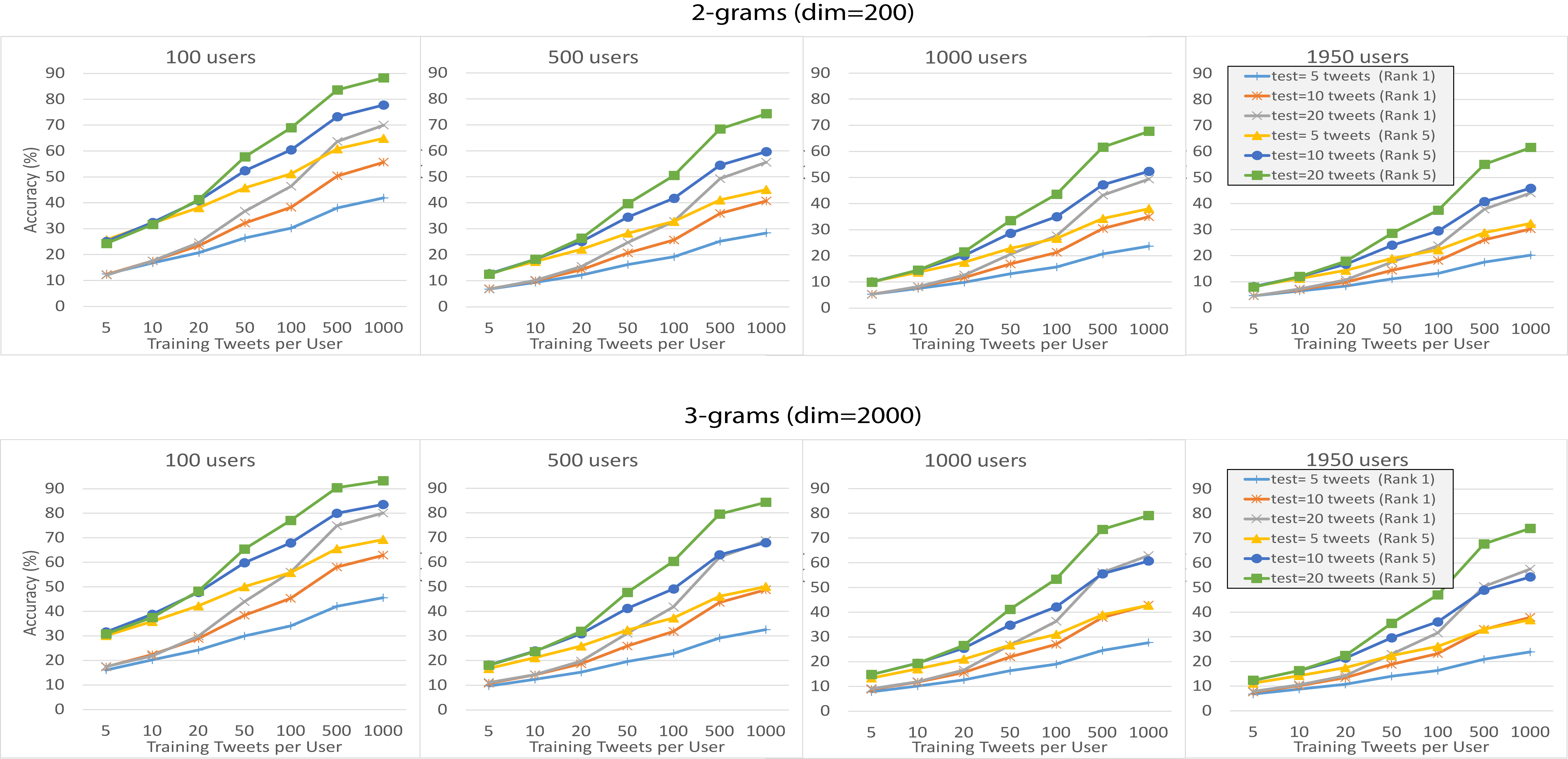}
\caption{Identification accuracy (Rank-1 and Rank-5) of character frequency features ($n$-grams, $n>1$) for a reduced number of test Tweets ($\chi^2$ distance). Results are given for the SMF \cite{[Rocha17tifsTweets]} database. Each figure corresponds to a different number of test authors (from left to right: 100, 500, 1000, and 1950 authors). Best in colour.}
\label{fig:results_char_ngrams_rocha_ranks}
\end{figure*}

\begin{figure}[t]
\centering
        \includegraphics[width=0.4\textwidth]{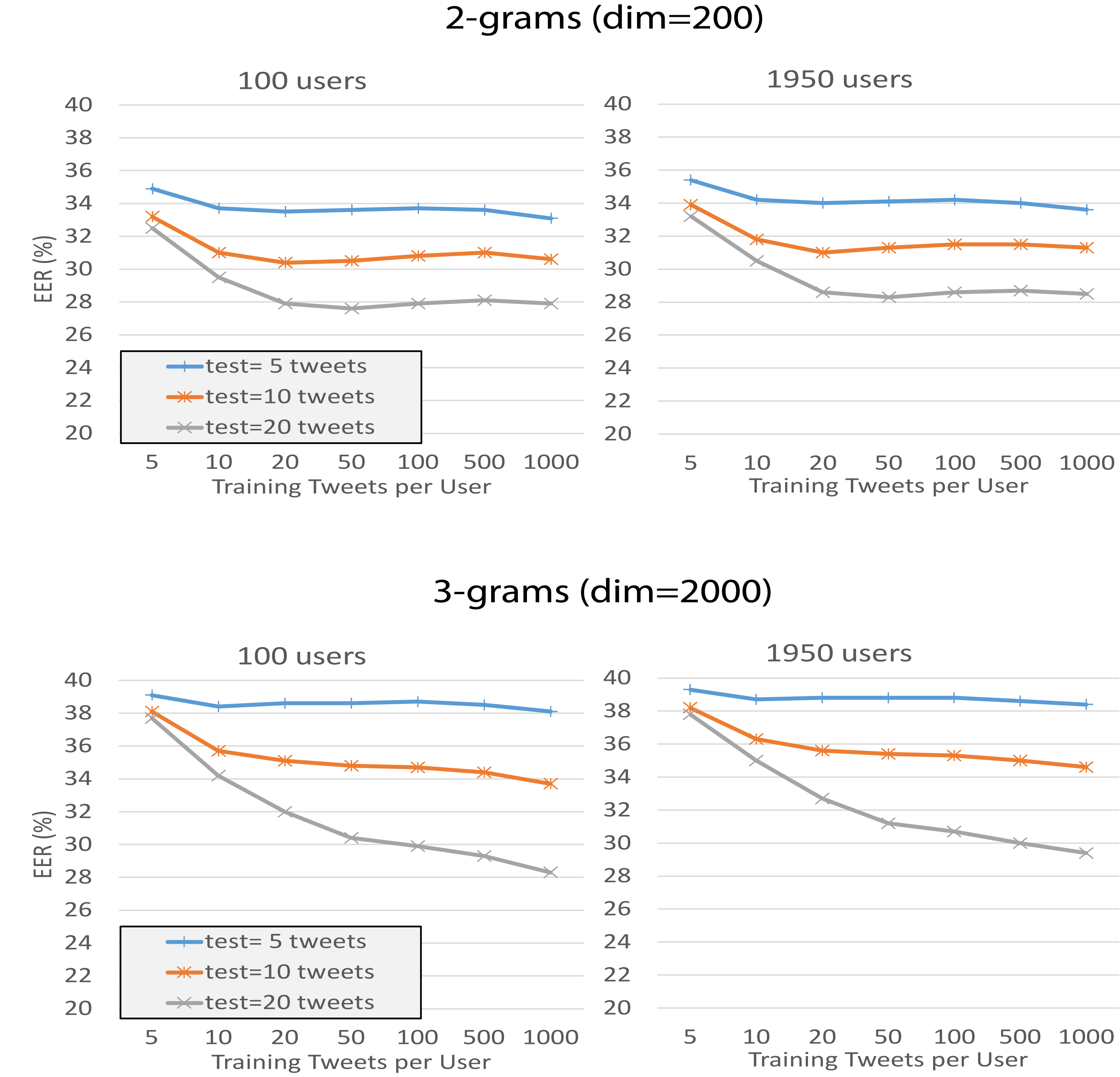}
\caption{Verification accuracy (EER) of character frequency features ($n$-grams, $n>1$) for a reduced number of test Tweets ($\chi^2$ distance). Results are given for the SMF \cite{[Rocha17tifsTweets]} database. Each figure corresponds to a different number of test authors. Best in colour.}
\label{fig:results_char_ngrams_rocha_eer}
\end{figure}

We test two cases, one considering a big dictionary of $n$-grams (the 2000 most frequent $n$-grams), and another considering a reduced dictionary (the 200 most frequent only). 
Identification results with the ISOT database are shown in Figure~\ref{fig:results_char_ngrams} and Table~\ref{tab:results_char_ngrams}.
Given the inferiority of the Euclidean distance and cosine similarity in previous sub-sections, we omit their results in the reminder of the paper for space-saving purposes.
It can be observed that a dictionary of 2000 $n$-grams provides much better performance, to the point that the value of $n$ becomes irrelevant when a high amount of training data is employed (e.g. solid curves in 1000/100 plots of Figure~\ref{fig:results_char_ngrams}).
The use of fewer test data (plots of the right column) reveals some differences depending on the value of $n$, with $n$=6 (purple curves) being the worst case, followed by $n$=5 (green).
With a small dictionary of 200 elements, $n$=2 (bi-grams) is always the best case (dashed black curves).
With a big dictionary, on the other hand, $n$=3 (tri-grams) stands out as the best option (solid blue curves).
This means that if we allow a sufficient dictionary size, tri-grams surpasses the performance of any other option.
This is specially the case when there is few training or test Tweets, as it can be observed in Table~\ref{tab:results_char_ngrams}.

Regarding performance values, tri-grams provide a similar performance than uni-grams (previous sub-section) when there is sufficient  of training and test Tweets. 
This can be seen by comparing Tables~\ref{tab:results_char_unigrams} and \ref{tab:results_char_ngrams}.
With 1000 or 500 training Tweets, their performance is comparable, regardless of the number of test Tweets.
With 100 training Tweets, on the other hand, uni-grams shows better performance. 
For example, the 100/100 combination has a Rank-1/Rank-5 of 43.3\%/69.4\% with tri-grams, and 78.4\%/92.5\% with uni-grams.
The more data-scarce 100/20 combination shows ranks of 42.8\%/67.8\% (tri-grams), and 54.9\%/78.5\% (uni-grams).
This is interesting, given that uni-grams have a feature vector of only 99 elements, in comparison to the 2000 elements of tri-grams.
If we allow a hit list of 20 candidates, tri-grams shows an accuracy of $\sim$87\%, behind the 92\% seen with uni-grams in the previous sub-section.

%
%

\begin{figure}[t]
\centering
        \includegraphics[width=0.4\textwidth]{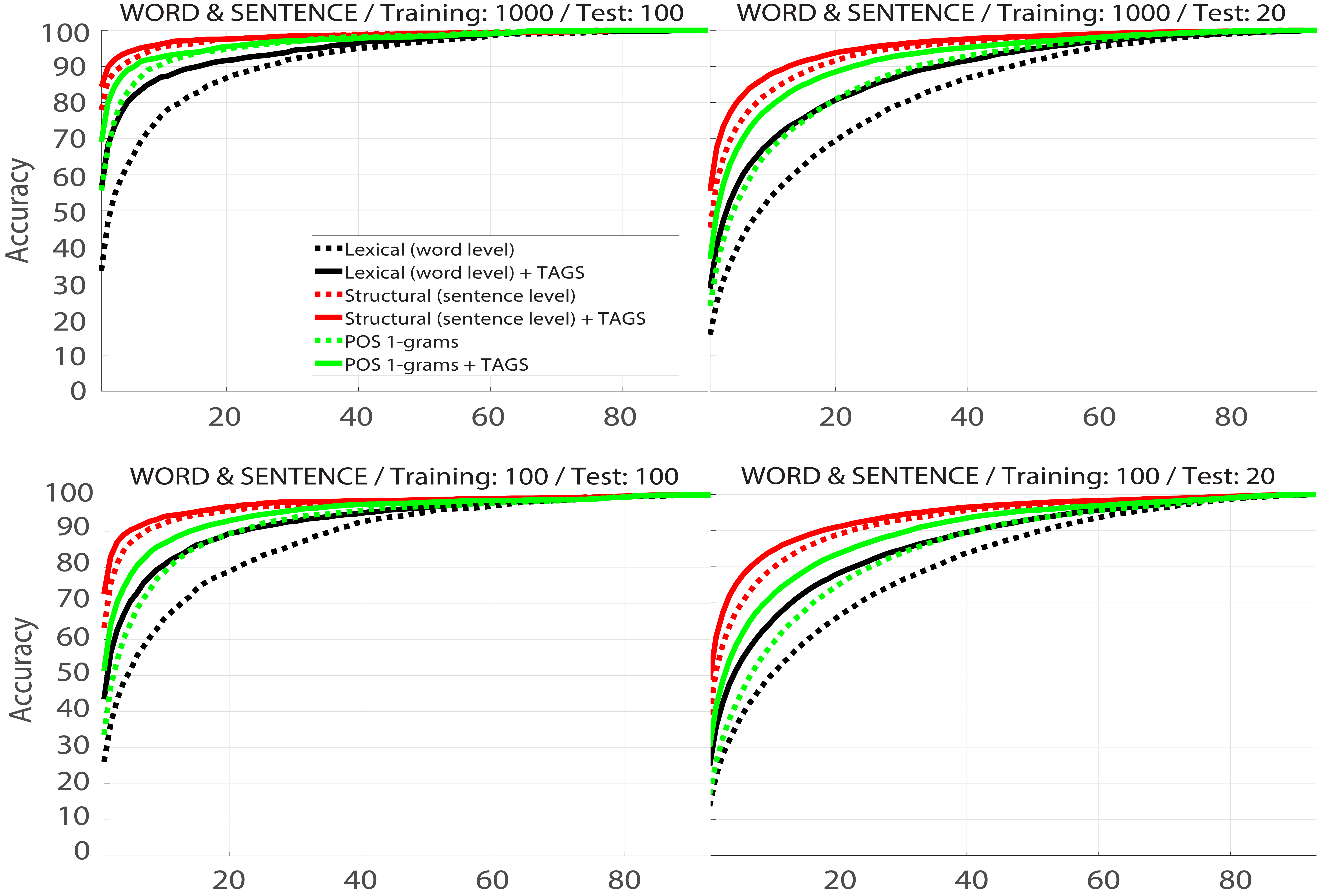}
\caption{Identification accuracy of word and sentence level features for a different number of training and test Tweets using the $\chi^2$ distance. Results are given for the ISOT \cite{Brocardo14pst} database with 93 test authors. Best in colour.}
\label{fig:results_word_sentence}
\end{figure}

\begin{table*}[t]
\begin{center}
\begin{tabular}{|c|c|||c|c||c|c|||c|c||c|c|||c|c||c|c|}

\multicolumn{8}{c}{} \\ \cline{3-14}

\multicolumn{2}{c|}{} & \multicolumn{4}{c|||}{Word} & \multicolumn{4}{c|||}{Sentence} & \multicolumn{4}{c|}{POS 1-grams}
 \\ \cline{3-14}
 
 
 \multicolumn{2}{c|}{} & \multicolumn{2}{c||}{\textbf{ISOT DB}} & \multicolumn{2}{c|||}{\textbf{SMF DB}} & \multicolumn{2}{c||}{\textbf{ISOT DB}} & \multicolumn{2}{c|||}{\textbf{SMF DB}} & \multicolumn{2}{c||}{\textbf{ISOT DB}} & \multicolumn{2}{c|}{\textbf{SMF DB}}  \\ \hline
 
\textbf{Train} & \textbf{Test} & \textbf{R-1} & \textbf{R-5} & \textbf{R-1} & \textbf{R-5} & \textbf{R-1} & \textbf{R-5} & \textbf{R-1} & \textbf{R-5} & \textbf{R-1} & \textbf{R-5} & \textbf{R-1} & \textbf{R-5} \\ \hline

1000 & 1000 & 83.8  & 93.9  &    +6.4 & +2.3
     & 94.9  & 98   &    +1 & +0.1
    & 92.9  & 97  &    +1.9 & +1.5
     \\  \cline{2-14}
 & 500 & 77  & 89.8  &    +7.8 & +5.1
      & 93.4 & 96.5  &    -3.2 & +0.3
      & 88.3  & 95.3  &    +0.2 & +1.7
      \\  \cline{2-14}
 & 100 & 56.1  & 80.2  &    +13.7 & +8.9
      & 84.3  & 94  &    -2.9 & -0.9
      & 69  & 89  &    +9 & +4.5
     \\  \cline{2-14}
 & 50 & 45  & 72.2  &    +16.2 & +11.4
     & 74.7  & 90.4  &    -0.4 & -0.4
     & 56.3  & 81.9  &    +13.8 & +8.1
     \\  \cline{2-14}
 & 20 & 28.5  & 56.6  &    +19.8 & +16.9
      & 55.5   & 79.9  &    +5.7 & +3
      & 36.6  & 66.8  &    +20.1 & +15
     \\  \hline
 
500 & 500 & 74.5  & 89.7  &    +3.8 & +3.1
      & 93.1  & 97.1  &    -5.1 & -1.7
      & 84.5  & 94.3  &    +1 & +0.6
     \\  \cline{2-14}
 & 100 & 55.2  & 78.5  &    +9 & +7.1
    & 81.2  & 93.8  &    -4 & -2.6
      & 67  & 87.3   &    +6.9 & +3.2
     \\  \cline{2-14}
 & 50 & 42.6  & 71.4  &    +13.7 & +9.1
      & 72.3  & 89.5  &    -1.9 & -1.8
      & 55.1  & 80.2  &    +11 & +7.1
      \\  \cline{2-14}
 & 20 & 27.8  & 56.4  &    +16.4 & +13.9
      & 54  & 78.9  &    +3.6 & +1.4
      & 36.7  & 66.2   &    +16 & +12.7
     \\  \hline
 
100 & 100 & 43.4  & 70.5  &   +8.7 & +5.9
     & 72.6  & 90.3  &    -6.7 & -4.7
      & 51.3  & 77.7   &    +10.3 & +6
     \\  \cline{2-14}
 & 50 & 36.1  & 62.7  &    +10 & +9.2
      & 65  & 85.7  &    -4.2 & -3.7
      & 42.7  & 71.2    &    +13.1 & +8.6
     \\  \cline{2-14}
 & 20 & 24.9  & 51.3  &    +12.6 & +11.8
     & 48.7  & 75.1  &    +1.7 & -0.9
      & 30.1  & 58.3    &    +15.3 & +13.6
     \\  \hline

\end{tabular}

\end{center}
\caption{Identification accuracy (Rank-1 and Rank-5) of word and sentence level features for a different number of training and test Tweets ($\chi^2$ distance). The features include meta tags counts (solid curves in Figure~\ref{fig:results_word_sentence}). Results are given for the ISOT \cite{Brocardo14pst} and SMF \cite{[Rocha17tifsTweets]} databases with 93 test authors. 
Results with the SMF database indicate the accuracy change with respect to the ISOT database.}
\label{tab:results_word_sentence}
\end{table*}
\normalsize

Table~\ref{tab:results_char_ngrams} also shows the identification accuracy of the SMF database with 93 authors. When the number of training and test Tweets is high, ISOT and SMF show similar accuracy. But contrarily to the previous sub-sections, when there are few data for training and testing, the SMF database surpass the ISOT database by a large margin (10-20\% better accuracy). This is interesting, since the $n$-grams dictionaries are computed from the ISOT database, and the Tweets of the ISOT database have more length on average (Figure~\ref{fig:dbstats}), so one would expect precisely the opposite when there is scarcity in the amount of Tweets.
Next, author identification and verification accuracy with a smaller number of test Tweets (5, 10 and 20) is given in Figures~\ref{fig:results_char_ngrams_rocha_ranks} and \ref{fig:results_char_ngrams_rocha_eer}. 
The identification curves confirm some of the conclusions of the previous paragraphs, i.e. $i$) that tri-grams with a big dictionary of 2000 elements is better than bi-grams with a dictionary of 200 elements (observe that values on the $y$-axes are higher); and $ii$) 
that uni-grams (previous sub-section) and $n$-grams (this sub-section) provide a similar accuracy with a sufficient number of data, but with less training or test data, uni-grams surpasses the features of this sub-section.
The latter can be seen in the exponential-like shape of the curves of Figure~\ref{fig:results_char_ngrams_rocha_ranks}, which reach comparable values than the curves of Figure~\ref{fig:results_char_unigrams_rocha_ranks} on the right side of the $x$-axes, but they go down quickly as we move left to the origin, instead of decreasing linearly. 
In other words, $n$-grams appear to be more sensitive to reductions in the amount of training and test data. 
For example, with 1950 authors and 1000 Tweets for training, the Rank-5 accuracy here (3-grams) is 55-75\% (50-65\% in the previous sub-section).
But if we reduce the training Tweets to 50, the Rank-1 accuracy here is 35\% or less, while in the previous sub-section, it was 35-45\%. 
Regarding verification experiments, the curves of Figure \ref{fig:results_char_ngrams_rocha_eer} show some interesting differences with the identification case. For example, the EER with 2-grams is better than with 3-grams (in identification, these features behave the other way around). Another interesting phenomenon is that the EER plateaus in most cases after 20 training Tweets. This contrasts with other sub-sections, where more training Tweets always translates to a better EER. 
The only exception is tri-grams with 20 test Tweets (gray curves), but the EER that it reaches with 1000 training Tweets is achievable with bi-grams and only 20 training Tweets.
Comparatively speaking, with few training Tweets, bi-grams and unigrams (previous sub-section) show EERs in a comparable range, for example: 32-35\% vs. 30-34\% (5 training Tweets), 29-34\% vs. 26-33\% (10 training Tweets), or 28-34\% vs. 24-31\% (20 training Tweets). But when the number of training Tweets is increased, the mentioned plateau with $n$-grams appears. With a high amount of training Tweets (1000), the EER is 28-34\% (bi-grams) vs. 17-30\% (uni-grams).

\begin{figure*}[t]
\centering
        \includegraphics[width=0.9\textwidth]{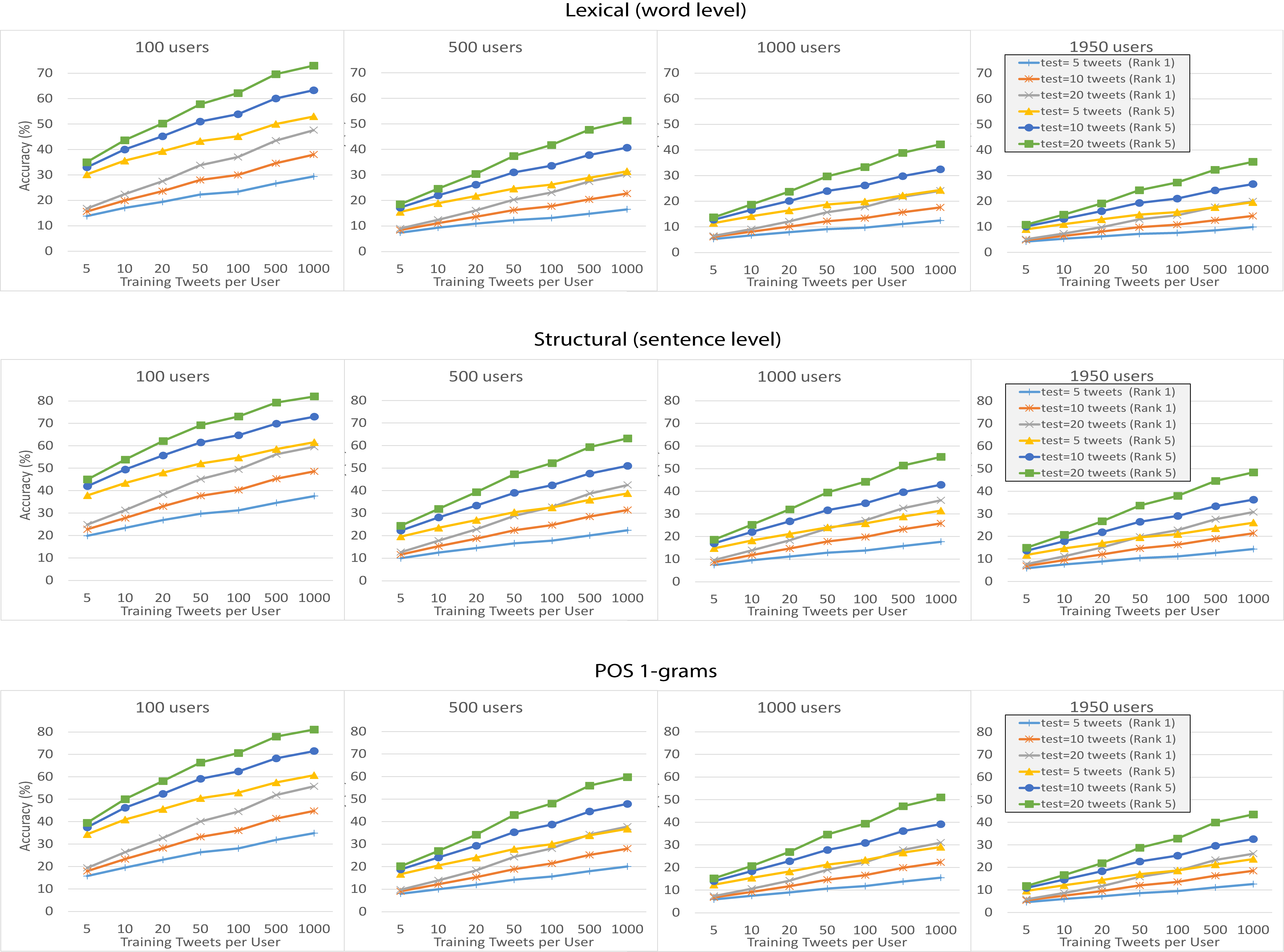}
\caption{Identification accuracy (Rank-1 and Rank-5) of word and sentence level features for a reduced number of test Tweets ($\chi^2$ distance). Results are given for the SMF \cite{[Rocha17tifsTweets]} database. Each figure corresponds to a different number of test authors (from left to right: 100, 500, 1000, and 1950 authors). Best in colour.}
\label{fig:results_word_sentence_rocha_ranks}
\end{figure*}

\begin{figure}[t]
\centering
        \includegraphics[width=0.4\textwidth]{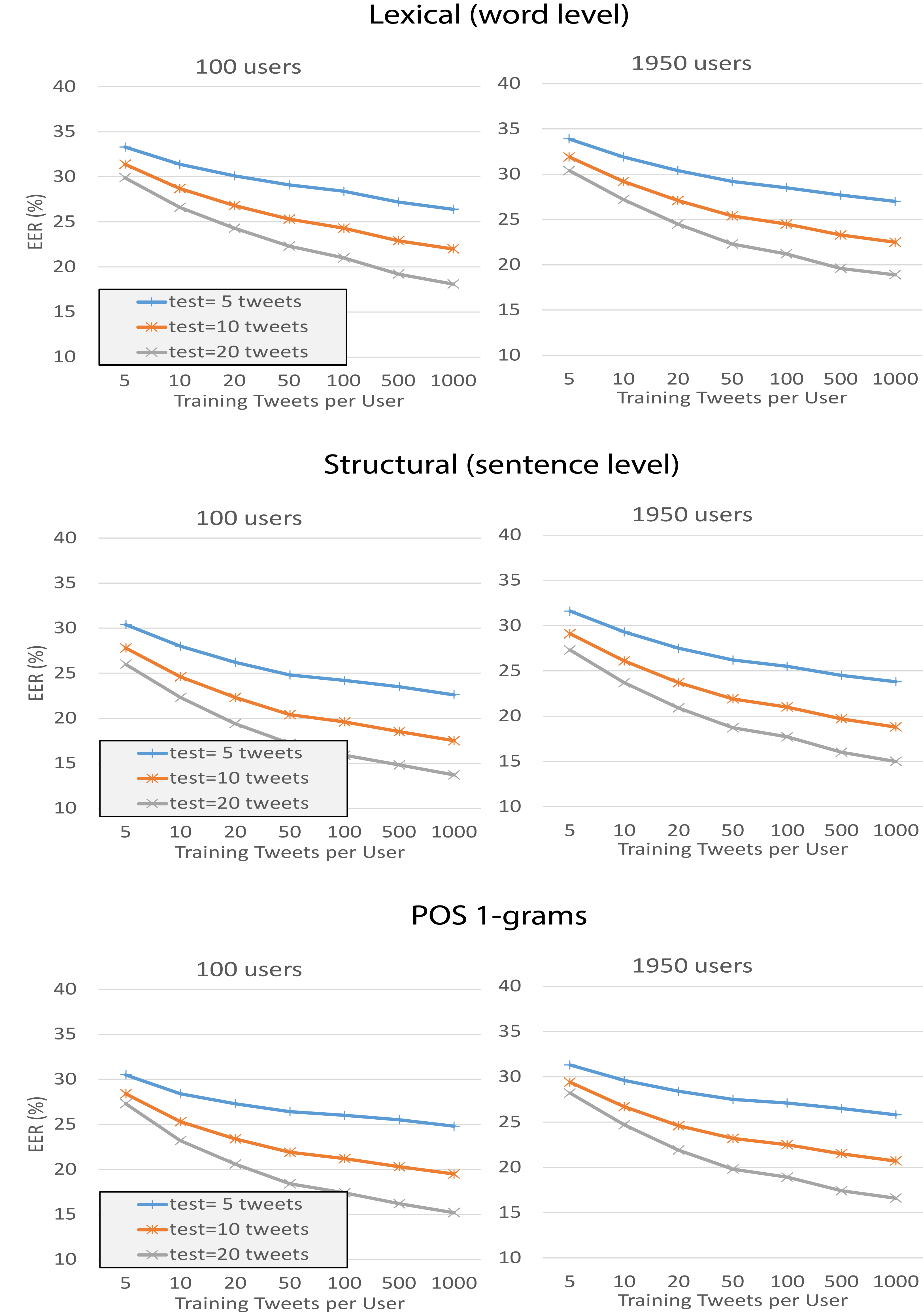}
\caption{Verification accuracy (EER) of word and sentence level features for a reduced number of test Tweets ($\chi^2$ distance). Results are given for the SMF \cite{[Rocha17tifsTweets]} database. Each figure corresponds to a different number of test authors. Best in colour.}
\label{fig:results_word_sentence_rocha_eer}
\end{figure}

\subsection{Word- and Sentence-Level Features}
\label{sect:results:word_sentence}

In this sub-section, we analyze together the performance of several lexical, structural and syntactic features that capture properties at the word and sentence level.
%
As done in some previous sub-sections, we also test the different groups of features with and without adding the meta tags that characterize Twitter posts (URLs, user mentions, hashtags, and quotations per Tweet).
Identification results with the ISOT database are given in Figure~\ref{fig:results_word_sentence} and Table~\ref{tab:results_word_sentence}.

According to the plots, structural features capturing information at the sentence level (red curves) provides better performance than the other two.
The worst features are the lexical features operating at the word level.
This is interesting, given that Tweets are likely to have few sentences given its limited length (140 characters). 
Accuracy of POS tag uni-grams is in the middle of the other two groups.
The features of this section have a similar dimensionality, see Table~\ref{tab:features}, so differences in performance cannot be attributed to the size of feature vectors.
%
Also, it is relevant here the improvement observed with the inclusion of meta tags (solid vs. dashed curves). 
Comparatively speaking, the performance of the features of this section is behind uni-grams and $n$-grams of Sections~\ref{sect:results:unigrams} and \ref{sect:results:ngrams}, and ahead of the count features of Section~\ref{sect:results:char_count}.
Only when there are few training or test Tweets available, the performance of lexical word features (the worst performing of this section) becomes similar to the features of Section~\ref{sect:results:char_count} (compare Tables~\ref{tab:results_char_counts} and \ref{tab:results_word_sentence}).

Table~\ref{tab:results_word_sentence} also provides the identification accuracy of the SMF database with 93 authors. 
Interestingly, the accuracy with word features and POS tag uni-grams is significantly better with the SMF database.
This is observed regardless of the number of training and test Tweets, although it is more pronounced when few test samples are used (see the rows with 20 or 50 test Tweets).  
This is in contract to previous sub-sections, where the accuracy between the two databases was not so different in the majority of combinations. 
Word features quantify aspects like number of words, word length or use of capitalization, while POS tag uni-grams analyze the use of different syntactic elements. These results suggest idiosyncratic differences in the use of language words by influential tweeters and the general public, while the features of the previous sub-sections are more agnostic to such differences.
Sentence-level features, on the other hand, show a comparable accuracy between the two databases. 
Finally, Figures~\ref{fig:results_word_sentence_rocha_ranks} and \ref{fig:results_word_sentence_rocha_eer} show the author identification and verification accuracy with a small number of test Tweets (5, 10 and 20). 
The identification curves show the same correlation as in previous sub-sections between accuracy and authors in the database or number of training/test Tweets, while the verification curves also show the resilience of these features to the number of authors in the database.


\begin{figure}[t]
\centering
        \includegraphics[width=0.33\textwidth]{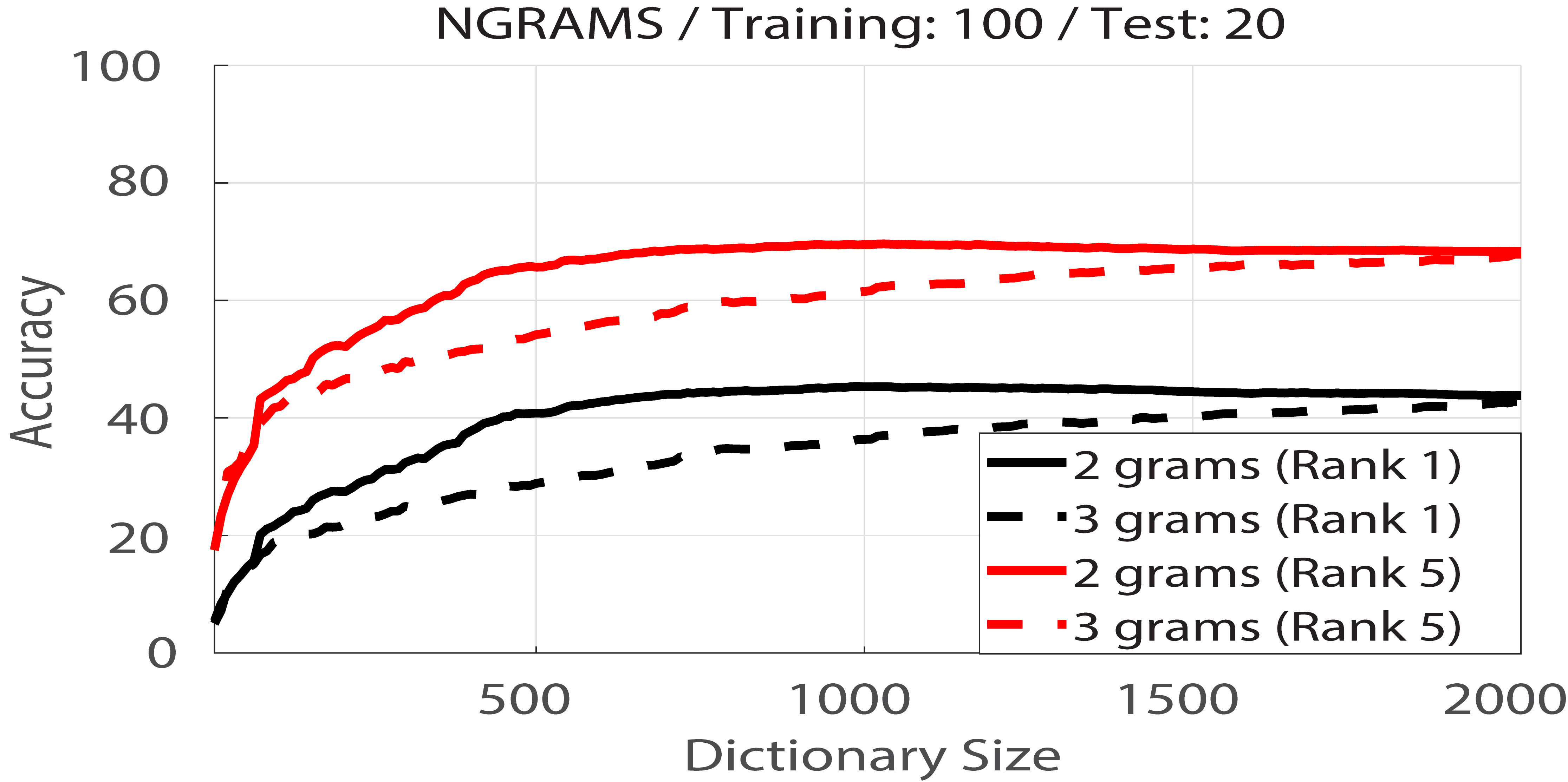}
\caption{Identification accuracy (Rank-1 and Rank-5) of 2-grams and 3-grams for different dictionary sizes using the $\chi^2$ distance (training: 100 Tweets, test: 20 Tweets). Results are given for the ISOT \cite{Brocardo14pst} database with 93 test authors. Best in colour.}
\label{fig:results_ngrams_dictionary}
\end{figure}

\begin{figure}[t]
\centering
        \includegraphics[width=0.33\textwidth]{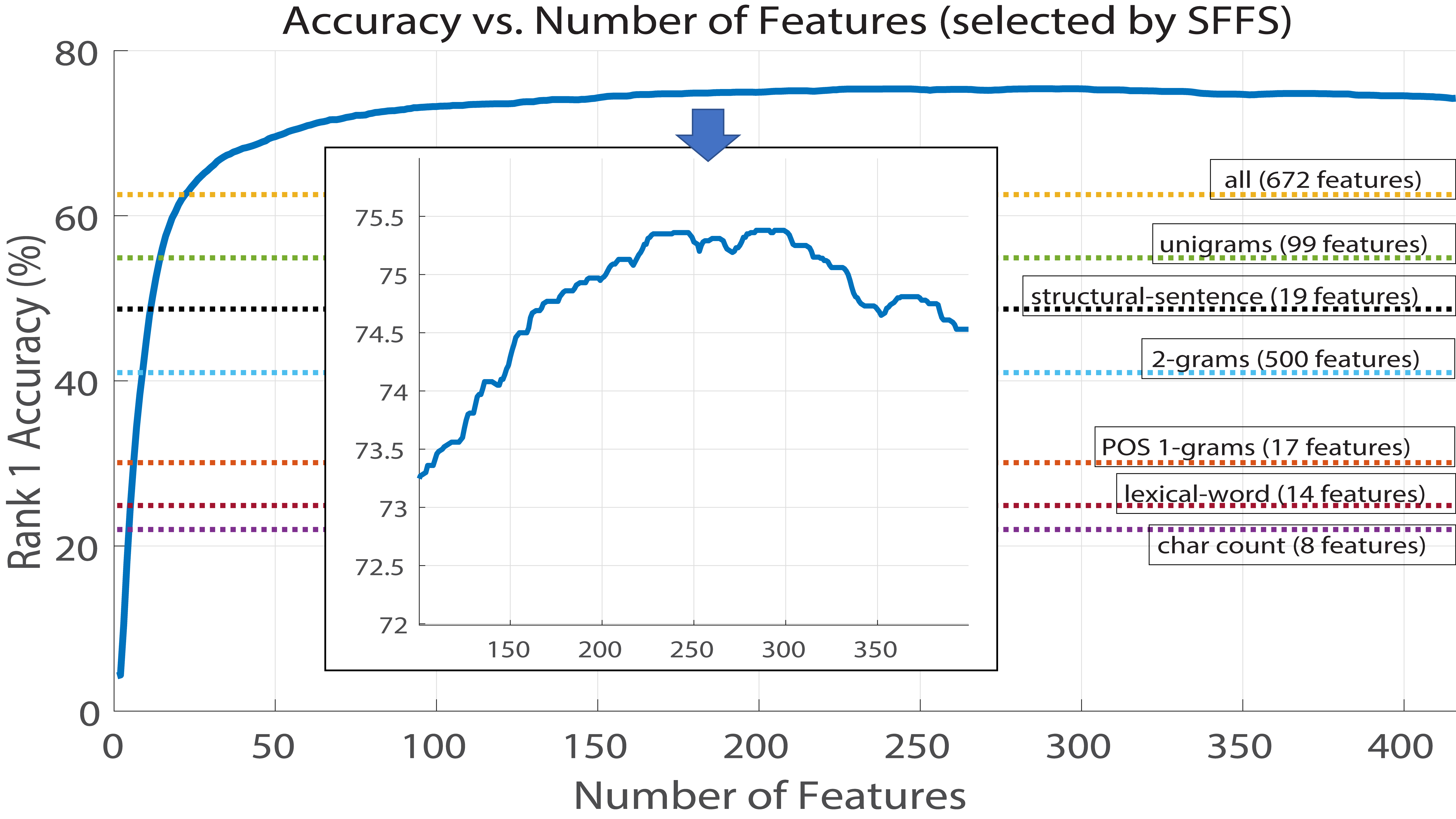}
\caption{Identification accuracy (Rank-1) for an increasing number of features selected with SFFS (training: 100 Tweets, test: 20 Tweets). Results are given for the ISOT \cite{Brocardo14pst} database with 93 test authors. Performance of the best individual features of Sections~\ref{sect:results:char_count}-\ref{sect:results:word_sentence} is also given for reference (horizontal lines, see Figure~\ref{fig:results_SFFS_CMC_best} for Rank-1 values). Best in colour.}
\label{fig:results_SFFS}
\end{figure}

\begin{figure}[t]
\centering
        \includegraphics[width=0.4\textwidth]{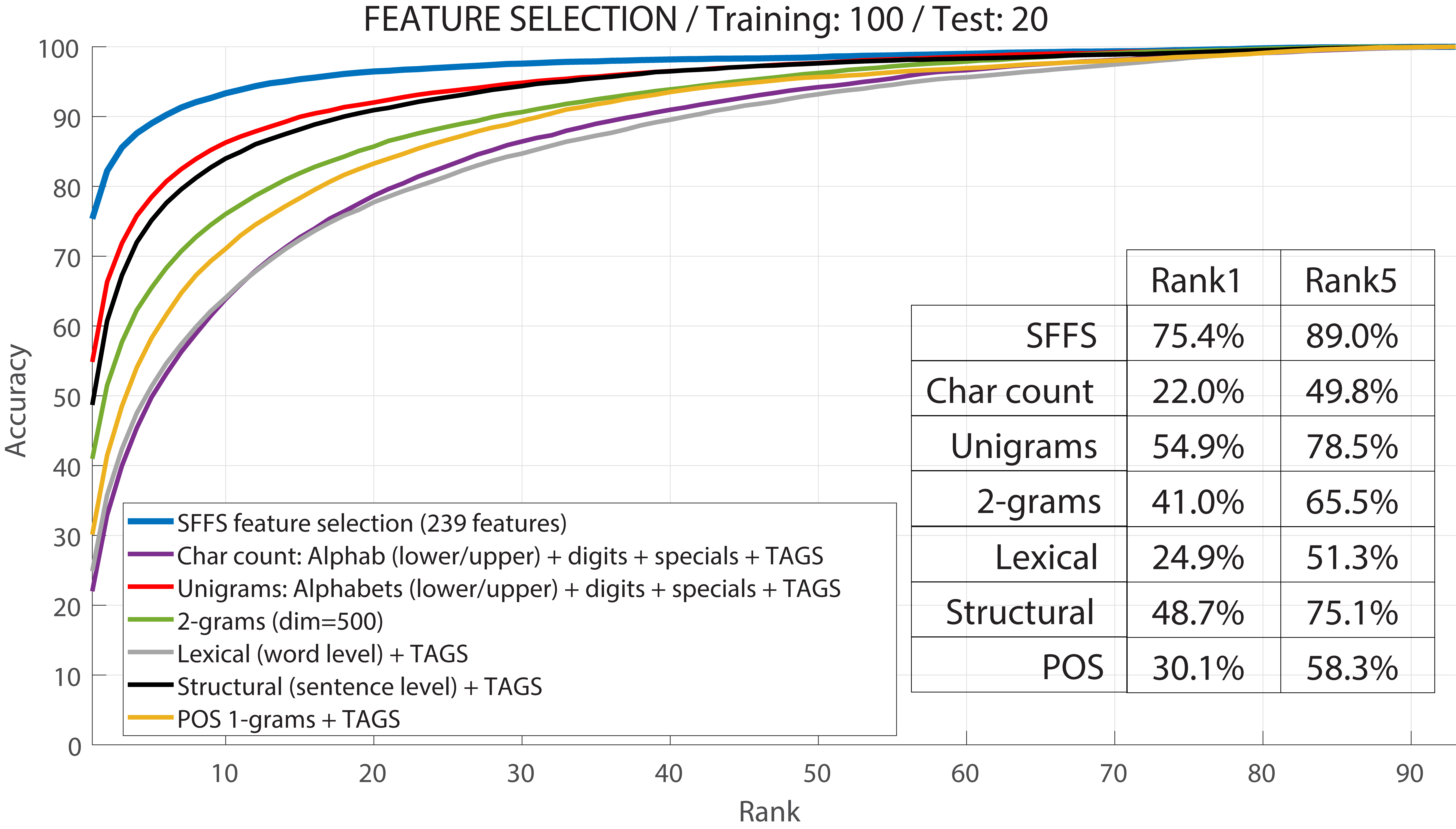}
\caption{Identification accuracy of feature combination with SFFS (training: 100 Tweets, test: 20 Tweets). Results are given for the ISOT \cite{Brocardo14pst} database with 93 test authors. Performance of the best individual features of Sections~\ref{sect:results:char_count}-\ref{sect:results:word_sentence} is also given for reference. Best in colour.}
\label{fig:results_SFFS_CMC_best}
\end{figure}

\begin{figure*}[t]
\centering
        \includegraphics[width=0.9\textwidth]{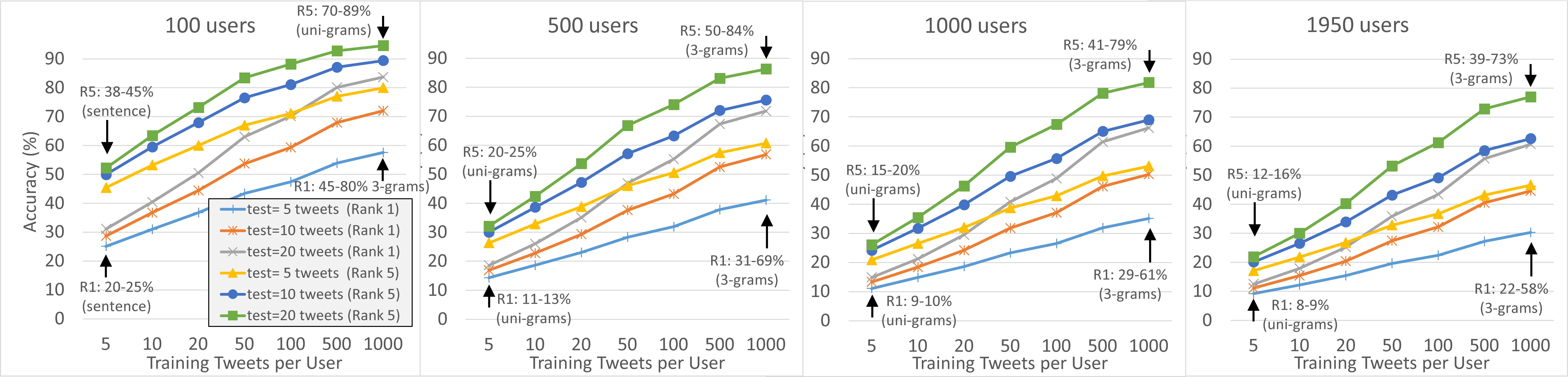}
\caption{Identification accuracy (Rank-1 and Rank-5) of feature combination with SFFS for a reduced number of test Tweets ($\chi^2$ distance). Results are given for the SMF \cite{[Rocha17tifsTweets]} database. Each figure corresponds to a different number of test authors (from left to right: 100, 500, 1000, and 1950 authors). The arrows indicate the best accuracy of the individual features of the previous sub-sections, with indication of which feature is providing such accuracy value. Best in colour.}
\label{fig:results_SFFS_rocha_ranks}
\end{figure*}

\begin{figure}[t]
\centering
        \includegraphics[width=0.4\textwidth]{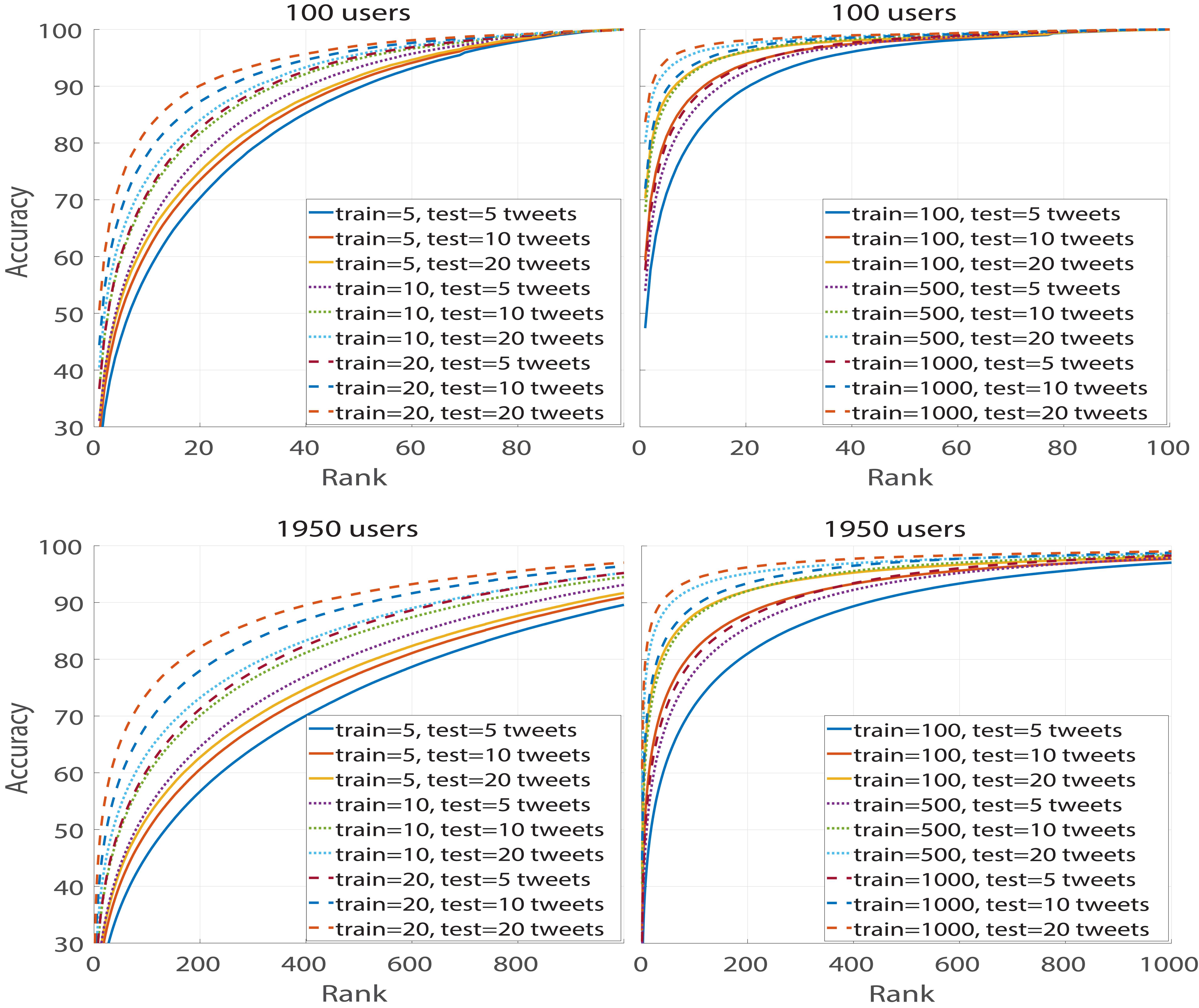}
\caption{Identification accuracy of feature combination with SFFS for a reduced number of test Tweets ($\chi^2$ distance). Results are given for the SMF \cite{[Rocha17tifsTweets]} database. 
Best in colour.}
\label{fig:results_SFFS_rocha_CMCs}
\end{figure}

\begin{figure}[t]
\centering
        \includegraphics[width=0.4\textwidth]{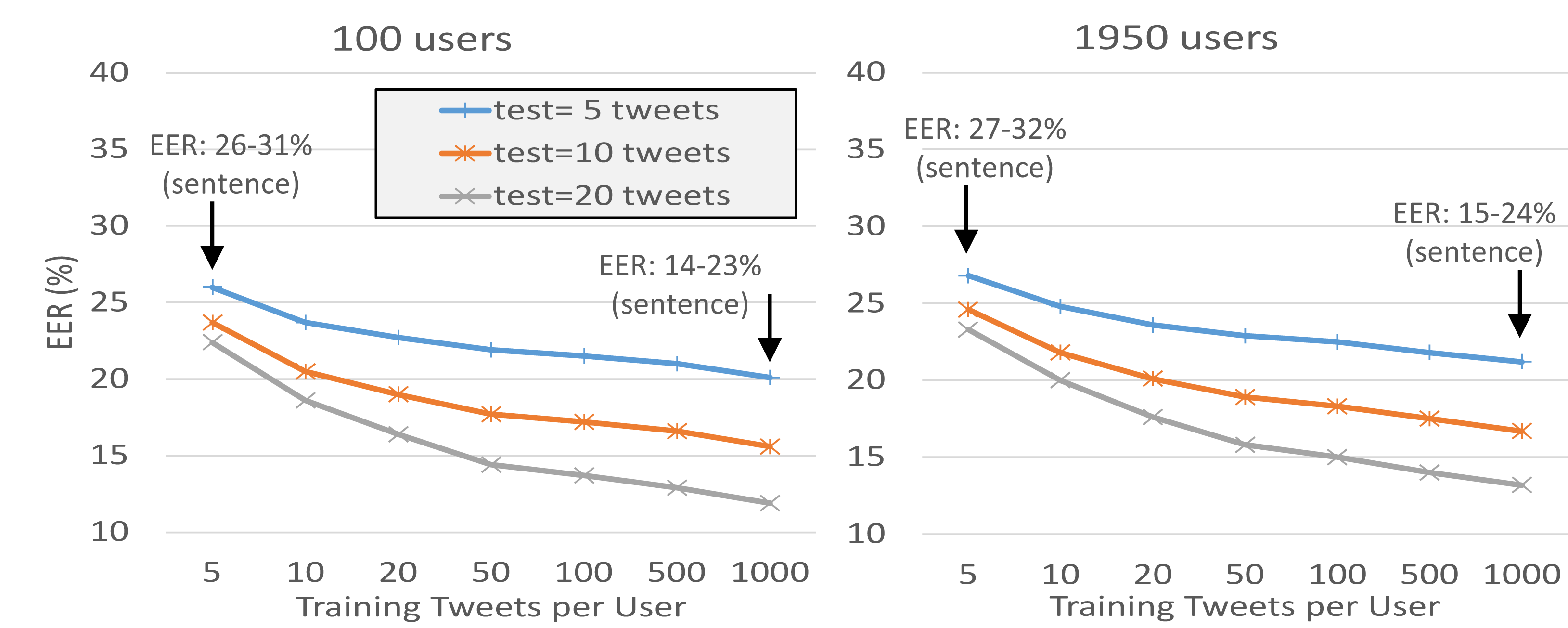}
\caption{Verification accuracy (EER) of feature combination with SFFS for a reduced number of test Tweets ($\chi^2$ distance). Results are given for the SMF \cite{[Rocha17tifsTweets]} database. Each figure corresponds to a different number of test authors. The arrows indicate the best accuracy of the individual features of the previous sub-sections, with indication of which feature is providing such accuracy value. Best in colour.}
\label{fig:results_SFFS_rocha_eer}
\end{figure}

\begin{figure}[t]
\centering
        \includegraphics[width=0.4\textwidth]{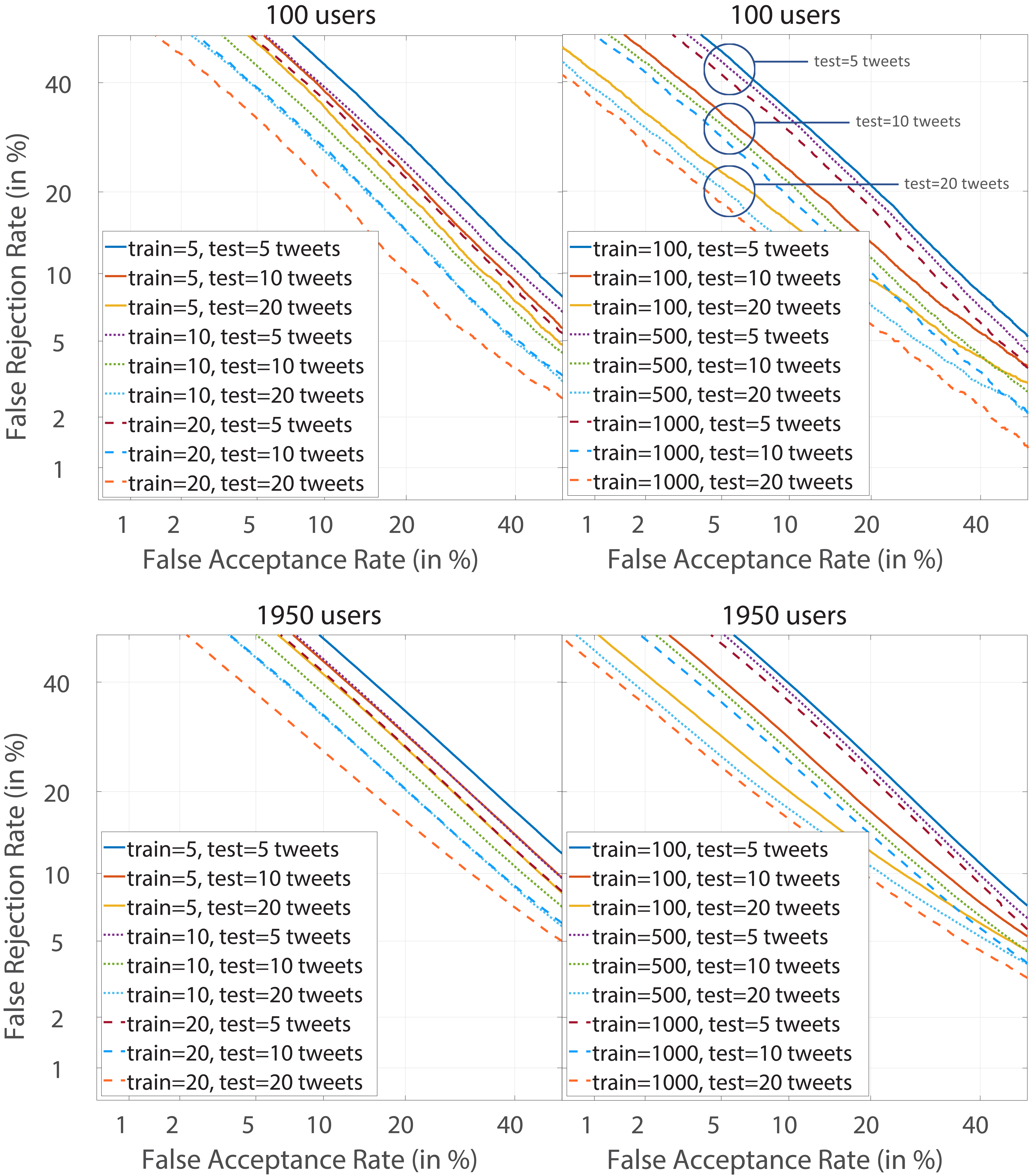}
\caption{Verification accuracy (DET curves) of feature combination with SFFS for a reduced number of test Tweets ($\chi^2$ distance). Results are given for the SMF \cite{[Rocha17tifsTweets]} database. 
Best in colour.}
\label{fig:results_SFFS_rocha_DETs}
\end{figure}

\subsection{Feature Combination by Automatic Selection}
\label{results:feature_selection}

The features considered in this paper work at several levels of analysis, capturing different aspects of writing individuality.
While they are not uncorrelated, improvements can be expected if  combined \cite{[Fierrez18]}.
%
We evaluate their combination at the feature level, concatenating them to form an extended feature vector for each Tweet. Then, 
experiments are carried out with the same protocol than previous sub-sections. 
The best combination to build the extended feature vector is found by Sequential Forward Floating Selection
(SFFS) \cite{[Pudil94]}. Given $n$ features to combine, we employ as
criterion value of the SFFS algorithm the Rank-1 identification accuracy.

In this sub-section, we focus our study on
the case where there is few data available 
(test = 20 Tweets or less), which is more suited to social media forensics, where the amount of available data might be scarce \cite{[Jain15],[Rocha17tifsTweets]}.
%
%
Since the accuracy with few data is observed to be worse, more significant gains can also be expected in comparison to employing a large amount of training and test data.
%
%
For the purposes of this sub-section, we consider all the features presented in Table~\ref{tab:features}, namely: 9 Lexical-Character-Count, 121 Lexical-Character-Frequency uni-grams (note that `white spaces' already appears in the character count features, so it is removed here), 14 Lexical-Word-Count (including meta tags), 15 Structural-Sentence, and 13 Syntantic-Word features (POS tag uni-grams). %
%
%
%
%
%
%
%
%
%
Regarding $n$-grams ($n>$1), the best accuracy is given by tri-grams with a dictionary of 2000 elements (Table~\ref{tab:results_char_ngrams}, Figure~\ref{fig:results_char_ngrams_rocha_ranks}), but to reduce the search space of the SFFS algorithm, we analyze if there is a smaller dictionary with a similar accuracy. Figure~\ref{fig:results_ngrams_dictionary} shows the performance of bi-grams and tri-grams for different dictionary sizes with the ISOT database and 93 authors (training/test Tweets=100/20). Interestingly, bi-grams works better than tri-grams for the considered size range, with the performance stabilizing from 500 elements, and reaching a plateau at around 900 elements. Therefore, for this section, we will consider bi-grams with a dictionary of 500 elements.

According to the described procedure, the SFFS search space consists of 9 + 121 + 14 + 15 + 13 + 500 = 672 features.
Figure~\ref{fig:results_SFFS} 
shows the identification accuracy with the ISOT database and 93 authors (training/test Tweets=100/20) for an increasing number of features selected with SFFS.
The performance increases quickly with the addition of new features, and stabilizes at around 100 features. The peak is reached at 239 features and remains approximately constant until 300 features, after which the performance deteriorates.
Interestingly, only 15 features are enough to surpass the best individual system of the previous sub-sections (unigrams, green horizontal line, with dimensionality 99). 
It can be also seen that even if the combination of all available features (orange line) surpasses any individual system, its performance is sub-optimal in comparison to the optimal set selected by SFFS.
This `peaking' effect, with a quick increase in performance by proper selection of a small number of features, is well documented \cite{[RaudysJain91smallsampleclassifiers]}, and it has been observed in many other biometric studies \cite{[Roli02MCScomparison],[Fierrez05d],[Alonso07],[Garcia-Salicetti06],[Alonso15a]}.
%
%
%
%
Figure~\ref{fig:results_SFFS_CMC_best}
shows the CMC curve with the vector of 239 features selected by SFFS, together with the features that provided the best accuracy in the previous sub-sections. Compared with the best individual system (unigrams, red curve), the accuracy by feature selection is 10-20\% better for small sizes of the output list. 
In just five positions (Rank-5), the accuracy of SFFS reaches 89\%, and with a bigger list (Rank-10/20), it increases up to 93.4/96.5\%. 

Using the features selected by SFFS, we then report experiments with the larger SMF database and a small number of test Tweets (5, 10 and 20).
Identification accuracy is shown in Figures~\ref{fig:results_SFFS_rocha_ranks} and \ref{fig:results_SFFS_rocha_CMCs}, while verification accuracy is shown in Figures~\ref{fig:results_SFFS_rocha_eer} and \ref{fig:results_SFFS_rocha_DETs}. In this case, we also report CMC and DET curves, which where omitted in previous sub-sections due to space.
To assess improvements in identification accuracy, Figure~\ref{fig:results_SFFS_rocha_ranks} should be compared against Figures~\ref{fig:results_char_counts_rocha_ranks}, \ref{fig:results_char_unigrams_rocha_ranks}, \ref{fig:results_char_ngrams_rocha_ranks}, and \ref{fig:results_word_sentence_rocha_ranks}. 
To facilitate the task, we show in Figure~\ref{fig:results_SFFS_rocha_ranks} the best accuracy obtained by the individual features of the previous sub-sections for the two extremes of the $x$-axes (5 and 1000 training Tweets). This allow to see, for example, that the best individual feature with few training Tweets is uni-grams, while the best feature with many training Tweets is tri-grams. 
It can be seen as well that feature fusion provides additional improvements, with the Rank-1 and Rank-5 ranges in general moving towards higher values.
%
%
%
Another phenomenon, that is also observed in previous sub-sections but has not been mentioned yet, is that as the curves progress to the right side of the $x$-axes (i.e. we use more training Tweets), they separate more among them.
In other words, feeding the training set with more Tweets makes the features employed more sensitive to changing the amount of test Tweets. This makes sense because increasing the training set means that more Tweets are combined to create a feature vector, thus averaging Tweets of a more varied style. If the test set contains very few Tweets, we increase the chances that its style is far away from the 'average style' of the training set for that user, becoming more similar to another author of the database (in the feature sense).   
For example, with 1950 authors and 1000 training Tweets, the Rank-1/Rank-5 accuracy with 5 test Tweets is 30/46\%. With 10 test Tweets, it goes up to 44/62\%, and with 20 test Tweets, to 60/78\%. 
While having 1000 training Tweets may not be the norm, the results show that in such case, the features employed here can provide good accuracy with few test data and 1950 authors in the database. Given the average Tweet length of the SMF database (56 characters plus meta-tags), 20 Tweets represent an average of 1120 characters, and 10 Tweets only 560 characters per user.
Additionally, Figure~\ref{fig:results_SFFS_rocha_CMCs} shows the CMC curves of the identification experiments with 100 and 1950 authors. 
It can be seen that the slope of the curves increases quickly with the addition of more training Tweets, contributing to reduce more the search space of potential candidates. 
With 100 authors, a 80\% accuracy with 5 training Tweets is not obtained until Rank-26 (20 test Tweets, yellow solid curve in Figure~\ref{fig:results_SFFS_rocha_CMCs}, top left). With 10 training Tweets, such accuracy can be obtained at Rank-15, and with 20 training Tweets, at Rank-9. All results are with 20 test Tweets. This shows the possibility of reducing the search space to 10\% of the database of less if just 20 Tweets per author are available, and still retrieving the correct author among the candidates with 80\% probability.
If 100 training Tweets are available (top right plot), we could even get an accuracy of 90\% at Rank-7, or 95\% at Rank-16. 
Interestingly, the proportion to which the search space is reduced is kept constant if we increase the number of authors.
In the more difficult case of 1950 authors, 80\% of accuracy with 5 training Tweets is obtained at Rank-529 (27\% of the database, with 100 authors it was Rank-26, or 26\%). With 10 training Tweets, such accuracy can be obtained at Rank-318 (16\% of the database), and with 20 training Tweets, at Rank-166 (8.5\% of the database). Again, all results are with 20 test Tweets. 
For 100 training Tweets (bottom right plot), a 90\% accuracy is obtained at Rank-138 (7\% of the database).
These results indicate that the employed features reduce the search space in the same proportion for a given number of training and test Tweets, regardless of the amount of authors in the database.
%
%
%

Regarding verification experiments, Figure~\ref{fig:results_SFFS_rocha_eer} should be compared against Figures~\ref{fig:results_char_counts_rocha_eer}, \ref{fig:results_char_unigrams_rocha_eer}, \ref{fig:results_char_ngrams_rocha_eer}, and \ref{fig:results_word_sentence_rocha_eer}
to assess to what extent feature combination is capable of improving accuracy.
We facilitate the task by showing in Figure~\ref{fig:results_SFFS_rocha_eer} the best accuracy obtained by the individual features of the previous sub-sections for the two extremes of the $x$-axes. Interestingly, the best individual feature in all cases is the structural (sentence level) type.
In an equal manner as the identification experiments, it can be seen that the feature fusion improves accuracy, with the ranges of the EER moving towards smaller values. Here it is observed as well that by progressing to the right side of the $x$-axes (i.e. using more training Tweets), the curves separate more among them. 
Figure~\ref{fig:results_SFFS_rocha_DETs} shows the DET curves of the verification experiments with 100 and 1950 authors, where the same progress towards a better accuracy as more data is available can be observed as well.
Nevertheless, the accuracy remain poor when very few data is available (e.g. 5 or 10 Tweets for training and testing), highlighting the need for more research investments to solve this difficult problem.

\begin{figure}[t]
\centering
        \includegraphics[width=0.4\textwidth]{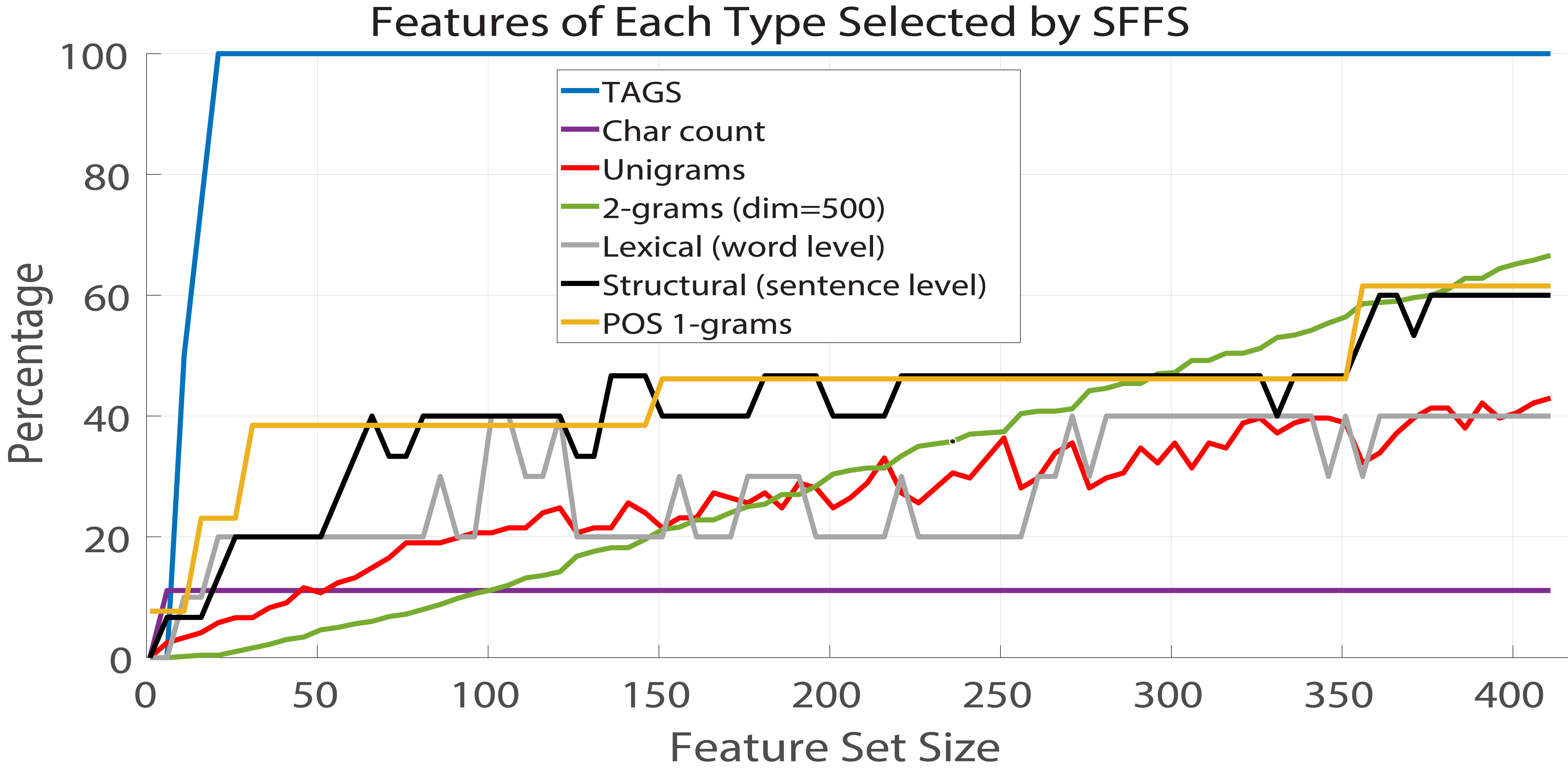}
\caption{Percentage of features from each group selected by SFFS (training: 100 Tweets, test: 20 Tweets). Results are given for the ISOT \cite{Brocardo14pst} database with 93 test authors. Best in colour.}
\label{fig:results_SFFS_feature_percentage}
\end{figure}

Lastly, we analyze to what extent each feature type is being selected by SFFS.
Figure~\ref{fig:results_SFFS_feature_percentage} plots the percentage of features from each group selected as the size of the set increases. For a better assessment, the four meta tags are plotted separately. 
Interestingly, at the beginning, the majority of selected features are from the POS tag and character count sets, which are among the worst performing ones individually. 
At 15 features (when unigrams, the best individual system is surpassed, recall Figure~\ref{fig:results_SFFS}), SFFS selects features from all types, except 2-grams.
At 239 features (performance peak), all groups are represented to some extent. Approximately 47\% of the POS tag and sentence level features are selected, 
40\% of the 2-grams, 30\% of the unigrams, and 20\% of the lexical word level features. 
This highlight the complementarity of the features employed, with all levels of stylometric analysis being retained to some extent in the combination.
All meta tags features are also selected in the majority of the search range. 
An interesting observation is that POS tag, which are not among the most accurate features, are one of the most preferred when combined with others. 
The representation of lexical word level features, another of the worst performing individual groups, is not negligible either. 
On the other hand, char count (the worst individual group) is the least selected group, with only 11\% of the features taken at any point.

\begin{table}[htb]
\begin{center}
\begin{tabular}{|c|c|c|c|c|}



\multicolumn{5}{c}{} \\ \hline

& &   & \multicolumn{2}{c|}{\textbf{Comparison}} \\  \cline{4-5}

 & \textbf{Vector} &  \textbf{Extrac-} & 1000/ & 100/ \\ 

Feature &  \textbf{Size} & \textbf{tion} & 1000 & 20 \\ \hline \hline

Pre-processing  & - & 8.99 ms (10\%) & - & -  \\ \hline

Character Count  & 8 & 0.31 ms (0.3\%) & 2.49 $\mu$s &  0.48 $\mu$s \\ \hline

Uni-grams  & 99 & 0.88 ms (1\%) & 7.75 $\mu$s &  1.78 $\mu$s  \\ \hline

2-grams  & 500 &  5.99 ms (6.7\%) & 48.18 $\mu$s &  4.31 $\mu$s  \\ \hline

Lexical-Word  & 14 & 3.5 ms (3.9\%) & 1.79 $\mu$s &  0.45 $\mu$s\\ \hline

Sentence  & 19 &  34.5 ms (38.6\%) & 1.76 $\mu$s &  0.45 $\mu$s \\ \hline

POS 1-grams  & 17 & 35.3 ms (39.5\%) & 1.73 $\mu$s &  0.45 $\mu$s  \\  \hline \hline 

TOTAL  & - &  89.47 ms & 63.7 $\mu$s & 7.92 $\mu$s  \\  \hline

\end{tabular}

\end{center}
\caption{Feature extraction (average per Tweet) and comparison times (average per distance computation between an enrolment and a test vector). Results are given for the ISOT \cite{Brocardo14pst} database.}
\label{tab:time_complexity}
\end{table}
\normalsize

%
%

\subsection{Time Complexity}
\label{results:time_complexity}

Table~\ref{tab:time_complexity} shows the extraction and comparison times using the $\chi^2$ distance of the different features considered (best combinations obtained in the previous sub-sections, indicated in Figure~\ref{fig:results_SFFS_CMC_best}).  
Experiments have been done with Matlab r2019b using a Dell Latitude 7390 laptop with an i7-8650U (1.9 GHz)
processor, and 16 Gb DDR4 RAM. 
The OS is Microsoft Windows 10 Enterprise.
The extraction time is in the range of milliseconds per Tweet, and it scales with the size of the feature vector for features involving only character analysis (rows 2-4).
The last three rows refer to features that work at the word or sentence level. Their higher computation times are because they require to erase punctuation, tokenize the text to extract words and sentences, and compute part-of-speech tags.
If all features are to be extracted, the total time per Tweet is of 89.47 ms. 
Pre-processing (Section~\ref{sect:prepro}) takes 8.99 ms per Tweet on average, representing 10\% of the process. 
Considering the size of the ISOT database (241,985 Tweets), it takes 6 hours on average to pre-process and extract all the features, and 194 hours with the SMF database.
This is not a problem in forensic analysis, since computations are mostly done off-line, and real time is not a requisite \cite{[Jain15]}. 
Vector comparison is made very efficiently (in microseconds), since it involves templates of fixed size and a fixed number of calculations using a distance measure. 
Enrolment and test vectors are generated by averaging feature vectors of several Tweets (according to the protocol of Section~\ref{sect:db_protocol}). This explains the higher comparison times in the 1000/1000 column (1000 Tweets for training and testing), with respect to the 100/20 column, since the vectors have to be averaged prior to distance computation. The comparison times indicated in Table~\ref{tab:time_complexity} only considers averaging of the test vectors, since training vectors would be usually pre-computed off-line.
In a real case-work scenario, it would only be necessary to extract features from the test Tweets (89.47 ms per Tweet), average them, and compare against the database. 
The features employed would scale well with bigger database sizes. With 1M users, for example, a complete comparison with the entire database in the 100/20 case
would take 7.92 seconds in a regular laptop. 
%

\subsection{Feature Permanence Over Time}
\label{results:time_analysis}

\begin{figure*}[t]
\centering
        \includegraphics[width=0.9\textwidth]{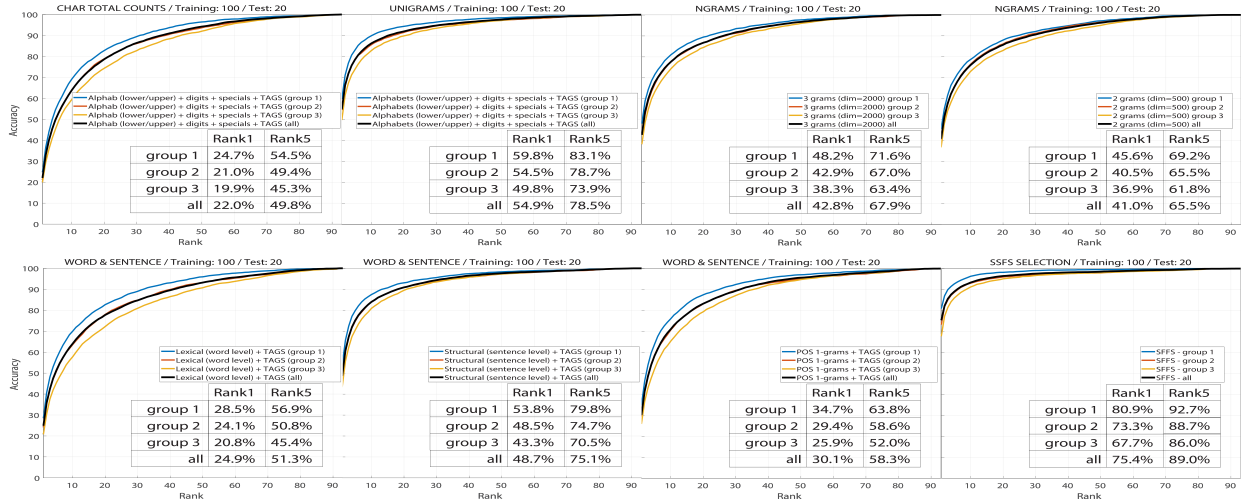}
\caption{Identification accuracy of different features for different time spans between training and test Tweets using the $\chi^2$ distance (training: 100 Tweets, test: 20 Tweets). Results are given for the ISOT \cite{Brocardo14pst} database with 93 test authors. Best in colour.}
\label{fig:results_time_span}
\end{figure*}

Finally, 
we analyze the permanence of the employed features over time. For this, we divide the test Tweets of each author of the ISOT database into three disjoint sets.
The Tweets 
are in chronological order, so the first group is closer in time to the training Tweets, while the Tweets of the third group are the furthest.
Figure~\ref{fig:results_time_span} reports the identification performance with the ISOT database of the features that provided the best performance in Sections~\ref{sect:results:char_count}-\ref{sect:results:word_sentence} (indicated in Figure~\ref{fig:results_SFFS_CMC_best}), as well as the optimal feature set found in Section~\ref{results:feature_selection}.
The results correspond to the case of 100 Tweets for training and 20 for testing.
Although the performance of all features decreases over time, their robustness is different (measured by the observed relative separation between CMC curves).
The biggest differences can be seen in Character Count (top left), Lexical Word (bottom left), and POS tag (bottom, third column). With these features, the variation of Rank-5, for example, is of 9-14\% between groups 1 and 3.
With $n$-grams, such variation is of 7.4-8.2\%, and with SFFS selection, it is of 6.7\%. The latter two types of features have the biggest feature vectors, which may be behind their higher resilience to time variability.
Feature stability for social media authorship analysis has been hardly explored \cite{[Rocha17tifsTweets]}.
While the results of this study can be considered promising (given that the 
vast majority of the users has a time span between Tweets of 6 months of more, Figure~\ref{fig:dbstats_time}), they highlight the need for methods that counteract variability of the writing style over time \cite{[Azarbonyad15tweet_time_evolution]}.
%
%

\section{Conclusion}
\label{sect:conclusion}

Authorship attribution of texts has its roots in the 19th century, well before the technology revolution.
In the early days, the target was to identify the author of literary works \cite{Williams75} or essays \cite{HOLMES95} by stylometric techniques \cite{Rong2006}. 
Nowadays, digital texts are massively in use, for example short messages, emails, blogs, etc.
Thus, 
authorship analysis has evolved 
to other fields such as cybercrime, law enforcement, education, fraud detection, etc.
In this context, it is of huge interest developing methods for authorship attribution to aid in forensic investigation of cybercrimes \cite{Nirkhi2013}.

Accordingly, we address the topic of authorship attribution of digital text. 
In popular social media platforms, 
the amount of text available might be limited, and some even limit the number of character per message to a few hundred.
Here, we concentrate on short texts (Twitter posts), which are currently limited to 280 characters, and to 140 at the time when our databases were captured.
We evaluate several feature experts based on stylometric features (Table~\ref{tab:features}), which are among the most widely used features both in traditional literary analysis and in more modern ways of writing \cite{Williams75,Nirkhi2013,Bouanani2014,Stamatatos2009}.
They capture properties of the writing style at different levels, such as: $i$) the number of individual characters, the frequency of individual characters (uni-grams), the frequency of sequences of characters ($n$-grams, $n>$1), or the type of words than an individual uses (\textit{lexical} level); $ii$) the way that the writer organizes the sentences (\textit{structural} level); or $iii$) the use of different categories of words (\textit{syntantic} level).
An advantage of these features is that the comparison between feature vectors is made simply by distance measures.
They also allow coping with texts of different lengths, since the features are extracted per post. When several posts from the same writer are combined for a richer identity model, the feature vectors just have to be averaged.
We have left out some features that are of less applicability to Twitter posts, 
such as properties related to paragraphs (given the limited length of Tweets),
or to the analysis of keywords of other particularities (given the generality of Twitter posts, which in principle are not restricted to any topic).
We have also added to our feature set several meta tags specific of Twitter text. They quantify the use of URLs, the mentions to other users, the use of hashtags, or the quotation of somebody's else text.
Although these tags are widely employed by Twitter users, they have been hardly employed in authorship studies \cite{Sultana14icc,Sultana17thms,Sultana18icsmc}.
In our experiments, we have observed a significant boost in performance when these particular features are considered.

Experimental results are given with two databases in English language: the ISOT database, with 93 authors from the list of some of the UK's most influential tweeters \cite{Brocardo14pst}, and the Social Media Forensics (SMF) database, with 3957 authors from the general public \cite{[Rocha17tifsTweets]}.
The database of influential tweeters contain more messages per author (2692 vs. 1972), and more characters per message (68 vs. 56, without considering meta-tags). Influential tweeters also show higher preference for the use of meta-tags specific of Twitter text.
This highlights the expected higher activity level in social media of such writers in comparison to the general public.
%
%
Identification accuracy of the features are assessed individually, with those capturing the frequency of individual characters (uni-grams) or sequences of characters ($n$-grams, $n>$1) showing superior performance.
This back-ups the popularity of $n$-grams \cite{Nirkhi2013,Bouanani2014,Stamatatos2009}.
In our first investigation with each feature type, we consider the 93 authors of the ISOT database, and a varying number of enrolment/test Tweets per user between 1000 and 20. 
%
%
When a high number of Tweets is employed, the Rank-1 accuracy of uni-grams or $n$-grams easily reaches 97-99\%.
In a more restricted case, e.g. only 20 Tweets for testing, Rank-1 ranges from 54.9\% to 70.6\%, depending on the number of training Tweets.
Regarding the other features, their performance is comparatively worse, and could be ranked as (from better to worse): features at the sentence level (structural), word category level (syntactic), word level (lexical), and count of characters (lexical).
It is worth noting that, for example, sentence level features work better than word level features, even if it is expected that Twitter posts have few sentences.
It is also observed that the performance of the individual features correlates somehow with their dimensionality, although a higher feature dimensionality in biometrics does not necessarily imply a better accuracy, but we point out as an observation that would deserve further analysis.

Next, we evaluate the features with an increasing number of authors from the SMF database (up to 1950), and by reducing further the amount of training and test Tweets (down to only 5). 
First, it is observed that when the same number of authors are employed (93), both databases show a comparable performance with the majority of features. Two remarkable exceptions are word-level features and POS tag uni-grams, suggesting idiosyncratic differences in the use of language words between the two populations. However, all the other features (which mainly work at the character level) remain resilient to the potential differences in writing style between influential tweeters and the general public. 
Another effect that is to be expected is that as
the number of authors increases, the identification accuracy
decreases accordingly. This is exacerbated if the number of training and test Tweets is low (e.g. 5). In such case, the Rank-1 accuracy of the best features is 20\% with 100 authors, and it is reduced to 9\% with 1950 authors, highlighting the difficulty of the problem \cite{[Rocha17tifsTweets]}.
For completeness \cite{[DeCannRoss13btasROCCMCmenagerie]}, we also report author verification results by rearranging ranked identification scores into mated and non-mated verification scores. 
The experiments show that the size of the database does not have the same impact on the verification performance, with the EER not increasing substantially when augmenting the number of authors. Nevertheless, the EER of the best features is above 25\% with very few training and test data (5 Tweets). With a large amount of training Tweets (1000), the EER can be brought down to less than 14\% (with 20 test Tweets), but such amount of training data cannot be expected to be always available. 

We also study the benefit of feature combination when there are few data available (20 Tweets or less for testing).
%
%
By applying automatic feature selection techniques, 
performance is shown to improve accuracy at all levels of database size (number of authors) and amount of training and test data, showing the potential of complementarity of the employed features. With both 20 Tweets for training and testing, the CMC curves show that the search space of candidates can be reduced to 9\% of the database with feature combination, and still retrieving the correct author among the candidates with 80\% probability. 
With 100 training Tweets, the space search is reduced to 7\% with a probability of retrieving the correct author of 90\%.
These percentages have been validated in this paper up to a database size of 1950 authors.
Even if the correct author does not come in the first positions with very high accuracy, these results allow significant time savings in the number of candidate accounts that should be examined by a forensic expert.
%
%
In terms of time complexity, the entire set of features can be extracted in less than 90 ms per Tweet with a regular laptop, and comparison times are in the range of microseconds, which would scale well to bigger databases with dozens of thousands or even millions of users, specially if more powerful dedicated hardware is employed.
%
The employed features have been also observed to be susceptible to evolution of writing style over time, although their robustness is different. In general, there is a correlation between feature dimensionality and resilience to time variability. The database employed with 93 authors contains users whose Tweets spans from 1.5 months up to 33 months, with the majority of users spanning more than 6 months, and half of the database spanning more than 12 months. With this setup, observed performance drops (Rank-5 accuracy) are between 7 and 14\% depending on the feature type (with 100 training and 20 test Tweets). 


The majority of features employed are language-independent, so this study is readily applicable to other languages, e.g. \cite{Abbasi05,Cristani12,Ragel13,Marukatat14}.
The analysis of these results 
suggests that the proposed
approach can be effectively used for writer identification using short digital texts if the amount of authors is small (a few hundreds). With larger database sizes, getting the correct author in the first positions of the output list is still far from achievable. Our analysis, however, shows that forensic investigators can be helped by greatly reducing the search space of candidates. Works like the present one are just the beginning of research considering databases with several thousands of authors \cite{[Rocha17tifsTweets]}.
%
%
Previous studies consider a few dozens of users, or over two hundreds at most \cite{Donais13,Ragel13,[Azarbonyad15tweet_time_evolution],Sultana18icsmc,Belvisi20}. 
%
%
The work where the SMF database 
is presented \cite{[Rocha17tifsTweets]} employs 50 authors in the majority of the paper, scaling up to 1000 authors in one of the experiments only.
In a recent paper by authors of the SMF database \cite{[Rocha19icasspTweets]}, they captured a database of 50K authors, but they restricted the experiments to only 50 authors.
Therefore, future work includes the use of 
a bigger database which simultaneously contain texts posted over a longer period, so as to test the correlation between accuracy and authors set size \cite{Okuno14,Ragel13}, as well as counteract time variability efficiently \cite{[Azarbonyad15tweet_time_evolution]}. 
%
%
%
%
Our databases are public distributions made by another researchers, so the study can be replicated or compared. 
Another avenue for improvement will be the application of trained classifiers such as Support Vector Machines \cite{[Vapnik95]} or Random Forests \cite{[Ma05rffusion]}. 

An issue in forensic contexts is the imbalance between classes, and high variability in the amount of available data per person can be expected, both in the training database of suspects, and in the test data collected during investigations. 
To study this more thoroughly, machine learning methods exist which can deal with class imbalance \cite{Stamatatos08classimbalancetext}, constituting another source of future work. 
%
%
In this direction, we are also considering the study of user-dependent selection approaches \cite{[Fierrez05ePRL]}, so that the features that are most discriminative for each user are employed, allowing to better counteract class imbalance or time variability.
It will also be of interest to couple the stylometric features employed in this paper with interaction patterns and temporal behaviour of Twitter users
\cite{Sultana14icc,Sultana17thms,Sultana18icsmc} in order to assess if performance can be further improved, specially when few text data is available.
The latter would also allow to study connections between users, since people having common interests may have common vocabularies or topics, that could impact the performance or utility of the features employed.
In a similar vein, deep learning solutions capable of learning dependencies of temporal data such as texts can be another promising avenue to solve scalability, class imbalance or time variability. However, these solutions have been proposed in the context of topic classification \cite{Sari19textLSTM} or sentiment analysis \cite{Goularas19TwitterDLsentiment}, needing proper adaptation to the identity classification problem.

We are also aware of the limitations of our work.
Our study has revealed some idiosyncratic differences in the use of language words between influential writers and the general public, as shown by the different performance obtained with the features that analyze the use of words at lexical and syntactic levels.
%
%
%
It is possible that public figures have developed their own, distinct writing style that distinguishes them from the general public, in order to enhance their influence.
Intuitively, this will lead to a different use of vocabulary, but we suspect that such differences can be traced down even to the individual author level, or maybe to a group of authors (even from the general public) that share common interests or interact regularly among them in social media. 
In this sense, incorporating patterns of interaction and behaviour can be a decisive aid \cite{Sultana14icc,Sultana17thms,Sultana18icsmc} that we will explore.
%
%
%
Also, our database does not have `impostors' which try to imitate another author. How successfully a 
`criminal' can impost the writing style of an innocent celebrity or regular citizen while committing a crime is thus a source of future study.
It can be possible as well that several writers produce Tweets pretending to be the same person. 
Celebrities may pay marketing agencies to write Tweets on their behalf and help them to build an online presence. There are semi-automatic software tools that aid in this purpose too, which are accessible by the general public as well. One would expect that this inevitably produces deviations in the form of bigger user intra-variability if several people post as if they were the same person. 
Conversely, several criminals can act together, which maybe could help to improve the chances of detection if the deviation introduced by each criminal contributes towards separating from the writing style of the person that they try to mimic.
The possibility of having multiple writers both in the \textit{genuine} and \textit{impostor} sides is, therefore, a challenge which would deserve further attention.
In social network contexts, data includes a variety of sources, including images or videos, and text from different sources (Twitter, Facebook, Instagram, Whats-app, blogs, chats, etc.), suggesting a hybrid approach beyond just texts \cite{[Rocha17tifsTweets]} that could help to overcome some of the above-mentioned issues. However, such approach exacerbates even more the mentioned limitations in the distribution of data from such platforms, or in their collection using public APIs.


%

\ifCLASSOPTIONcompsoc
  \section*{Acknowledgments}
\else
  \section*{Acknowledgment}
\fi

This work was supported in part by the project 2016-03497 of the Swedish Research Council. Naveed Muhammad has been funded by European Social Fund via IT Academy programme. The authors also thank the CAISR Program of the Swedish Knowledge Foundation.

\ifCLASSOPTIONcaptionsoff
  \newpage
\fi



\bibliographystyle{IEEEtran}
%

%
%

\bibliography{fernando1,bibliography_twitter_ID}

%
\begin{IEEEbiography}[{\includegraphics[width=1in,height=1.25in,clip,keepaspectratio]{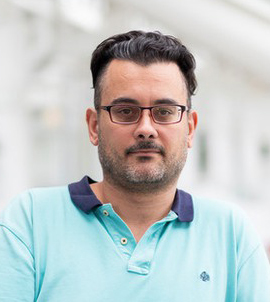}}]{Dr. Fernando Alonso-Fernandez}
received the M.S./Ph.D. in telecommunications from Universidad Politecnica de Madrid, Spain,
in 2003/2008. Since 2010, he is with 
Research, 
Halmstad University, Sweden, first as the
recipient of a Marie Curie IEF and a Post-Doctoral
Fellowship of the Swedish Research Council, and
later as the recipient of a Project Research Grant
for Junior Researchers of the Swedish Research
Council. Since 2017, he is Associate
Professor at Halmstad University. He has been involved in multiple
EU and National projects
focused on biometrics and human–machine interaction. He has over 90 international
contributions at refereed conferences and journals, and
several book chapters. His interests include signal and image processing,
pattern recognition, and biometrics. He also co-chaired ICB2016, the 9th IAPR Intl
Conference on Biometrics.
\end{IEEEbiography}

\begin{IEEEbiographynophoto}{Nicole Mariah Sharon Belvisi}
earned her Master in Network Forensics from Halmstad University, Sweden, in 2019. Her master research covers the domains of digital forensics and textual analysis.
\end{IEEEbiographynophoto}

\begin{IEEEbiography}[{\includegraphics[width=1in,height=1.25in,clip,keepaspectratio]{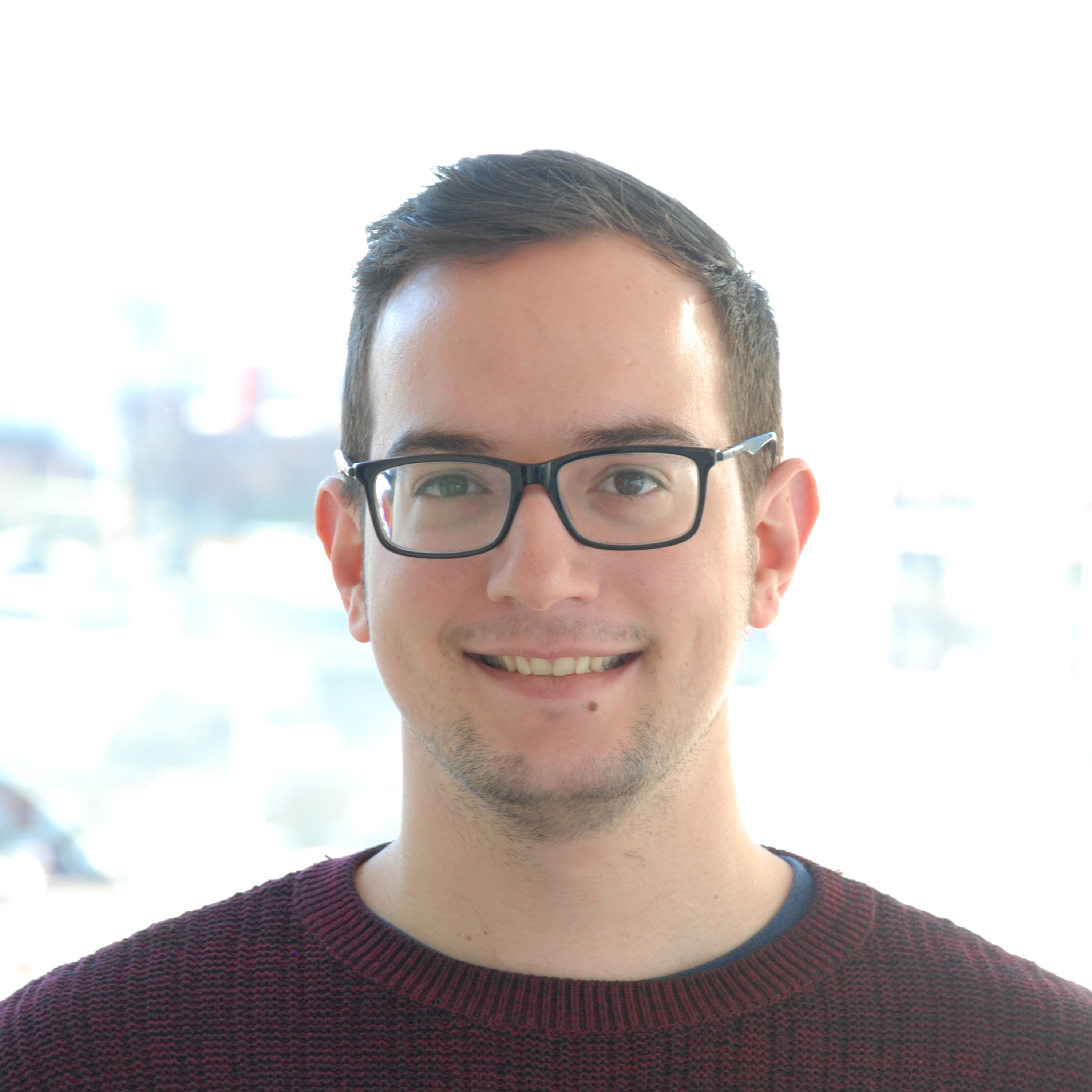}}]{Kevin Hernandez-Diaz}
received the B.S. in telecommunication engineering from Universidad de Sevilla, Spain (2016), and the MSc in Data Science and Computer Engineering from Universidad de Granada, Spain (2017). Since 2018, he is a Ph.D. candidate of the Centre for Applied Intelligent Systems Research, Halmstad University, Sweden. His interests include computer vision, pattern recognition, signal processing and biometrics. 
\end{IEEEbiography}

\begin{IEEEbiography}[{\includegraphics[width=1in,height=1.25in,clip,keepaspectratio]{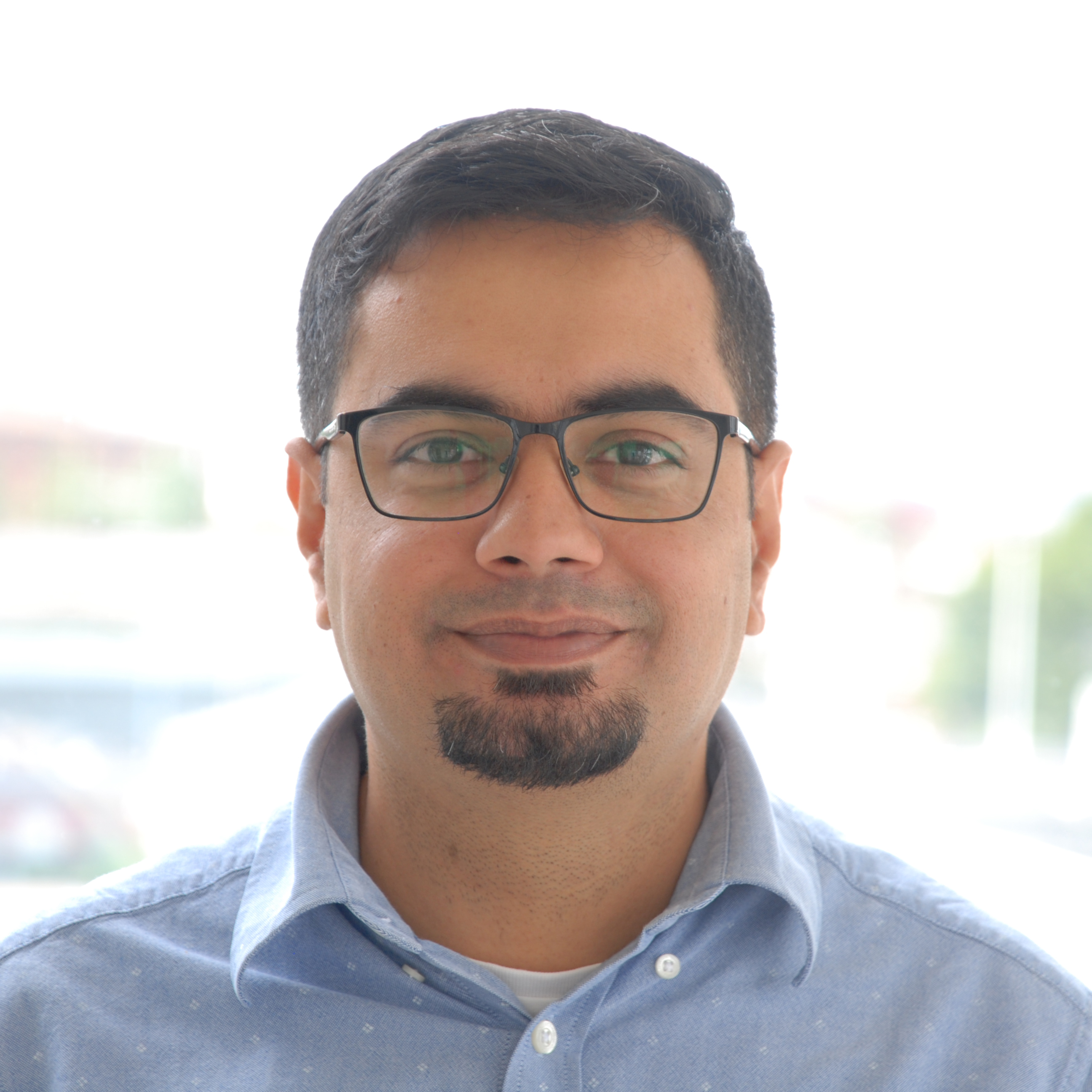}}]{Dr. Naveed Muhammad}
is assistant professor of autonomous driving at the University of Tartu, Estonia. He earned his PhD in robotics from INSA de Toulouse (research stay at LAAS-CNRS), France, in 2012. His master and bachelor education are in vision and robotics (European VIBOT masters), and mechatronics, respectively. He has had post-doctoral stays at Tallinn University of Technology, Estonia, and Halmstad University, Sweden, and has held faculty positions at National University of Sciences and Technology, Pakistan and Asian Institute of Technology, Thailand. His interests include autonomous driving, navigation, behaviour modeling, classification and prediction.
\end{IEEEbiography}

\begin{IEEEbiography}[{\includegraphics[width=1in,height=1.25in,clip,keepaspectratio]{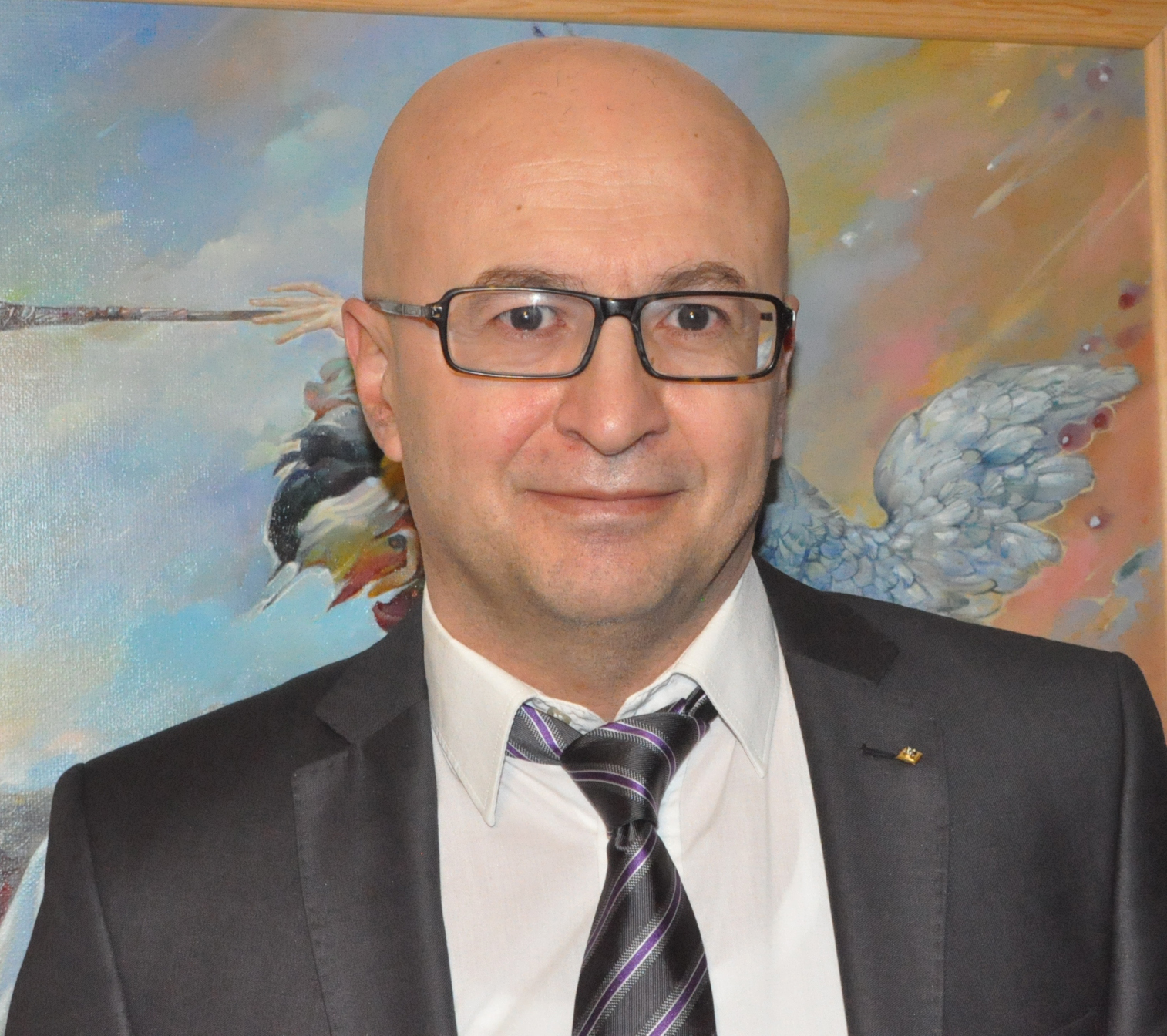}}]{Prof. Josef Bigun}
(M’88,SM’98,FM'03) obtained M.S./Ph.D. degrees from Linköping University, Sweden, in 1983/1988. Between
1988-1998, he was faculty member of the EPFL, Switzerland, as 'Adjoint Scientfique'. He was elected Professor to the Signal Analysis Chair (current position) at Halmstad University. His scientific interests include a broad field in computer vision, texture and motion analysis, biometrics, and the understanding of biological recognition mechanisms. Dr. Bigun has co-chaired several international conferences and contributed to initiate the ongoing Int. Conference on Biometrics, formerly known as Audio and Video based Biometric Person Authentication in 1997. He has been contributing as editorial board member of journals, including Pattern Recognition Letters, IEEE Trans. on Image Processing, and Image and Vision Computing. He has been keynote speaker of several international conferences on pattern recognition, computer vision and biometrics, including ICPR. He has served on the executive committees of several associations, including IAPR, and as expert for research evaluations, including Sweden, Norway, and EU-countries. He 
is 'Fellow' of IAPR and IEEE.

\end{IEEEbiography}






\end{document}